\documentclass[10pt,journal,compsoc]{IEEEtran}
\usepackage{color}

\ifCLASSOPTIONcompsoc
\usepackage[nocompress]{cite}
\else
\usepackage{cite}
\fi

\usepackage{multirow}
\usepackage{pdflscape}
\usepackage{booktabs}
\usepackage{enumitem}
\usepackage{amsfonts}
\usepackage{amssymb}
\usepackage{lscape}
\usepackage{amsmath}

\usepackage[ruled]{algorithm2e}
\usepackage[]{algorithmic}

\usepackage{esint}
\usepackage{xcolor}
\usepackage{units}
\usepackage{textcomp}
\usepackage{epigraph}
\usepackage{makecell}
\usepackage{tikz}
\usetikzlibrary{decorations.pathmorphing,calc,shapes,arrows,snakes,shapes.geometric,patterns, decorations.markings}
\usepackage{pstricks-add}
\usepackage{pst-node}

\usepackage[normalem]{ulem}
\usepackage{cancel}

\usepackage{hyperref}

\usepackage{xr}
\newtheorem{definition}{Definition}[section]

\graphicspath{{figures/}{figures/spatial_registration/}{figures/}{figures/Faces/}{figures/stats/human1/}{figures/stats/human2/}{figures/stats/face2/}{figures/teaser/}}

\newcommand{\parameter}{s}						

\newcommand{\surfaces}{\Space{F}} 				

\newcommand{\surface}{f} 						
\newcommand{\srnf}{h}							
\newcommand{\srnfmap}{H}                        			

\newcommand{\srvf}{q}							
\newcommand{\srvfmap}{Q}                        			
\newcommand{\curve}{\alpha} 						
\newcommand{\srnfcurve}{\beta} 					

\newcommand{\curvebetweenanim}{\Lambda}				
\newcommand{\curvebetweensrnfanim}{\curvebetweenanim_{\srnfcurve}}	
\newcommand{\curvebetweensrvfanim}{\curvebetweenanim_{\srvf}}	

\newcommand{\curveone}{\curve_1} 					
\newcommand{\curvetwo}{\curve_2} 	

\newcommand{\srnfcurveone}{\srnfcurve_1}
\newcommand{\srnfcurvetwo}{\srnfcurve_2}

\newcommand{\srvfone}{\srvf_1}
\newcommand{\srvftwo}{\srvf_2}

\newcommand{\meancurve}{\bar{\curve}}

\newcommand{\meansrvf}{\bar{\srvf}}

\newcommand{\jacobian}{J}

\newcommand{\preshapes}{\mathcal{C}}            		
\newcommand{\preshapesq}{\preshapes_{\srnf}}		
\newcommand{\srnfs}{\preshapesq}					

\newcommand{\curves}{\ensuremath{\mathcal{M}_{\surfaces}}}		
\newcommand{\srvfs}{\Space{Q}}                  			
\newcommand{\srnfcurves}{\mathcal{M}_{\srnf}} 		

\newcommand{\normalfield}{\textbf{n}}                   		 

\newcommand{\rotations}{SO(3)}                  			
\newcommand{\rotation}{O}                       				

\newcommand{\diffeos}{\Gamma}                   			
\newcommand{\diffeo}{\gamma}                    			

\newcommand{\cov}{K}							
\newcommand{\eigenval}{\sigma}
\newcommand{\eigenvect}{\Sigma}					

\newcommand{\nsurfaces}{n}

\newcommand{\noi}{\noindent}

\newcommand{\mespace}{\:}

\newcommand{\intspace}{\mespace}
\newcommand{\isp}{\intspace}

\newcommand{\eg}{{e.g.,\  }}
\newcommand{\ie}{\emph{i.e.,\ }}

\newcommand{\etal}{\emph{et al.}}

\usepackage{framed, color}
\definecolor{shadecolor}{rgb}{1,0.5,0}

\newcommand{\Space}[1]{\ensuremath{\mathcal{#1}}}

\newcommand{\real}{\mathbb{R}}

\newcommand{\rnonneg}{\real_{\geq 0}}

\newcommand{\rcubed}{\real^{3}}
\newcommand{\rthree}{\rcubed}
\newcommand{\rnth}{\real^{n}}

\newcommand{\s}{\ensuremath{\mathbb{S}}}
\newcommand{\stwo}{\ensuremath{\mathbb{S}^2}}

\newcommand{\sn}[1]{\s^{n}}
\newcommand{\domain}{\Omega}					

\newcommand{\ltwo}{\mathbb{L}^{2}}

\newcommand{\curvediffeos}{\Xi}
\newcommand{\curvediffeo}{\xi}

\newcommand{\cB}{{\mathcal B}}

\newcommand{\id}{\text{id}}

\newcommand{\map}{\to}

\newcommand{\norm}[1]{\lVert #1 \rVert}

\newcommand{\inner}[2]{\langle #1,#2 \rangle}

\newcommand{\innertwo}[2]{\langle #1, #2 \rangle_{2} }

\newcommand{\E}{E}

\newcommand{\Ereg}{\E_{\text{reg}}}

\def\argmin{\mathop{\rm argmin}}

\def\cov{K}

\usepackage{todonotes}

\begin{document}
\bstctlcite{IEEEexample:BSTcontrol}

\title{4D Atlas: Statistical Analysis of the Spatiotemporal Variability in Longitudinal 3D Shape Data}
\author{Hamid Laga,  Marcel Padilla, Ian H. Jermyn,  Sebastian Kurtek,  Mohammed Bennamoun~\IEEEmembership{Senior Member,~IEEE}  \and Anuj Srivastava
	\IEEEcompsocitemizethanks{
		\IEEEcompsocthanksitem Hamid Laga  is with the Information Technology Discipline, Murdoch University, Murdoch, 6150 (Australia), with the Harry Butler Institute, Murdoch University, Murdoch, 6150 (Australia), Email: H.Laga@murdoch.edu.au, and with the University of South Australia, The Phenomics and Bioinformatics Research Centre, SA 5000, Australia.

		\IEEEcompsocthanksitem Marcel Padilla is with TU Berlin, Germany. Email: padilla@math.tu-berlin.de
		
		\IEEEcompsocthanksitem Ian H. Jermyn is with Durham University. Email: i.h.jermyn@durham.ac.uk
		
		\IEEEcompsocthanksitem Sebastian Kurtek is with the Ohio State University, US. Email: kurtek.1@stat.osu.edu
		
		\IEEEcompsocthanksitem Mohammed Bennamoun is with the University of Western Australia, Perth, WA 6009, Australia. Email: mohammed.bennamoun@uwa.edu.au
		
		\IEEEcompsocthanksitem Anuj Srivasta is with Florida State University, US. Email: anuj@stat.fsu.edu
		
	}
	\thanks{Manuscript received XXXX, 2021; revised XXX, 2021.}}

\IEEEcompsoctitleabstractindextext{
\begin{abstract}
We propose a novel framework to learn the spatiotemporal variability in longitudinal 3D shape data sets, which contain observations of objects that evolve and deform over time.  This problem is challenging since surfaces come with arbitrary parameterizations and thus, they need to be spatially registered. Also, different deforming objects, also called \emph{4D surfaces}, evolve at different speeds and thus they need to be temporally aligned. We solve this spatiotemporal registration problem using a Riemannian approach. We treat a 3D surface as a point in a shape space equipped with an elastic Riemannian metric that measures the amount of bending and stretching that the surfaces undergo. A 4D surface can then be seen as a trajectory in this space. With this formulation, the statistical analysis of 4D surfaces can be cast as the problem of analyzing trajectories embedded in a nonlinear Riemannian manifold. However, performing the spatiotemporal registration, and subsequently computing statistics, on such nonlinear spaces is not straightforward as they rely on complex nonlinear optimizations. Our core contribution is the mapping of the surfaces to the space of Square-Root Normal Fields (SRNF) where the $\ltwo$ metric is equivalent to the partial elastic metric in the space of surfaces. Thus, by solving the spatial registration in the SRNF space, the problem of analyzing 4D surfaces becomes the problem of analyzing trajectories embedded in the SRNF space, which has a Euclidean structure. In this paper, we develop the building blocks that enable such analysis. These include: \textbf{(1)} the spatiotemporal registration of arbitrarily parameterized 4D surfaces even in the presence of large elastic deformations and large variations in their execution rates; \textbf{(2)} the computation of geodesics between 4D surfaces; \textbf{(3)} the computation of statistical summaries, such as means and modes of variation, of collections of 4D surfaces; and \textbf{(4)} the synthesis of random 4D surfaces. We demonstrate the performance of the proposed framework using 4D facial surfaces and 4D human body shapes. 
\end{abstract}

}

\maketitle


\IEEEdisplaynotcompsoctitleabstractindextext

%
\IEEEpeerreviewmaketitle

\section{Introduction}
\label{sec:introduction}

\IEEEPARstart{S}{hape}, an essential property of natural and man-made 3D objects, deforms over time as a result of many internal and external factors. For instance, anatomical organs such as bones, kidneys, and subcortical structures in the brain deform due to natural growth or disease progression; human faces deform as a consequence of talking, executing facial expressions, and aging. Similarly, actions and motions such as walking, jumping, and running are the result of deformations, over time,  of the human body shape. The ability to understand and model \textbf{(1)} the typical deformation patterns of a class of 3D objects and \textbf{(2)} the variability of these deformations within and across object classes has many applications. In medical diagnosis and biological growth modeling, one is interested in measuring the intensity of pain from facial deformations~\cite{werner2019automatic}, and in distinguishing between normal growth and disease progression using the deformation over time of body shape. In computer vision and graphics, the ability to statistically model such spatiotemporal variability can be used to summarize collections of 3D animations, and synthesize and simulate animations and motions. Similar to 3D morphable face models~\cite{egger20203d}, these tools can  also be used  in a generative model for synthesizing large corpora of labeled longitudinal 3D shape data, \eg 4D faces, for training deep neural networks.

This paper proposes a novel framework for the statistical analysis of longitudinal 3D shape data composed of objects that deform over time. Each object is represented as a closed manifold surface. We refer to an object captured at different points in time, \eg a 3D human face performing a facial expression or speaking a sentence, or a 3D human body shape growing or performing actions,  as a \emph{4D (or 3D + t) surface}.  Given a set of 4D surfaces, our goal is to:
\begin{itemize}
	\item Compute the mean  deformation pattern, \ie the statistical mean 4D surface. For example, the same person can smile in different ways. Similarly, different people smile differently. The goal is to learn, based on observed longitudinal shape data, the typical smile. 
	\item Compute the directions of variation, analogous to Principal Component Analysis (PCA) for modeling 3D shape variability~\cite{allen:2003,laga2017numerical}, but here we focus on modeling variability in 4D surface collections.
	\item Characterize a population of 4D surfaces using statistical models. 
	\item Synthesize new 4D surfaces by sampling,  randomly or in a controlled fashion, from these statistical models.
\end{itemize}



\noi  We refer to these tasks as the process of constructing a 4D atlas. Achieving this goal requires solving important fundamental challenges. In fact, 3D objects such as faces, human body shapes, and  anatomical organs, which come with arbitrary parameterizations,  exhibit large elastic deformations within the same object and across different objects. This makes their spatial registration, \ie finding one-to-one correspondences between each pair of shapes, very challenging. In the case of 4D surfaces, there is an additional temporal variability due to different execution rates (speeds) of evolution within and across objects. For instance, a walking action can be executed at variable speeds even by the same person. Thus, the statistical analysis of the spatiotemporal variability in samples of 4D surfaces requires efficient spatiotemporal registration of these samples. \emph{Spatial registration} refers to the process of finding a one-to-one correspondence between two 3D surfaces of the same individual, captured at different points in time,  or of different individuals. \emph{Temporal registration} refers to the problem of finding the optimal time warping that aligns 4D surfaces, \eg walking actions, performed at different execution rates.


We treat a 4D surface as a trajectory in a high-dimensional nonlinear space. We then formulate the problem of analyzing the spatiotemporal variability of 4D surfaces as the statistical analysis of elastic trajectories, where elasticity corresponds to variations in the execution rates of the 4D surfaces. However, performing statistics on trajectories embedded in nonlinear spaces of high dimension is computationally expensive since it relies on nonlinear optimizations. Our core contribution in this paper is the mapping of the surfaces to the space of Square-Root Normal Fields (SRNF)~\cite{jermyn:2012,laga2017numerical}, which has a Euclidean structure (see Section~\ref{sec:shapespace_surfaces}---in particular, the $\ltwo$ metric in the space of SRNFs is equivalent to the partial elastic metric in the space of surfaces), meaning that the problem of analyzing 4D surfaces becomes the problem of analyzing trajectories, or curves, embedded in the Euclidean space of SRNFs. 

This paper develops the building blocks that enable such analysis.  We use these building blocks to compute statistical summaries, such as means and modes of variation of collections of 4D surfaces, and for the synthesis, either randomly or in a controlled fashion, of 4D surfaces. We demonstrate the utility and performance of the proposed framework using 4D facial surfaces from the VOCA dataset~\cite{VOCA2019}, 4D human body shapes from the Dynamic FAUST (DFAUST) dataset~\cite{dfaust:CVPR:2017}, and dressed 4D human body shapes from the CAPE dataset~\cite{ma2020learning}. Our approach is, however, general and applies to all spherically-parameterized surfaces. 
In summary, the main contributions of this article are:
\begin{itemize}
	\item We represent 4D surfaces as trajectories in the space of SRNFs, which has a Euclidean structure (Section~\ref{sec:shapespace_surfaces}). This key contribution enables the usage of standard computational tools for the analysis and modeling of 4D surfaces (Section~\ref{sec:analyzing_4D_expressions}).
	\item We propose efficient algorithms for the spatiotemporal registration of 4D surfaces and  the computation of geodesics between such 4D surfaces, even in the presence of large elastic deformations and significant variations in execution rates (Sections~\ref{sec:temporal_alignment} and~\ref{sec:4Dgeodesics}).
	\item The framework does not explicitly or implicitly assume  that the correspondences between the surfaces are given. It simultaneously solves  for the spatial and temporal registrations, and for the 4D geodesics that are optimal under the proposed  metrics.
	\item We develop computational tools for \textbf{(1)} computing summary statistics of 4D surfaces and \textbf{(2)} synthesizing  4D surfaces from formal statistical models (Section~\ref{sec:4Dstatistics}).
\end{itemize}

\noi The remainder of this paper is organized as follows. We first discuss related work in Section~\ref{sec:related_work}. Section~\ref{sec:framework} describes  the proposed mathematical framework. Section~\ref{sec:4Dstatistics} discusses its application to various statistical analysis tasks. Section~\ref{sec:results} presents the results and discusses the performance of the proposed framework. Section~\ref{sec:summary} summarizes the main findings of this paper and discusses future research directions.




\section{Related work}
\label{sec:related_work}

We classify the state-of-the-art into two categories. Methods in the first category focus on cross-sectional shape data (Section~\ref{sec:related_work_shape}). Methods in the second category focus on longitudinal shape data (Section~\ref{sec:related_work_longitudinal}).

\subsection{Statistical models of cross-sectional 3D shape data} 
\label{sec:related_work_shape}
Modeling shape variability in 2D and 3D objects has been  studied  extensively in the literature. Early methods use Principal Component Analysis (PCA) to characterize the shape space of objects. Initially introduced for the analysis of planar shapes, the active shape model of Cootes \etal~\cite{cootes1995active} has been extended to  3D faces~\cite{blanz1999morphable} and 3D human bodies~\cite{allen:2003}; see~\cite{egger20203d} for a detailed survey. These methods represent 3D objects as discrete sets of landmarks, \eg vertices, which are assumed to be in correspondence across a population of objects, and use standard Euclidean metrics for their comparison. Thus, they are limited to 3D objects that undergo small elastic deformations. 

To handle large nonlinear variations, \eg elastic deformations such as the bending and stretching observed in 3D human body shapes, Anguelov \etal~\cite{anguelov2005scape} introduced SCAPE, which represents body shape and pose-dependent shape in terms of triangle deformations instead of vertex displacements. Hasler \etal~\cite{hasler2009statistical} learn two linear models: one for pose and one for body shape. Loper \etal~\cite{loper2015smpl} introduce SMPL, a vertex-based linear model for human body shape and pose-dependent shape variation. This model has been extensively used in the literature for the analysis of the human body shape. It has also been adapted to  other types of objects such as  animals~\cite{zuffi2019three} and human body parts~\cite{pavlakos2019expressive}. While these models can capture large variations, they exhibit two fundamental limitations. \textbf{First}, they rely on separate models for pose-independent shape, pose-dependent shape, and pose. Thus, they are limited to specific classes of objects, \eg human bodies. Changing the target application, \eg to animals~\cite{zuffi2019three} or infants~\cite{hesse2018learning}, requires redefining the model. \textbf{Second},   they either assume a given registration between the surfaces of the 3D objects or solve for registration separately by matching vertices across the surfaces using an unrelated optimization criterion.
 

Recently, there has been a growing interest in analyzing variability in 3D shape collections using tools from differential and Riemannian geometry~\cite{kurtek2013landmark,kilian2007geometric,jermyn:2012,xie-iccv:2013,xie2014numerical,jermyn2017elastic,laga2017numerical}; see~\cite{laga2018survey} for a detailed survey. The work most relevant to ours is the Square-Root Normal Field (SRNF) representation introduced in~\cite{jermyn:2012}. In this work,  parameterized surfaces are compared using a partial elastic Riemannian metric defined as a weighted sum of a bending term and a stretching term. More importantly, Jermyn \etal~\cite{jermyn:2012} show that by carefully choosing the weights of these two terms, the complex partial elastic metric reduces to the $\ltwo$ metric in the space of SRNFs. Thus, by treating shapes of objects as points in the SRNF space, a straight line between two points in this space is equivalent to the geodesic (or shortest) curve in the original space of surfaces under the partial elastic metric, and represents the optimal deformation between them. As a result, one can perform statistical analysis in the SRNF space using standard vector calculus, and then map the results back to the space of surfaces (for visualization), using the approach of Laga \etal~\cite{laga2017numerical}. Another important property of SRNFs is that both registration and optimal deformation (geodesic) are computed jointly, using the same partial elastic metric. 


One of the fundamental problems in statistical shape analysis is correspondence and registration, see~\cite{laga20183d}. Past methods do not define a shape space and a metric that enable the computation of geodesics and statistics. Also,  correspondence methods that are based on the intrinsic properties of surfaces, \eg Generalized Multidimensional Scaling~\cite{bronstein2006generalized}, spectral descriptors~\cite{litman2013learning}, or functional maps (which rely on  the availability of descriptors)~\cite{ovsjanikov2012functional,ovsjanikov2016computing}, are primarily suited for surfaces that deform in an isometric manner. They also require  landmarks to resolve  symmetry ambiguities.


\subsection{Statistical models for longitudinal shape data} 
\label{sec:related_work_longitudinal}

As stated in~\cite{dfaust:CVPR:2017}, we live in a 4D world of 3D shapes in motion. With the availability of a variety of range sensing devices that can scan dynamic objects at high temporal frequency, there is a growing interest in capturing and modeling the 4D dynamics of objects~\cite{wand2007reconstruction,beeler2011high,tevs2012animation}. For instance,  Wand \etal~\cite{wand2007reconstruction} and Tevs \etal~\cite{tevs2012animation} propose methods to reconstruct the deforming geometry of time-varying point clouds.  Li \etal~\cite{li2017learning} use sequences of 4D scans to learn a statistical 3D facial model. This model, referred to as FLAME, has been later used by Cudeiro \etal~\cite{VOCA2019} to capture, learn, and synthesize 3D speaking styles. Bogo \etal~\cite{dfaust:CVPR:2017}  build a 4D human dataset by registering a 3D human template to sequences of 3D human scans performing various types of actions. These methods focus  on the 3D reconstruction of deforming objects. The literature on the statistical analysis of their spatiotemporal variability is rather limited.


Early works focused on longitudinal 2D shape data. For instance, Anirudh \etal~\cite{anirudh2015elastic} represent the contour of planar shapes that evolve as trajectories on a Grassmann manifold. They then use the Transported Square-Root Vector Fields (TSRVF) representation for their rate-invariant analysis. This approach was later extended to the analysis of the trajectories of sparse features or landmarks measured on the surface of a deforming 3D object. For instance, Akhter \etal~\cite{akhter2012bilinear} introduced a bilinear spatiotemporal basis to model the spatiotemporal variability in 4D surfaces. The approach treats surfaces as $N$ discrete landmarks and uses the $\ltwo$ metric and PCA in $\real^{4N}$ for their analysis. Thus, the approch is not suitable for highly articulated shapes that undergo large articulated and elastic motion (\eg human bodies). The approach also assumes that the landmarks are in correspondence, both spatially and temporally.


Anirudh \etal~\cite{anirudh2015elastic} and Ben Amor \etal~\cite{amor2015action} represent human body actions using dynamic skeletons. By treating each skeleton, represented by a set of landmarks, as a high-dimensional point on Kendall's shape space \cite{dryden:1998}, motions become trajectories in a high-dimensional, Euclidean space. Thus, one can use the rich literature on the statistical analysis of high-dimensional curves~\cite{srivastava2011shape} to build a framework for the statistical analysis of human motions and actions. This approach, however, has two fundamental limitations. \textbf{First}, the $\ltwo$ metric on Kendall's shape space is not suitable for large articulated motions. \textbf{Second}, skeletons and landmarks do  not capture surface elasticity, and thus, cannot be used to model growth processes and surface deformations due to motion. While this can be addressed by using two separate models, one for shape and another for motion, it will fail to capture motion-dependent shape variations.



Using the LDDMM framework~\cite{beg2005computing}, Debavelaere \etal~\cite{debavelaere2020learning} and Bone \etal~\cite{bone2020learning}  represent a 4D surface as a flow of deformations of the 3D volume around each surface and then code deformations as geodesics on a Riemannian manifold. However, in general, natural deformations do not correspond to geodesics but can be arbitrary paths on the shape space. Also, deforming 3D volumes is expensive in terms of computation and memory requirements.  Finally, this approach relies on manually specified landmarks to efficiently register the 3D volumes. Our approach, which can handle large articulated and elastic motions,  works directly on surfaces,  does not assume that deformations are (piecewise) geodesics, and does not rely on landmarks for the spatiotemporal registration. 

\section{Mathematical framework}
\label{sec:framework}

We describe in this section the proposed mathematical framework for the spatiotemporal registration and comparison of 4D surfaces.  Section~\ref{sec:4Dstatistics} discusses its application to various statistical analysis tasks. A 4D surface, where the fourth dimension refers to time, is a 3D surface that evolves over time. Examples of such 4D surfaces include facial expressions (\eg a smiling face), a human body shape performing an action such as walking or jumping, or an anatomical organ that evolves over time due to natural growth or disease progression. It can be represented as a path $\curve(t),\ t \in[0,1]$ such that $\curve(0)$ and $\curve(1)$ are the initial and final surfaces, respectively, and $\curve(t),\ 0<t<1$ are the intermediate surfaces. The main challenges posed by the statistical analysis of such 4D surfaces are two-fold. \textbf{First}, surfaces within the same 4D surface and across different 4D surfaces come with arbitrary poses and registrations. \textbf{Second}, 4D surfaces can have different execution rates, \eg two smiling expressions performed at different speeds. Thus, to compare and perform statistical analysis on samples of 4D surfaces, we first need to spatiotemporally register them. 

We solve the spatiotemporal registration problem using tools from differential geometry. We treat surfaces as points in a Riemannian shape space equipped with an elastic metric that captures shape differences using  bending and stretching energies. We then formulate the elastic registration problem, \ie the problem of computing spatial correspondences, as that of finding the optimal rotation and reparameterization that align one surface onto another. This enables comparing and spatially registering surfaces, even in the presence of large elastic deformations (Section~\ref{sec:shapespace_surfaces}).

With this representation, a 4D surface becomes a time-parameterized trajectory in the above-referenced Riemannian shape space. Thus, the problem of analyzing 4D surfaces is reduced to the problem of analyzing curves. Similar to surfaces, we define a space of curves equipped with a Riemannian metric, which quantifies the amount of elastic deformation, or time warping, needed to align two 4D surfaces (Section~\ref{sec:analyzing_4D_expressions}). 

\subsection{The elastic shape space of surfaces} 
\label{sec:shapespace_surfaces}

\begin{figure}[t]
	\includegraphics[trim={0 75 0 75},clip, width=\linewidth]{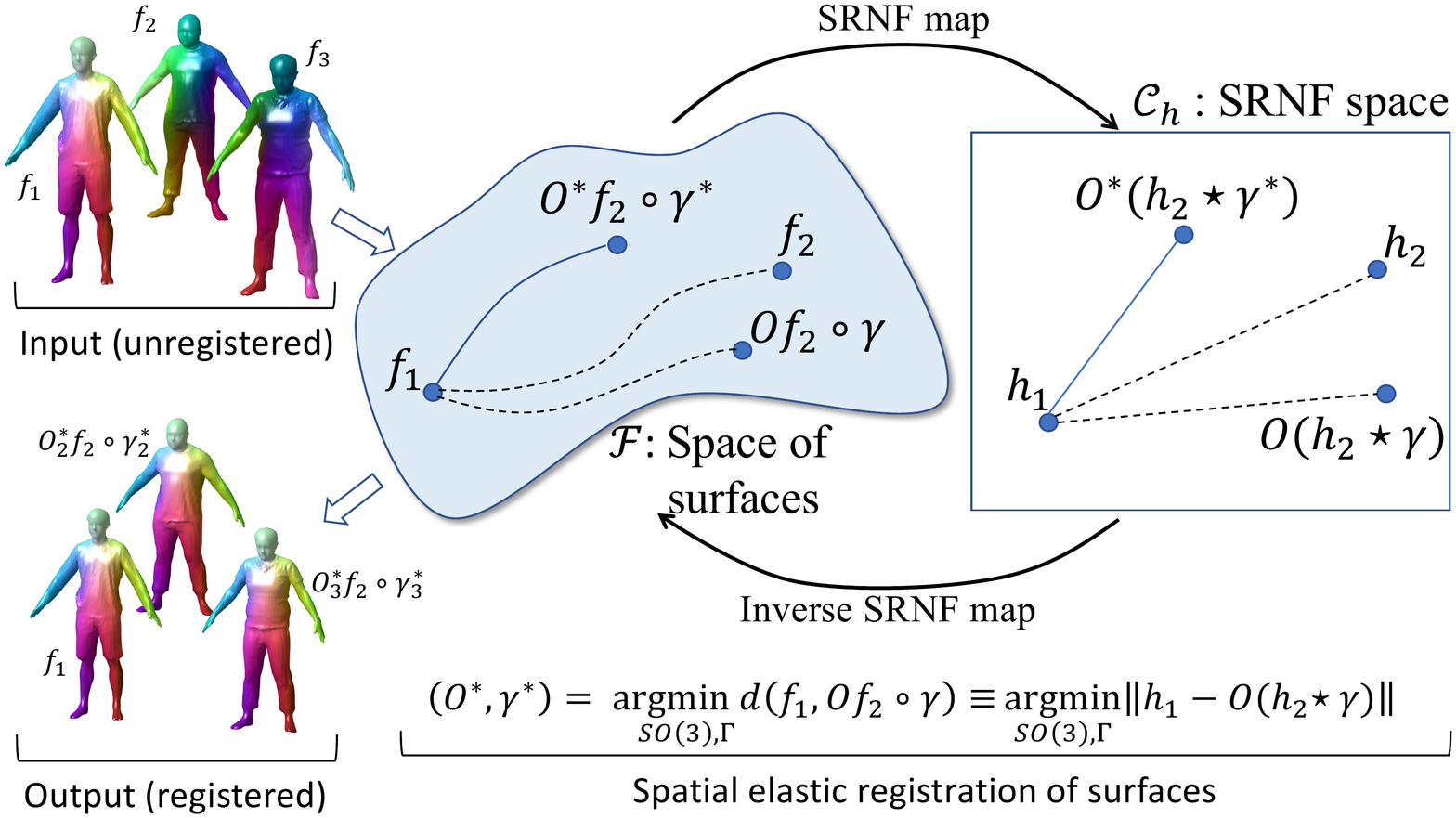}
	\caption{\label{fig:overview_spatial_registration} Overview of the proposed spatial registration framework. Surfaces are first mapped onto the space of Square-Root Normal Fields (SRNF) and spatially registered using the $\ltwo$ metric, which is equivalent to the partial elastic metric in the original space of surfaces.  4D surfaces can then be treated as curves embedded in the $\ltwo$ space of SRNFs. The operator $\star$ refers to the composition of functions in the SRNF space.} 
\end{figure}

Fig.~\ref{fig:overview_spatial_registration} overviews the proposed spatial registration framework. We consider a surface as a function $\surface$ of the form: 
\begin{eqnarray}
	\surface &:& \domain \to \rthree; \hspace{12pt} \parameter \mapsto \surface(\parameter) = (X(\parameter), Y(\parameter), Z(\parameter)), 
\end{eqnarray}

\noi where $\domain$ is a parameterization domain and $\parameter\in\domain$ is the parameter in this domain. The choice of  $\domain$  depends on the nature of the surfaces of interest. When dealing with closed surfaces of genus-0, $\domain$ is a sphere, \ie $\domain = \stwo$, and $\parameter= (u, v)$, with $u \in [0, \pi]$ and $v \in [0, 2\pi[$, are the spherical coordinates. In practice, surfaces come as unregistered triangulated meshes. We then use the spherical parameterization algorithm of~\cite{praun:2003} to map them to a spherical domain.

To remove shape-preserving transformations, we  first translate the surfaces so that their center of mass is located at the origin, and  then scale them  to have unit surface area. The space normalized surfaces,  denoted by $\surfaces$, is called the \emph{preshape space}. 


Having removed translation and scale, we still need to account for rotations and reparameterizations. Those are handled algebraically. For any surface $\surface \in \surfaces$ and for any rotation $\rotation \in \rotations$, $\rotation \surface$ and $\surface$ have equivalent shapes. Similarly, any reparameterization of a surface with an orientation-preserving diffeomorphism  preserves its shape. Let $\diffeos$ be the space of all  orientation-preserving diffeomorphisms of $\domain$. Then, $\forall\ \diffeo \in \diffeos$, $\surface$ and $\surface \circ \diffeo$, \ie the reparameterization of $\surface$ with $\diffeo$, have the same shape. (Here, $\circ$ refers to the composition of two functions.)  Note  that reparameterizations provide dense correspondences across surfaces. If one wants to put a surface $\surface_2$ in correspondence with another surface $\surface_1$, then we  need to find a rotation $\rotation^*$ and a reparameterization $\diffeo^*$ such that $\rotation^*(\surface_2 \circ \diffeo^*)$ is as close as possible to $\surface_1$. This is precisely the process of 3D surface registration. It  is defined mathematically as:
\begin{equation} 
 (\rotation^*, \diffeo^*) =\argmin_{\rotation \in SO(3), \diffeo\in \diffeos} d_{\surfaces}(f_1, \rotation (f_2 \circ \diffeo)),
 \label{eq:spatial_reg_in_surfaces}
\end{equation}
where $d_{\surfaces}$ is a measure of distances between surfaces in $\surfaces$. 

\subsubsection{SRNF representation of surfaces}

For efficient registration and comparison of surfaces, the distance measure, or metric, $d_{\surfaces}$  should quantify interpretable shape differences, \ie the amount of bending and stretching one needs to apply to one surface to deform it into another. It should also be simple enough to facilitate efficient computation of correspondences and geodesic paths. Jermyn \etal~\cite{jermyn:2012} introduced a partial elastic metric that measures differences between surfaces as a weighted sum of the amount of bending and stretching that one needs to  apply to a surface  to align it to another. In this approach, bending is measured in terms of changes in the orientation of the unit normal vectors, while stretching is measured in terms of changes in the infinitesimal surface areas. More importantly, Jermyn \etal~\cite{jermyn:2012} showed that by using a special representation of surfaces, called the Square-Root Normal Field (SRNF), the complex partial elastic metric reduces to the simple $\ltwo$ metric on the SRNF space. 

\begin{definition}[SRNF maps] The SRNF map $\srnfmap(\surface)$ of a surface $\surface \in \surfaces$ is defined as the normal vector field of the surface scaled by the square-root of the local area around each surface point: 
\small{\begin{align}
 \srnfmap:\ &\surfaces \rightarrow \srnfs \nonumber \\
   & \surface \mapsto \srnfmap(\surface) = \srnf \text{, such that } \srnf(u, v) = \frac{\normalfield(u,v)}{\| \normalfield(u,v)\|^{\frac{1}{2}}_2},
\end{align}
}
\noi where $\srnfs$ is the space of all SRNFs, $\normalfield=\frac{\partial \surface}{\partial u} \times \frac{\partial \surface}{\partial v}$ is the normal field to  $\surface$ and $\|\cdot\|_2$ is the $\ltwo$  norm in $\rthree$.
\end{definition}

\noi The SNRF representation of surfaces has nice  properties that make it suitable for the various analysis tasks at hand:
\begin{itemize}
    \item It is translation invariant. Also, the SRNF of a rotated surface is simply the rotation of the SRNF of that surface, \ie $\srnfmap(\rotation \surface) = \rotation \srnfmap(\surface)$.
    
    \item $\forall \diffeo \in \diffeos$, $\srnfmap(\surface \circ \diffeo) = \sqrt{|J_{\diffeo}|} (\srnf \circ \diffeo) \equiv \srnf \ast \diffeo$, where $J_{\diffeo}$ is the Jacobian of $\diffeo$ and $|\cdot|$ is its determinant. 

    \item Under the $\ltwo$ metric in the space of SRNFs, the action of $\Gamma$ is by isometries, \ie $\forall\ \diffeo \in \diffeos$ and $\forall\ \surface_1, \surface_2 \in \surfaces,\ \|\srnf_1 - \srnf_2 \| = \|\srnf_1 \ast \diffeo - \srnf_2 \ast \diffeo \|$, where $\srnf_i = \srnfmap(f_i),\ i=1,2$.
    
    \item The space of SRNFs is a subset of $\ltwo(\domain, \rthree)$. In addition, the $\ltwo$ metric in $\srnfs$ is equivalent to the partial elastic metric in the space of surfaces. As such, geodesics in $\surfaces$ become straight lines in the SRNF space $\srnfs$; see Fig.~\ref{fig:overview_spatial_registration}.  
    
    \item Currently, there is no analytical expression for the inverse SRNF map, and in fact, the injectivity and surjectivity of the SRNF remain open questions. However,  Laga \etal~\cite{laga2017numerical} showed that, for a given SRNF of a valid surface, one can always numerically estimate the original surface, up to translation~\cite{laga2017numerical}.
\end{itemize}

\noi The last three properties are critical for comparison and atlas construction of 4D surfaces.  One can perform elastic registration of surfaces using the standard $\ltwo$ metric in the space of SRNFs, which is computationally very efficient compared to using the complex elastic metric in the space of surfaces (Section~\ref{sec:elastic_spatial_registration}). Further, temporal evolutions of surfaces   can be interpreted as curves in the Euclidean space of SRNFs, making them amenable to statistical analysis. Thus, the problem of constructing 4D atlases becomes  the problem of statistical analysis of elastic curves in the space of SRNFs using standard statistical tools developed for Euclidean spaces. After analysis, the results can be mapped back to the original space of surfaces using efficient SRNF inversion procedures~\cite{laga2017numerical} (Section~\ref{sec:analyzing_4D_expressions}).

\subsubsection{Spatial elastic registration of surfaces}
\label{sec:elastic_spatial_registration}
Under the SRNF representation, the elastic registration problem in Eqn.~\eqref{eq:spatial_reg_in_surfaces} can be reformulated using the $\ltwo$ metric on $\srnfs$, the space of SRNFs, instead of the complex partial elastic metric on the preshape space $\surfaces$.  Let $\surface_1$ and $\surface_2$ be two surfaces in the preshape space $\surfaces$, and $\srnf_1$ and $\srnf_2$ their SRNFs. Then, the rotation and reparameterization that optimally register $\surface_2$ to $\surface_1$ are given by:
\begin{equation}
  (\rotation^*, \diffeo^*) =\argmin_{\rotation \in \rotations, \diffeo\in \diffeos} \|\srnf_1 - \rotation (\srnf_2 \ast \diffeo)\|,
 \label{eq:spatial_reg_in_srnfs}
\end{equation}

\noi where $\ast$ is the composition operator  between an SRNF and a diffeomorphism $\diffeo\in\diffeos$. This joint optimization over $SO(3)$ and $\Gamma$ can be solved by alternating, until convergence, between the two marginal optimizations (this is allowed due to the product structure of $SO(3)\times\Gamma$) \cite{kurtek:2010}:
\begin{itemize}
	\item Assuming a fixed parameterization, solve for the optimal rotation using Procrustes analysis via Singular Value Decomposition (SVD). 
	\item Assuming a fixed rotation, solve for the optimal reparameterization using a gradient descent algorithm. 
\end{itemize}

\noi To solve for the optimal reparameterization, we represent the space $\diffeos$  of diffeomorphisms $\diffeo$, which are functions on the sphere,  using spherical harmonic basis $\{ B_i\}_{ i=1, \dots, n}$. This way, every $\diffeo\in\diffeos$ can be written as a weighted sum of the harmonic basis: $\diffeo = \sum_{i=1}^{n}a_iB_i$. Thus, the search for the optimal diffeomorphism is reduced to the search for the optimal weights $\{a_i\}$. This procedure is described in detail in Section~\ref{sec:supp_spatial_registration_algorithm} of the Supplementary Material.

Although this  approach converges to a local optimum, in practice, it can be used in a very efficient way. Since a 4D surface $\curve$ is a sequence of discrete realizations $\curve(t_i) \in \surfaces,\ i=0, \cdots, \nsurfaces$, with $t_0 =0$ and $t_{\nsurfaces} = 1$, one can perform the elastic registration sequentially. Let $\srnfcurve = \srnfmap(\curve)$ be the SRNF map of the 4D surface $\curve$, \ie $\forall\ t \in [0, 1], \ \srnfcurve(t) = \srnfmap(\curve(t))$. Also, let $\curve_0$ be a reference surface randomly chosen from the population of surfaces being analyzed, and $\srnfcurve_0$ its SRNF map ($\curve_0$ can be, for example, $\curve(0)$). Then,
\begin{enumerate}
	\item Find $\rotation_0 \in \rotations $ and $\diffeo_0 \in \diffeos$ that register $\srnfcurve(t_0)$ (the start point of SRNF path) to the SRNF of the reference surface $\srnfcurve_0$,  by solving Eqn.~\eqref{eq:spatial_reg_in_srnfs}. 
	\item For $i=0, \dots, \nsurfaces$, 
		\begin{itemize}
			\item $\srnfcurve(t_i) \leftarrow \rotation_0 (\srnfcurve(t_i) \ast \diffeo_0$) and $\curve(t_i) \leftarrow \rotation_0 \curve(t_i) \circ \diffeo_0$.
		\end{itemize}
	\item For $i=1, \dots, \nsurfaces$, 
		\begin{itemize}
			\item Find, by solving Eqn.~\eqref{eq:spatial_reg_in_srnfs},  $\rotation_i \in \rotations$ and $\diffeo_i \in \diffeos$ that register $\srnfcurve(t_i)$ to $\srnfcurve(t_{i-1})$. 
			\item $\srnfcurve(t_i) \leftarrow \rotation_i (\srnfcurve(t_i) \ast \diffeo_i$) and $\curve(t_i) \leftarrow \rotation_i \curve(t_i) \circ \diffeo_i$.
		\end{itemize}
\end{enumerate}

\noi The first step ensures that, when given a collection of 4D surfaces $\curve_j, \ j=1, \cdots, n$, the surfaces $\curve_j(0), \ j=1, \cdots, n$  are registered to each other. The subsequent steps ensure that $\forall t, \ \curve_j(t)$ is registered to $\curve_j(0)$. This sequential approach is efficient since, in general, elastic deformations between two consecutive frames in a 4D surface are relatively small. In what follows, we assume that all surfaces within a 4D surface and across 4D surfaces are correctly registered, \ie they have been normalized for translation and scale, and optimally rotated and reparameterized using the approach described in this section. 

\subsection{The shape space of 4D surfaces} 
\label{sec:analyzing_4D_expressions}
\begin{figure}[t]
	\includegraphics[trim={0 30 0 30},clip, width=\linewidth]{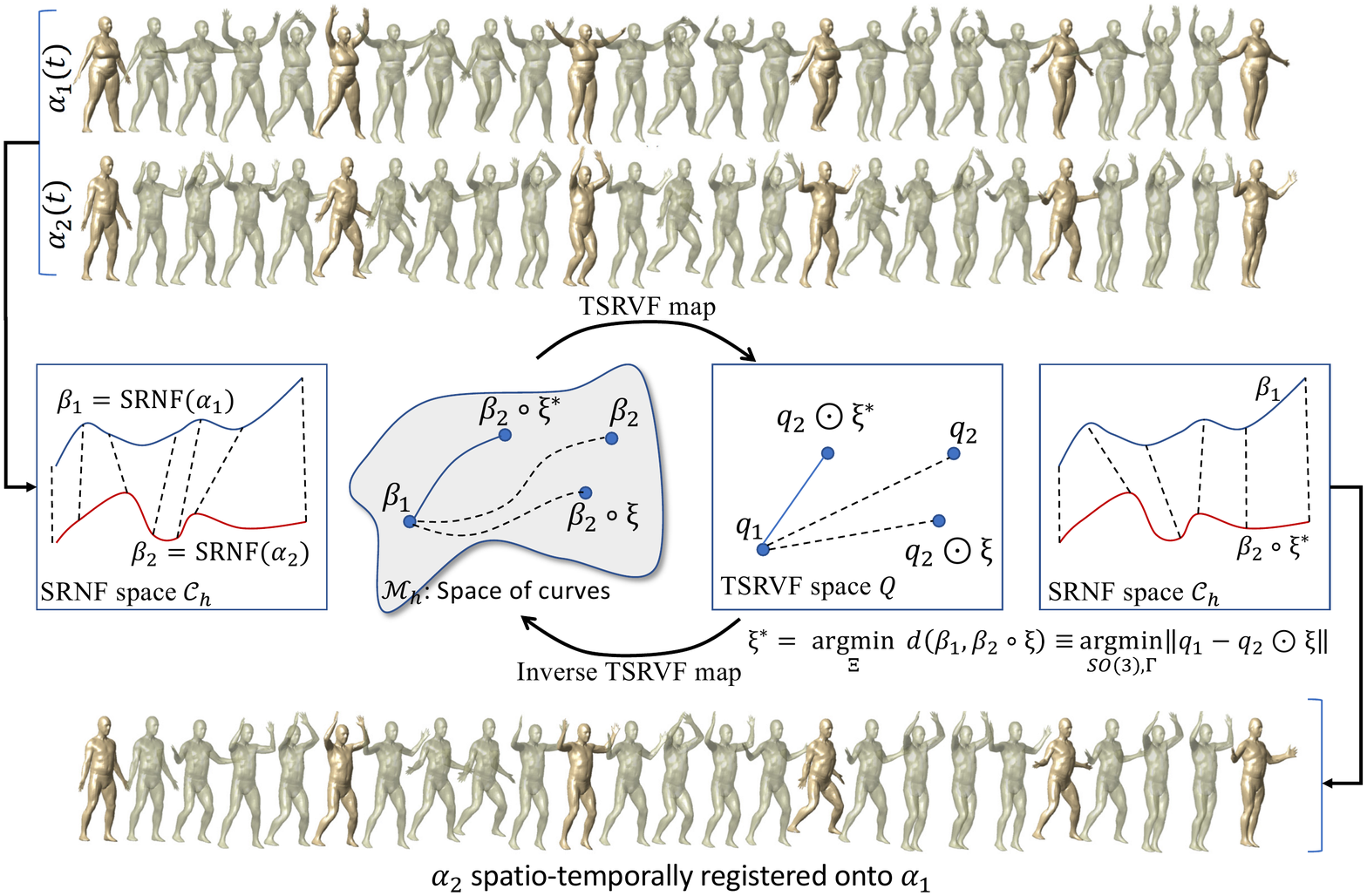}
 	\caption{\label{fig:overview_temporal_registration} In the proposed temporal registration framework, 4D surfaces, represented as curves in the SRNF space,  are first mapped to the space of Transported Square-Root Vector Fields (TSRVF) for their temporal registration. Points in the TSRVF space are  mapped back to the space of SRNFs and then  to the original space of surfaces for visualization. The operator  $\odot$ refers to the composition of functions  in the TSRVF space. 
}
\end{figure}

Under the setup of Section \ref{sec:shapespace_surfaces}, a 4D surface becomes a curve $\curve: [0, 1] \to \surfaces$. However, since $\surfaces$ is endowed with the partial elastic metric, which is non-Euclidean, we propose to further map the 4D surfaces to the SRNF space, which has a Euclidean structure. Thus, 4D surfaces become curves of the form $\srnfcurve: [0, 1] \to \srnfs$. With this representation, all statistical tasks are carried out in $\srnfs$ under the $\ltwo$ metric with results mapped back to the space of surfaces $\surfaces$ for visualization.

\subsubsection{TSRVF representation of SRNF trajectories} 

Let $\curve$ be a curve (path) in $\surfaces$ and $\srnfcurve$ its image under the SRNF map, \ie $\forall\ t \in [0, 1], \ \srnfcurve(t) = \srnfmap(\alpha(t))$; $\srnfcurve$ is also a curve, but in $\srnfs$. Let $\curves$ be the space of all paths in $\surfaces$, and $\srnfcurves$ be the space of all paths in $\srnfs$: $\srnfcurves = \{\srnfcurve: [0, 1] \to \srnfs | \srnfcurve = \srnfmap(\curve),\ \curve \in \curves\}$. 

To temporally register, compare, and summarize samples of such curves, we need to define an appropriate metric on $\curves$, or $\srnfcurves$, that is invariant to the rate (or speed) of the 4D surfaces. For example,  facial expressions that only differ in the rate of their execution should be deemed equivalent under such a metric. Let $\curvediffeos=\{\curvediffeo: [0, 1] \to [0, 1] \text{ such that } 0<\dot{\curvediffeo}<\infty, \curvediffeo(0) = 0 \text{ and } \curvediffeo(1) = 1 \}$ denote all reparameterizations of the temporal domain $[0, 1]$. Here, $\dot{\curvediffeo}=\frac{d\curvediffeo}{dt}$. Then, for any $\curvediffeo\in\curvediffeos$, $\srnfcurve \circ \curvediffeo$ and $\srnfcurve$ only differ in the rate of execution and are thus equivalent. The function $\curvediffeo$ is often referred to as a time warping of the domain $[0,1]$. Temporal registration of two 4D surfaces $\curve_1$ and $\curve_2$ then becomes the problem of registering their corresponding curves $\srnfcurve_1$ and $\srnfcurve_2$ in $\srnfs$. This requires solving for an optimal reparameterization $\curvediffeo^* \in \curvediffeos$ that minimizes an appropriate distance $d(\cdot, \cdot)$ between $\srnfcurve_1$ and $\srnfcurve_2$:
\begin{equation}
 \curvediffeo^* = \argmin_{\curvediffeo \in \curvediffeos} d(\srnfcurve_1, \srnfcurve_2 \circ \curvediffeo).
 \label{eq:general_registration_paths}
\end{equation}

\noi The optimization over $\curvediffeos$ in Eqn.~\eqref{eq:general_registration_paths} ensures rate invariance. Thus, we are left with defining a distance $d(\cdot, \cdot)$ that is invariant to time warping of the temporal domain $[0,1]$. To this end, we borrow tools from Srivastava \etal~\cite{srivastava2011shape} for analyzing shapes of curves in $\rnth, n\ge 2$. The associated elastic metric defined therein is invariant to reparameterizations of curves, and quantifies the amount of bending and stretching of the curves in terms of changes in the orientations and lengths of their tangent vectors, respectively. However, instead of directly working with such a complex elastic metric, Su \etal~\cite{su2014statistical} introduced the Transported Square-Root Vector Field (TSRVF) representation, which simplifies the complex elastic metric into the simple  $\ltwo$ metric. 

\begin{definition}[Transported Square-Root Vector Field (TSRVF)]
For any smooth trajectory $\srnfcurve \in \srnfcurves$, the transported square-root vector field (TSRVF) is a parallel transport of a scaled velocity vector field of $\srnfcurve$ to a reference point $c \in \srnfs$ according to
 \begin{equation}
   \srvfmap(\srnfcurve)(t) = \srvf(t) = \frac{\dot{\srnfcurve}(t)|_{\srnfcurve(t)\to c}}{ \sqrt{\|\dot{\srnfcurve}(t) \|}},
   \label{eq:tsrvf}
 \end{equation}
where $\dot{\srnfcurve} = \frac{\partial \srnfcurve}{\partial t}$ is the tangent vector field on $\srnfcurve$ and $\|\cdot\|$ is the $\ltwo$ metric on $\srnfs$.
\end{definition}

\noi Note that the parallel transport $\dot{\srnfcurve}(t)|_{\srnfcurve(t)\to c}$ is performed along the geodesic from $\srnfcurve(t)$ to $c$. The TSRVF representation has nice  properties that facilitate efficient temporal registration of 4D surfaces. Let $\srnfcurve_1$ and $\srnfcurve_2$ be two trajectories on $\srnfcurves$, and let $\srvf_1$ and $\srvf_2$ be their respective TSRVFs. Then,
\begin{itemize}
  \item The elastic metric on the space of trajectories $\srnfcurves$ reduces to the $\ltwo$ metric on the space of their TSRVFs. Thus, one can use the $\ltwo$ metric to compare two paths:
  \begin{equation}\label{eq:TSRVFltwo}
  d(\srnfcurve_1, \srnfcurve_2) = \|\srvf_1 - \srvf_2\| = \left(\int_{0}^{1} \|\srvf_1(t) - \srvf_2(t) \|^2 dt\right)^{\frac{1}{2}}, 
  \end{equation}
  where $\|\cdot\|$ is again the $\ltwo$ norm on $\srnfs$.
  
  \item For any $\curvediffeo \in \curvediffeos$, $\srvfmap (\srnfcurve \circ \curvediffeo) = ( \srvf \circ \curvediffeo) \sqrt{\dot{\curvediffeo}(t)}$, which we denote by  $ \srvf \odot \curvediffeo$.  
  
  \item Under the $\ltwo$ metric, the action of the reparameterization group $\curvediffeos$ on the space of TSRVFs is by isometries, \ie 
  $ \|\srvf_1 - \srvf_2 \| = \|(\srvf_1 \odot \curvediffeo) - (\srvf_2 \odot \curvediffeo)\|,\ \forall\ \curvediffeo \in \curvediffeos$.
  
  \item Given a TSRVF $\srvf$ and an initial trajectory point, one can reconstruct the corresponding path $\srnfcurve$, such that $\srvfmap(\srnfcurve) = \srvf$, by solving an ordinary differential equation~\cite{su2014statistical}. 
\end{itemize}

\noi As we will see next, these properties enable efficient temporal registration of trajectories and subsequent rate-invariant statistical analysis. In what follows, let $\srvfs$ denote the space of TSRVFs equipped with the $\ltwo$ metric defined in Eqn.~\ref{eq:TSRVFltwo}.


%

\subsubsection{Temporal registration}
\label{sec:temporal_alignment}
%

Under the TSRVF representation, the temporal registration problem in  Eqn.~\eqref{eq:general_registration_paths}, which involved optimization over $\curvediffeos$, can now be reformulated using the standard $\ltwo$ metric on the space of TSRVFs:
\begin{equation}
 \curvediffeo^* = \argmin_{\curvediffeo\in \curvediffeos} \|\srvf_1 - \srvf_2 \odot \curvediffeo\|.
 \label{eq:optimalTemporalRegistration}
\end{equation}

\noi This problem can be solved efficiently using a Dynamic Programming algorithm \cite{robinson2012functional,su2014statistical}. Then, the rate-invariant distance between two trajectories is given by:
\begin{equation}
 d(\srnfcurve_1, \srnfcurve_2) = \inf_{\curvediffeo\in \curvediffeos} \|\srvf_1 -  \srvf_2 \odot \curvediffeo\|.
 \label{eq:optimalTemporalRegistration_distance}
\end{equation}

%


\begin{figure*}[t]
\center{
	\includegraphics[trim={0 0 0 0},clip, width=\textwidth]{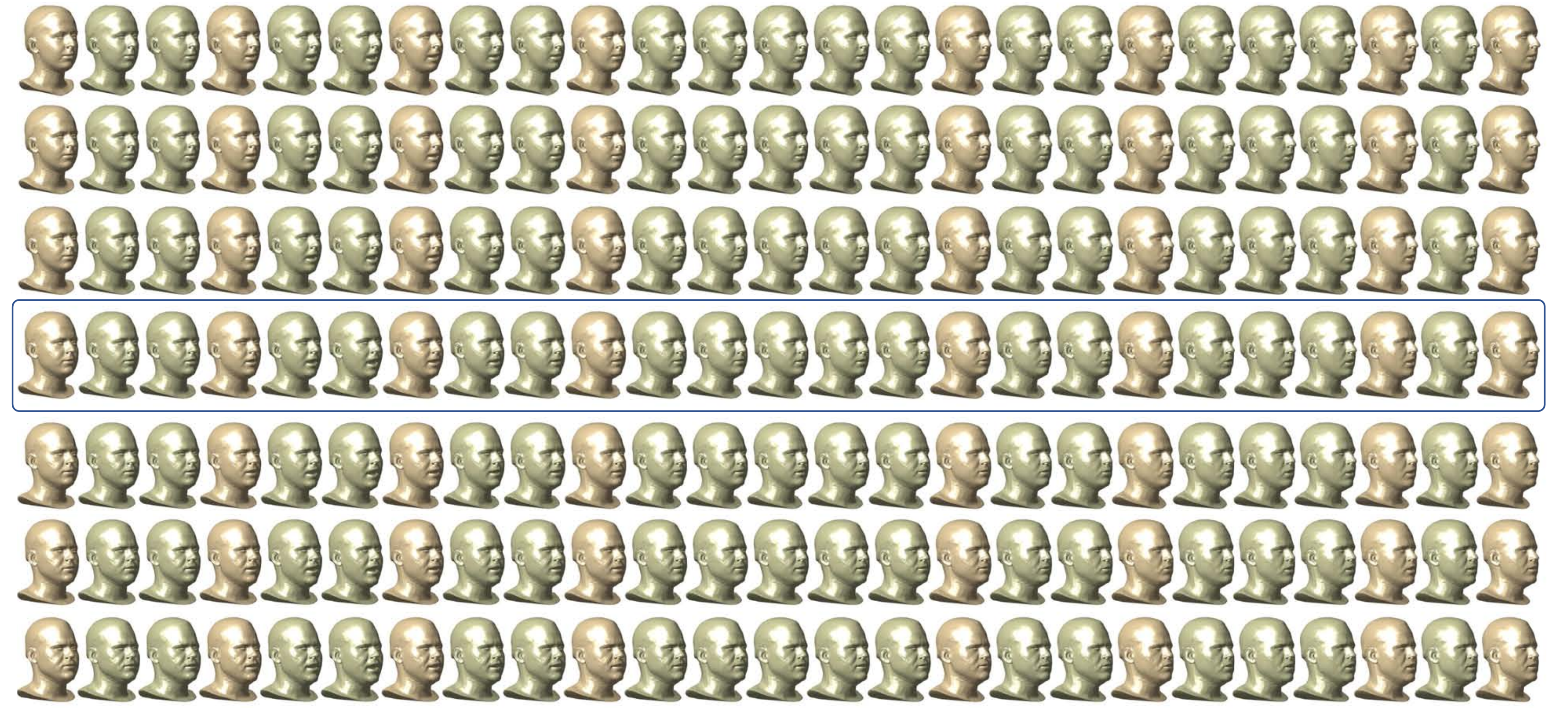} 
	\caption{\label{fig:4Dgedeosics_faces1} Example of a geodesic between the source 4D surface (top row) and the target 4D surface (bottom row) after spatiotemporal registration. The highlighted row corresponds to the mean 4D surface.  A video of the figure is included in the Supplementary Material. }
}
\end{figure*}

\subsubsection{Geodesics between 4D surfaces}
\label{sec:4Dgeodesics}

We now summarize our  entire pipeline. Let $\curveone,\curvetwo \in \curves$ be two 4D surfaces. The pipeline to spatiotemporally register them and compute the geodesic path between them can be summarized as follows.


\vspace{6pt}
\noi\textbf{(1) Proposed spatial registration. } The goal is to spatially register the surfaces in $\curveone$ and $\curvetwo$ to the same reference surface, which can be  any arbitrary surface. For simplicity, we choose it to be $\curveone(0)$, the first surface in the sequence $\curveone$. The spatial registration can then be performed in two steps.

\begin{itemize}
	\item Compute the SRNF maps, $\forall\ t\in [0, 1]$, \ $\srnfcurveone(t) = \srnfmap(\curveone(t))$ and $\srnfcurvetwo(t) = \srnfmap(\curvetwo(t))$. 
	
	
	\item Spatially register $\srnfcurveone$ and $\srnfcurvetwo$, and thus $\curveone$ and $\curvetwo$, to the reference surface, using the algorithm described in Section~\ref{sec:elastic_spatial_registration}. 
	
\end{itemize}

\noi For simplicity of notation, we also use $\srnfcurveone$ and $\srnfcurvetwo$ to denote the spatially-registered trajectories.

\vspace{6pt}
\noi\textbf{(2) Proposed temporal alignment.} $\srnfcurveone$ and $\srnfcurvetwo$ are elements of $\srnfcurves$. We perform temporal registration in three steps. 
		\begin{itemize}
			\item Map $\srnfcurveone$ and $\srnfcurvetwo$ to the TSRVF space $\srvfs: \srvfone = \srvfmap(\srnfcurveone)$ and $\srvftwo = \srvfmap(\srnfcurvetwo)$. 
			\item Find $\curvediffeo^*$, the optimal reparametrization that registers $\srvftwo$ to $\srvfone$ by solving Eqn.~\eqref{eq:optimalTemporalRegistration}.
			\item $\srvftwo^* \leftarrow \srvftwo \odot \curvediffeo^*$ and $\srnfcurvetwo^* \leftarrow \srnfcurvetwo \circ \curvediffeo^*$.
		\end{itemize}
		
\vspace{6pt}
\noi\textbf{(3) Proposed geodesic computation. } Since $\srvfs$ is Euclidean, the geodesic path $\curvebetweensrvfanim$ between $\srvfone$ and $\srvftwo^*$ is a straight line:
			 \begin{equation} 
			 	\curvebetweensrvfanim(\tau) = (1-\tau)\srvfone + \tau \srvftwo^*,\  \tau\in[0, 1].
			 \end{equation}
			
\noi Next, we map $\curvebetweensrvfanim$ back to $\srnfcurves$ using the inverse TSRVF map, \ie $\forall\ \tau,\ \curvebetweensrnfanim(\tau) = \srvfmap^{-1}(\curvebetweensrvfanim(\tau))$. The computation of the inverse mapping uses the starting point on the trajectory and has a closed-form solution, making it computationally efficient. This is described in detail in \cite{su2014statistical}. After applying the inverse mapping to the entire geodesic path, we have $\curvebetweensrnfanim(0) = \srnfcurveone,\  \curvebetweensrnfanim(1) = \srnfcurvetwo$, and $ \srnfcurve_{\tau} = \curvebetweensrnfanim(\tau),\ \tau \in (0, 1)$, \ie a geodesic path between the SRNF curves $\srnfcurveone$ and $\srnfcurvetwo$. 


\vspace{6pt}
\noi\textbf{(4) Visualization. } To visualize geodesic paths between 4D surfaces (and not their SRNFs), we need to further map all SRNFs on the trajectory $\curvebetweensrnfanim(\tau)$ to their corresponding surfaces in $\surfaces$. This is done using the inverse SRNF map, \ie $
	 	\forall\ \tau\in [0,1],\ t \in [0, 1], \ \curvebetweenanim(\tau)(t) = \srnfmap^{-1}(\curvebetweensrnfanim(\tau)(t)).
	$ Unlike the TSRVF map whose inverse can be computed analytically, inversion of the SRNF map, whose injectivity and surjectivity are yet to be determined, has to be accomplished numerically using the approach of Laga \etal~\cite{laga2017numerical}.
	
\begin{figure*}[!th]
\center{
	\begin{tabular}{@{}c@{}}
		\midrule
		\includegraphics[trim={0 0 0 0},clip, width=\textwidth]{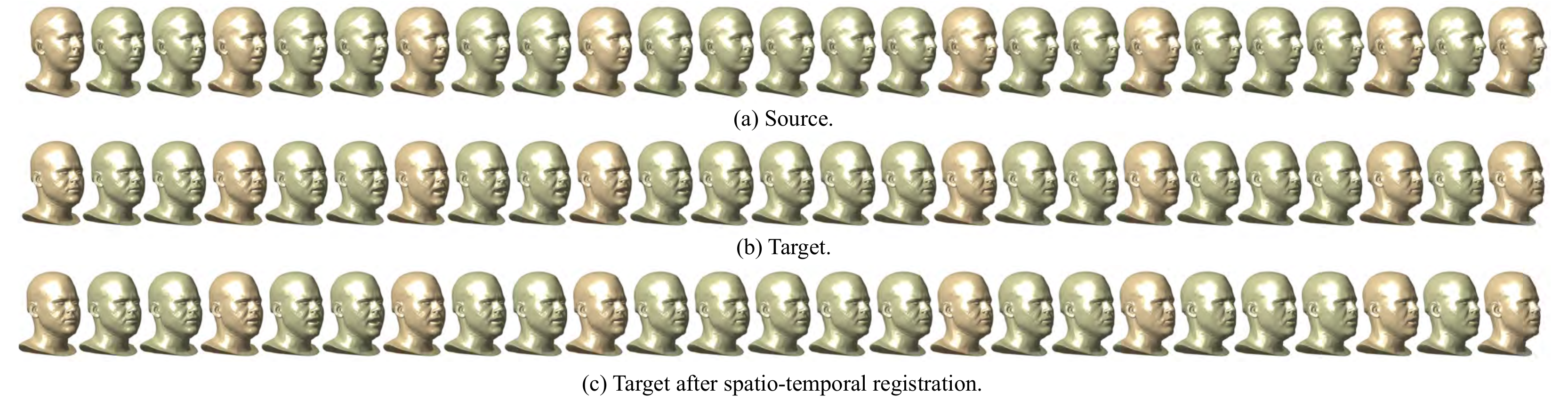} \\
		\midrule
		\includegraphics[trim={0 0 0 0},clip, width=\textwidth]{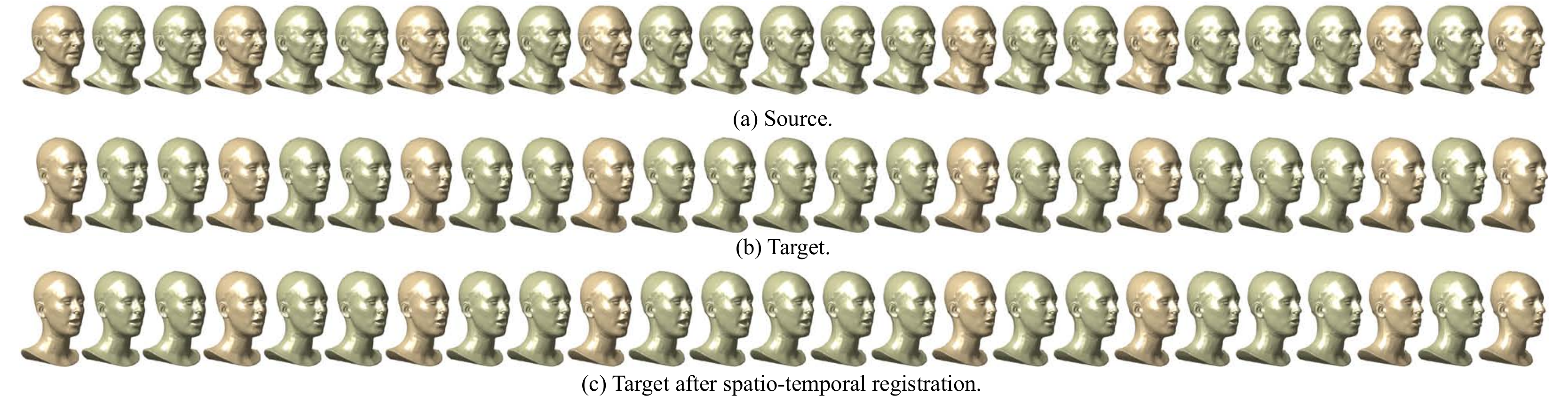}  \\
		\midrule
	\end{tabular}
		 \caption{\label{fig:faces_spatio_temporal1} \label{fig:faces_spatio_temporal2} Examples of the spatiotemporal registration of two facial expressions (4D faces).  In each example, we show \textbf{(a)} the source 4D face,  \textbf{(b)} the target 4D face, and  \textbf{(c)} the target 4D face after spatiotemporal registration using the proposed framework.  Note how the spatiotemporally registered target 4D surface became fully synchronised with the source 4D surface.   The full video sequence is provided in the supplementary material.} 
		 }
\end{figure*}

Now, $\curvebetweenanim$ is the geodesic path between the 4D surfaces $\curveone$ and $\curvetwo^{*}$, \ie $\curvebetweenanim(0) = \curveone, \curvebetweenanim(1) = \curvetwo^*$, and $\curve_\tau = \curvebetweenanim(\tau)$ is a 4D surface at time $\tau$ along the geodesic path. Fig.~\ref{fig:4Dgedeosics_faces1} shows an example of a geodesic between two 4D surfaces representing talking faces. Each row corresponds to one 4D surface. The top is the source, the bottom row is the target, after optimal spatiotemporal registration, and the highlighted row in the middle corresponds to the mean 4D surface.  The temporal registration is further illustrated in  Fig.~\ref{fig:faces_spatio_temporal1}, where we show the source 4D surface, the target 4D surface before the spatiotemporal registration, and the target   4D surface after the spatiotemporal registration. Section~\ref{sec:results} provides more examples of geodesics computed between various types of 4D surfaces.

\section{Statistical analysis of 4D surfaces}
\label{sec:4Dstatistics}

Now that we have devised all of the required mathematical tools for comparing 4D surfaces, we shift our focus to how these tools can be used to build a 4D atlas from a sample of 4D surfaces. Let $\curve_1, \cdots, \curve_n$ be a set of 4D surfaces and $\srnfcurve_1, \cdots, \srnfcurve_n$ be their corresponding trajectories in $\srnfs$. We assume that all of the surfaces, and their corresponding SRNFs, have  been spatially registered to a common reference; see Section \ref{sec:elastic_spatial_registration}. We proceed to map all of the 4D surfaces to their corresponding TSRVFs, hereinafter denoted by $\srvf_1, \cdots, \srvf_n$, and compute statistics in the space of TSRVFs. As before, all results are mapped at the end to the original space of surfaces $\surfaces$ for visualization. We will  use this framework to compute means and modes of variation, and  to synthesize novel 4D surfaces by sampling from probability distributions fitted to a set of exemplar 4D surfaces.

\vspace{6 pt}
\noi\textbf{Mean of 4D surfaces. } Intuitively, the mean of a collection of 4D surfaces is the 4D surface that is as close as possible to all of the 4D surfaces in the collection, under the specified distance measure (or metric). It is  also called  Karcher mean and is defined as the 4D surface that minimizes the sum of squared distances to all of the 4D surfaces in the given sample. In other words, we seek to solve the following optimization problem, defined in the space of TSRVFs:
\begin{equation}
	\meansrvf = \argmin_{\srvf \in \srvfs} \sum_{i=1}^{n} \min_{\curvediffeo_i\in\curvediffeos} \|\srvf - \srvf_i \odot \curvediffeo_i \|^2.
	\label{eq:karchermean}
\end{equation}

\noi Algorithm~\ref{alg:karcher_mean} in the Supplementary Material describes the proposed procedure for solving this optimisation problem. It outputs the TSRVF Karcher mean $\meansrvf$, the optimal temporal reparameterizations $\curvediffeo_i^*,\ i=1,\ldots,n$, and the temporally registered TSRVFs $q_i^*=q_i \odot \curvediffeo_i^*$; again, for simplified notation we simply use $\curvediffeo_i$ and $q_i$ to denote the optimal temporal reparameterizations and the temporally registered TSRVFs. The mean 4D surface can be obtained by TSRVF inversion of the mean TSRVF   followed by SRNF inversion~\cite{laga2017numerical}.

\vspace{6pt}
\noi\textbf{Principal directions of variation.} Since the TSRVF space is Euclidean, the principal directions of variation can also be computed in a standard way, \ie using the Singular Value Decomposition (SVD) of the covariance matrix. In the following, we assume that the TSRVFs are sampled using a finite set of points and appropriately vectorized. Let $ \cov = \frac{1}{n-1} (\srvf_i - \meansrvf) (\srvf_i - \meansrvf)^{\top}$ be the covariance matrix of the input sample, $\eigenval_i,\ i=1, \dots, k$ its $k-$leading eigenvalues, and $\eigenvect_i,\ i=1, \dots, k$ the corresponding eigenvectors. Then, one can explore the variability in the $i-$th principal direction using $\srvf_{\tau}= \meansrvf + \tau \sqrt{\eigenval_i} \eigenvect_i$, where $\tau \in \real$. To visualize this principal direction of variation, we again use TSRVF inversion followed by SRNF inversion to compute the 4D surface $\curve_{\tau}$, such that $\srvfmap(\srnfmap( \curve_{\tau})) = \meansrvf + \tau \sqrt{\eigenval_i} \eigenvect_i,\ \tau \in \real$.

\vspace{6pt}
\noi\textbf{Random 4D surface synthesis. } Given the mean and the $k$-leading principal directions of variation, any TSRVF $\srvf$ of a 4D surface $\curve$ can be approximately represented, in a parameterized form, as:
\begin{equation}
	 \srvf = \meansrvf + \sum_{i=1}^{k}\tau_i \sqrt{\eigenval_i} \eigenvect_i, \tau_i \in \real.
	 \label{eq:new_srnf}
\end{equation}
\noi Thus, to generate a random TSRVF, we only need to generate $k$ random values $\tau_i \in \real$ and plug them into Eqn.~\eqref{eq:new_srnf}. Then, to compute the corresponding random 4D surface, we apply the inverse TSRVF map followed by the inverse SRNF map. Also, by enforcing each $\tau_i$ to be within a certain range, \eg $[-1, 1]$, we can ensure that the generated random 4D surfaces are similar to the given sample and thus plausible. 

This procedure allows the generation of new random 4D surfaces. However, it does not offer any control over the generation process, which is entirely random. In many situations, we would like to control this process using a set of parameters. For instance, when dealing with 4D facial expressions, these parameters can be the degree of sadness, facial dimensions, etc. This type of control can be implemented using regression in the TSRVF space, a problem that we plan to explore in the future.


\begin{figure*}[!h]
\center{
	\includegraphics[trim={30 0 30 0}, width=\textwidth]{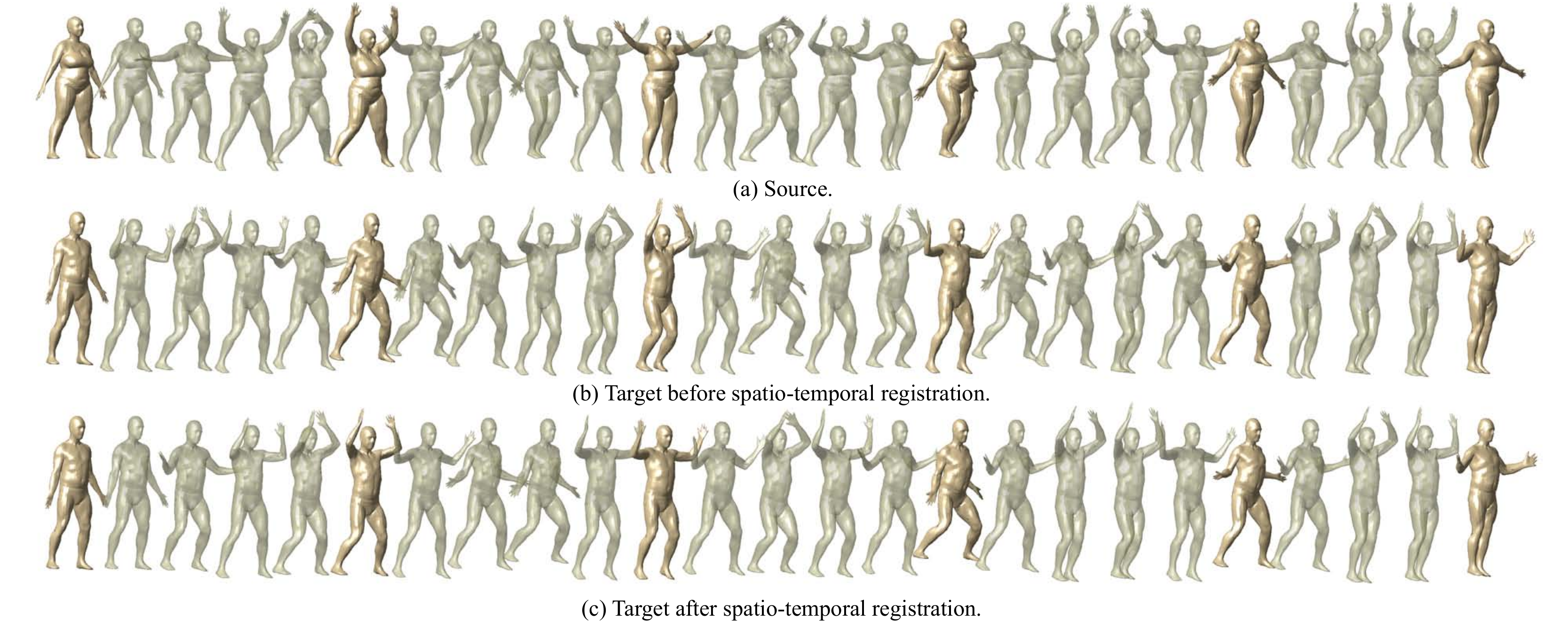}
	\caption{\label{fig:humans_spatio_temporal1} Example of the spatiotemporal registration, using the proposed algorithm, of two 4D human body shapes (from the DFAUST dataset) performing a jumping action at different speeds. Note how the spatiotemporally registered target 4D surface in (c) became synchronised with the source 4D surface in (a).   The full video sequence is provided in the supplementary material. } 
	}
\end{figure*}

\section{Results}
\label{sec:results}
This section demonstrates some results of the proposed framework and evaluates its performance. Section~\ref{sec:results_registration} focuses on spatiotemporal registration and geodesic computation between 4D surfaces. Section~\ref{sec:results_statistics} focuses on the computation of the  statistical summaries while   Section~\ref{sec:results_synthesis} focuses on the random synthesis of 4D surfaces. Finally, Section~\ref{sec:ablation_study} provides an ablation study to demonstrate the importance of each component of the proposed framework. We use three datasets:  \textbf{(1)}  VOCA~\cite{VOCA2019}, which contains 4D facial scans, captured at $60$fps, of $12$ subjects speaking various sentences; \textbf{(2)}  MPI DFAUST~\cite{dfaust:CVPR:2017}, which contains high-resolution 4D scans of $10$ human subjects in motion, captured at $60$fps; in total, the dataset contains $129$ dynamic performances; and \textbf{(3)}  MPI 4D CAPE~\cite{ma2020learning}, which contains  high-resolution 4D scans of $10$ male and $5$ female subjects in clothing. These datasets come as polygonal meshes with consistent triangulation and given registration across the meshes. We spherically parameterize them using Kurtek \etal's implementation~\cite{kurtek2013landmark}  of the spherical parameterization approach of~\cite{praun:2003}. We also apply randomly generated spatial diffeomorphisms to simulate non-registered surfaces. Our framework does not use, either explicitly or implicitly, the provided vertex-wise correspondences.

\begin{figure*}[t]
	\center{
	\includegraphics[trim={30 0 30 0}, width=\textwidth]{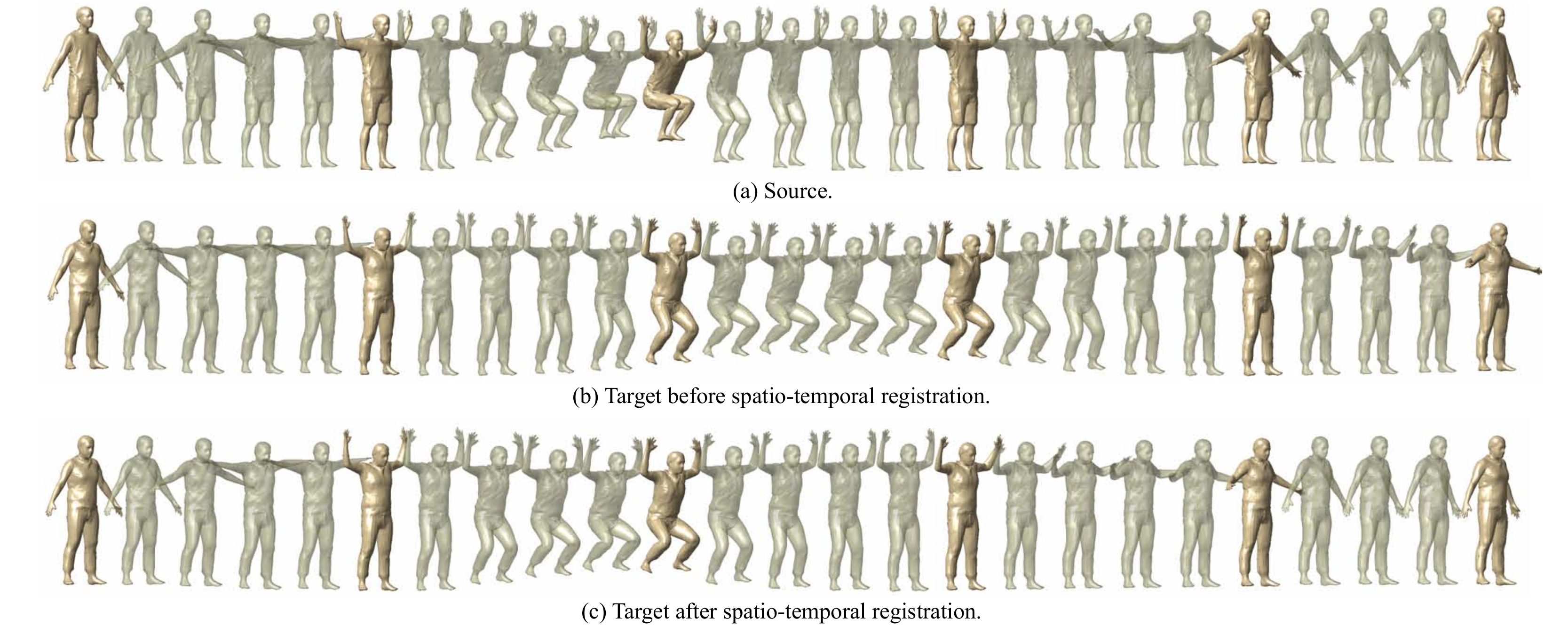} 
	\caption{\label{fig:humans_spatio_temporal2} Example of the spatiotemporal registration,  using the proposed algorithm,  of two 4D body shapes with different clothing (	from the CAPE dataset). Note how the spatiotemporally registered target 4D surface in (c) became fully synchronised with the source 4D surface in (a).   The full video sequence is provided in the supplementary material. } 
	}
\end{figure*}

\begin{figure*}[!ht]
\center{
\begin{tabular}{@{}cc@{}}
	\midrule
	\includegraphics[trim={20 0 20 0},clip, width=\textwidth]{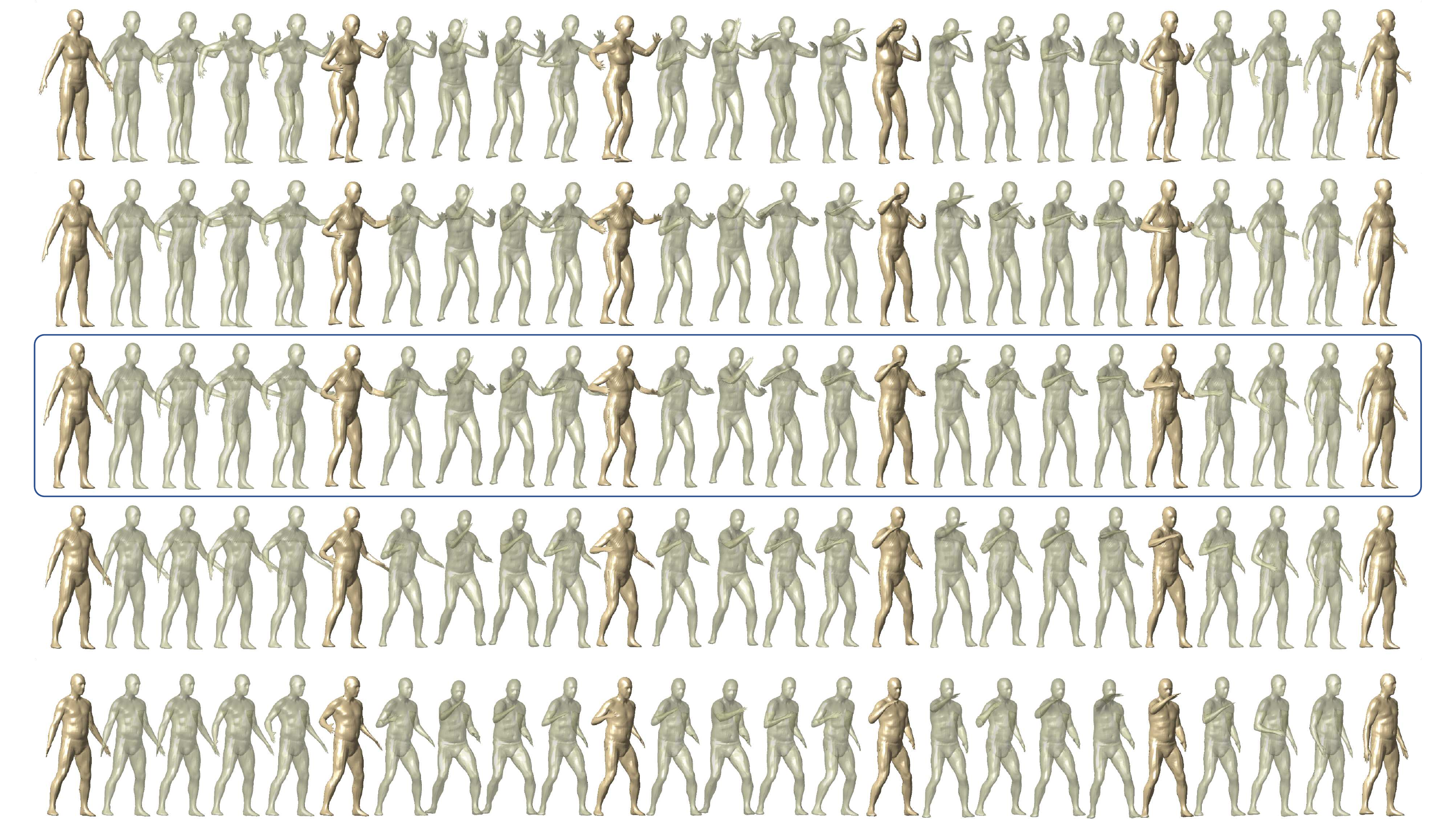}\\ 
	\small{(a) Before registration. The highlighted middle row corresponds to the mean 4D surface.} \\
	\midrule
	\includegraphics[trim={20 0 20 0},clip, width=\textwidth]{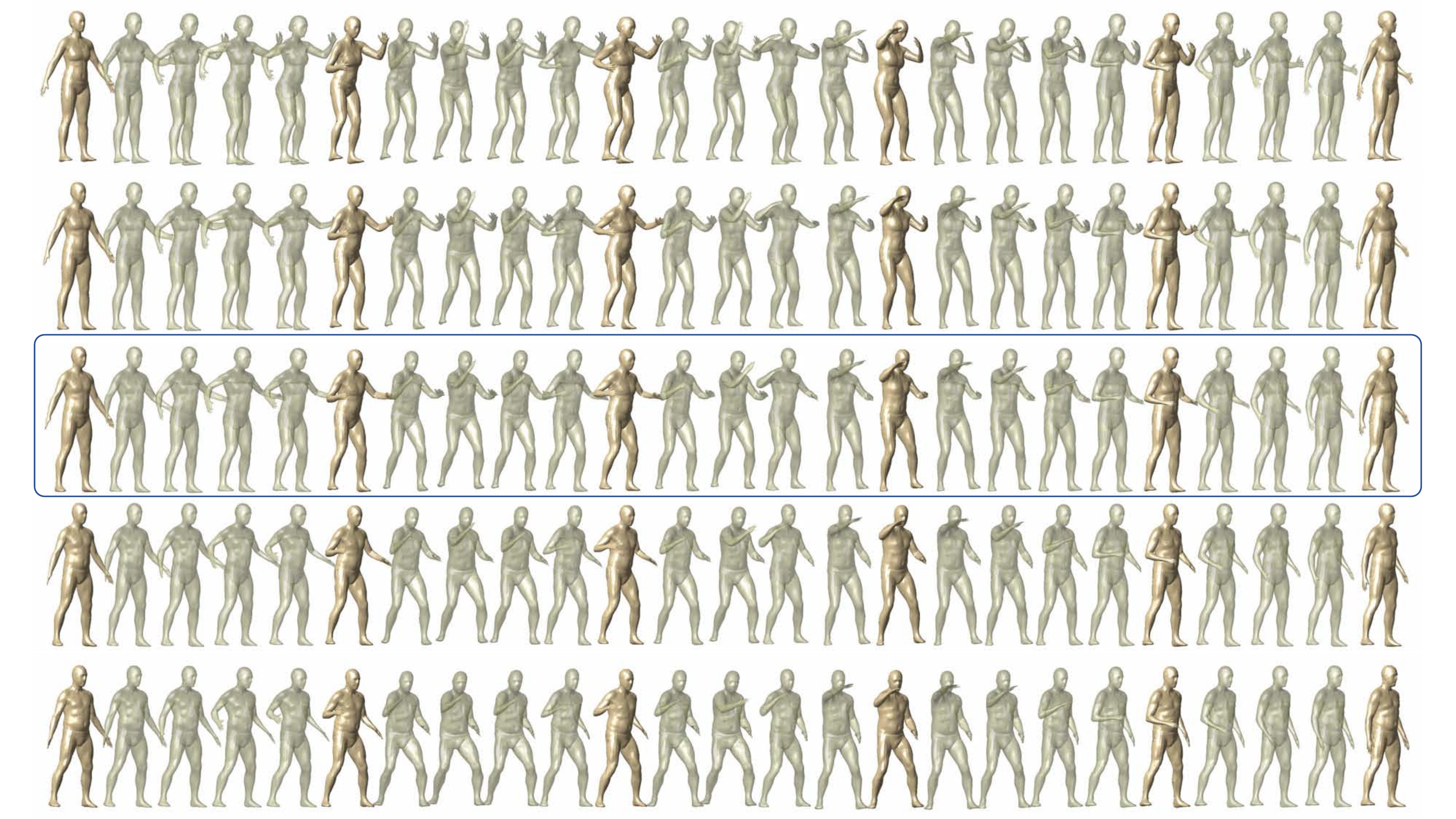}\\
	\small{(b) After registration. The highlighted middle row corresponds to the mean 4D surface.} \\
	\midrule
\end{tabular}
	\caption{\label{fig:4Dgedeosics_humans1} Example of a geodesic between 4D surfaces corresponding to punching actions ((a) before registration and (b) after registration).  In each example, we show  the source 4D surface in the first row, the target 4D surface in the last row, and three intermediate 4D surfaces along the geodesic between the source and the target. Observe how misaligned are the highlighted frames  before registration. A video illustrating these sequences is included in the Supplementary Material.}
}
\end{figure*}

\subsection{Spatiotemporal registration and 4D geodesics}
\label{sec:results_registration}
We consider pairs of 4D facial expressions from the VOCA dataset. We first reparameterize each 4D surface using randomly generated  time-warping functions to simulate facial expressions performed at different execution rates. We then apply the framework proposed in this paper to spatiotemporally register them.  Fig.~\ref{fig:faces_spatio_temporal1} shows an example of such spatiotemporal registration. In this example, we show \textbf{(a)} the source 4D surface, \textbf{(b)} the target 4D surface before spatiotemporal registration, and \textbf{(c)} the target 4D surface after spatiotemporally registering it to the source. We also highlight some key frames. As one can see, the original 4D surfaces differ significantly in their execution rates. The spatiotemporal registration framework synchronizes the  source and target expressions, thus  enabling their comparison, interpolation and averaging.  We also perform a similar experiment on the human body shapes of the DFAUST~\cite{dfaust:CVPR:2017} and CAPE~\cite{ma2020learning} datasets; see Figs.~\ref{fig:humans_spatio_temporal1},~\ref{fig:humans_spatio_temporal2}, and~\ref{fig:4Dgedeosics_ablation1}(a)-(c). Compared to faces, human body shapes are very challenging to analyze since they perform complex articulated motions, which result in large bending and stretching of their surfaces.


\vspace{6pt}
\noi\textbf{4D geodesics. } Fig.~\ref{fig:4Dgedeosics_humans1} shows geodesics between 4D human body shapes. In this example, both the source and target perform a punching action but at different rates.  We show the geodesic before and after the spatiotemporal  registration of the target 4D surface onto the source. Unlike the jumping action in Fig.~\ref{fig:humans_spatio_temporal1}, the left hand of the target surface does not perform the same action as the left hand of the source surface. Nevertheless, our framework can bring these two 4D surfaces as close as possible to each other. The supplementary material includes a video of the sequence and more examples of geodesics between 4D faces (from VOCA),  4D human bodies (from DFAUST),  and clothed 4D human bodies (from CAPE). 


\begin{table*}[!ht]
    \centering
    \caption{\label{tab:performance_spatial_registration}Comparison of the performance and accuracy of the proposed spatial registration with state-of-the-art techniques such as MAP Tree~\cite{ren2020maptree} and Fast Sinkhorn filters~\cite{pai2021fast}, which are based on functional maps~\cite{ovsjanikov2016computing}. 
    }
    \begin{tabular}{|@{ }c| ccc | ccc| ccc@{ }|}
        \hline
                 &\multicolumn{3}{c|}{\textbf{COMA}~\cite{ranjan2018generating}}&\multicolumn{3}{c|}{\textbf{CAPE}~\cite{ma2020learning}}  &\multicolumn{3}{c|}{\textbf{DFAUST}~\cite{dfaust:CVPR:2017}}  \\
       \cline{2-4} \cline{5-7} \cline{8-10}
                & \textbf{Mean} & \textbf{Std} & \textbf{Median}     &  \textbf{Mean} & \textbf{Std} & \textbf{Median}   & \textbf{Mean} & \textbf{Std} & \textbf{Median}   \\
        \hline 
          MapTree~\cite{ren2020maptree}          & $1.6042$ & $0.6956$ & $1.6069$ & 
                             $1.4447$ & $0.7227$ & $1.4483$ &    
                             $1.5344$ & $0.7065$ & $1.5211$  \\
                                                                       
           ICP-NN~\cite{pai2021fast}           & $1.6028$ & $0.6973$ & $1.6056$ & 
                              $1.4684$ & $0.7087$ & $1.5116$  & 
                              $1.5185$ & $0.6905$ & $1.5082$ \\
                                                                       
            ICP-Sinkhorn~\cite{pai2021fast}     & $1.6019$ & $0.6974$ & $1.6053$ & 
                               $1.4684$ & $0.7037$ & $1.5058$& 
                               $1.5275$ & $0.6888$ & $1.5227$ \\
                                                                       
            Zoomout-NN~\cite{pai2021fast}   & $1.5997$ & $0.6902$ & $1.6002$ & 
                               $1.4743$ & $0.7029$ & $1.4781$  & 
                               $1.5101$ & $0.6922$ & $1.4857$ \\
                                                                       
            Zoomout-Sinkhorn~\cite{pai2021fast} & $1.6016$ & $0.6908$ & $1.5968$ & 
                               $1.4737$ & $0.6937$ & $1.4925$ & 
                               $1.5019$ & $0.6948$ & $1.4731$ \\
            \hline
            SRNF (ours)      & $\textbf{0.0012}$ & $\textbf{0.0008}$ & $\textbf{0.0003}$ & 
                               $\textbf{0.0008}$ &$\textbf{0.0008}$ & $\textbf{0.0006}$ & 
                               $\textbf{0.0008}$ & $\textbf{0.0008}$ & $\textbf{0.0007}$ \\
        \hline
    \end{tabular}
\end{table*}

\vspace{6pt}
\noi\textbf{Evaluation of the spatial registration.}  We quantitatively evaluate the accuracy of the proposed spatial registration method and compare it to the latest functional map-based techniques such as  MapTree~\cite{ren2020maptree} and Fast Sinkhorn filters~\cite{pai2021fast}. Similar to our method, functional maps operate on clean manifold surfaces and do not use any form of (deep) learning. We take the surfaces of COMA~\cite{ranjan2018generating}, CAPE~\cite{ma2020learning}, and DFAUST~\cite{dfaust:CVPR:2017} datasets, which come with ground-truth correspondences, and apply random spatial diffeomorphisms to them to simulate unregistered surfaces. We then compute the correspondence map between each pair of surfaces in the dataset.  We measure the spatial registration error in terms of the geodesic distance, on the parameterization domain, between the ground-truth and the computed correspondence. Table~\ref{tab:performance_spatial_registration} reports the mean, standard deviation, and median of the registration errors computed  across all the models in each data set. As one can see, the proposed SRNF-based spatial registration method significantly outperforms state-of-the-art algorithms~\cite{ren2020maptree,pai2021fast}.  We refer the reader to the supplementary material, which  includes visual examples of pairs of surfaces before and after spatial registration. It also includes additional spatial registration experiments using the quadruped animal data set of Kulkarni \etal~\cite{kulkarni2020articulation}.


An important property of the proposed approach is that it finds a one-to-one mapping between the source and target surfaces. This is not the case with functional map-based methods, which can map a point on the source  to multiple points on the target. Thus, they cannot be used to compute geodesics, interpolations, and statistical  summaries.

\vspace{6pt}
\noi\textbf{Evaluation of the temporal registration.} We use the FLAME fitting framework~\cite{li2017learning} to generate random 4D facial surfaces with known ground-truth temporal registrations. We first generate two random SMPL parameters, each  corresponding to a 3D surface, and then linearly interpolate them to simulate a deforming 4D facial surface. Let $\curve_i, \  i \in \{1,\dots, 100\}$ be the resulting 4D surfaces.  Next, we generate $100$ random temporal diffeomorphisms $\curvediffeo_i$; see Fig.~\ref{fig:suppmat_diffeos}-(b) of the supplementary material. These will be used to simulate 4D facial surfaces that have different execution rates. 
 
 \begin{figure}[t]
	\begin{tabular}{@{}c@{}c@{}}
		\includegraphics[ trim={18 1 20 1}, clip, width=0.24\textwidth]{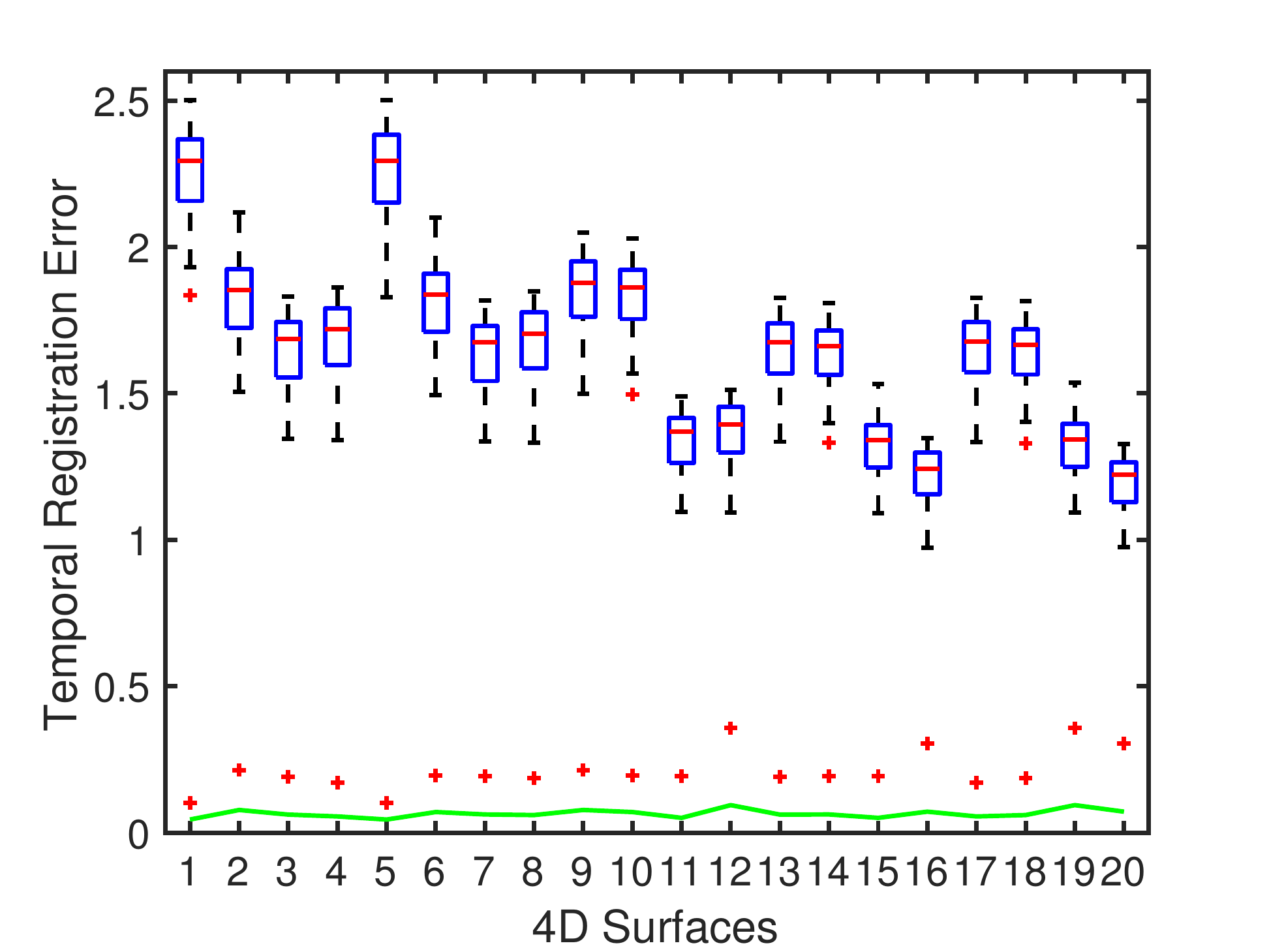} & 
		\includegraphics[trim={18 1 20 1}, clip, width=0.24\textwidth]{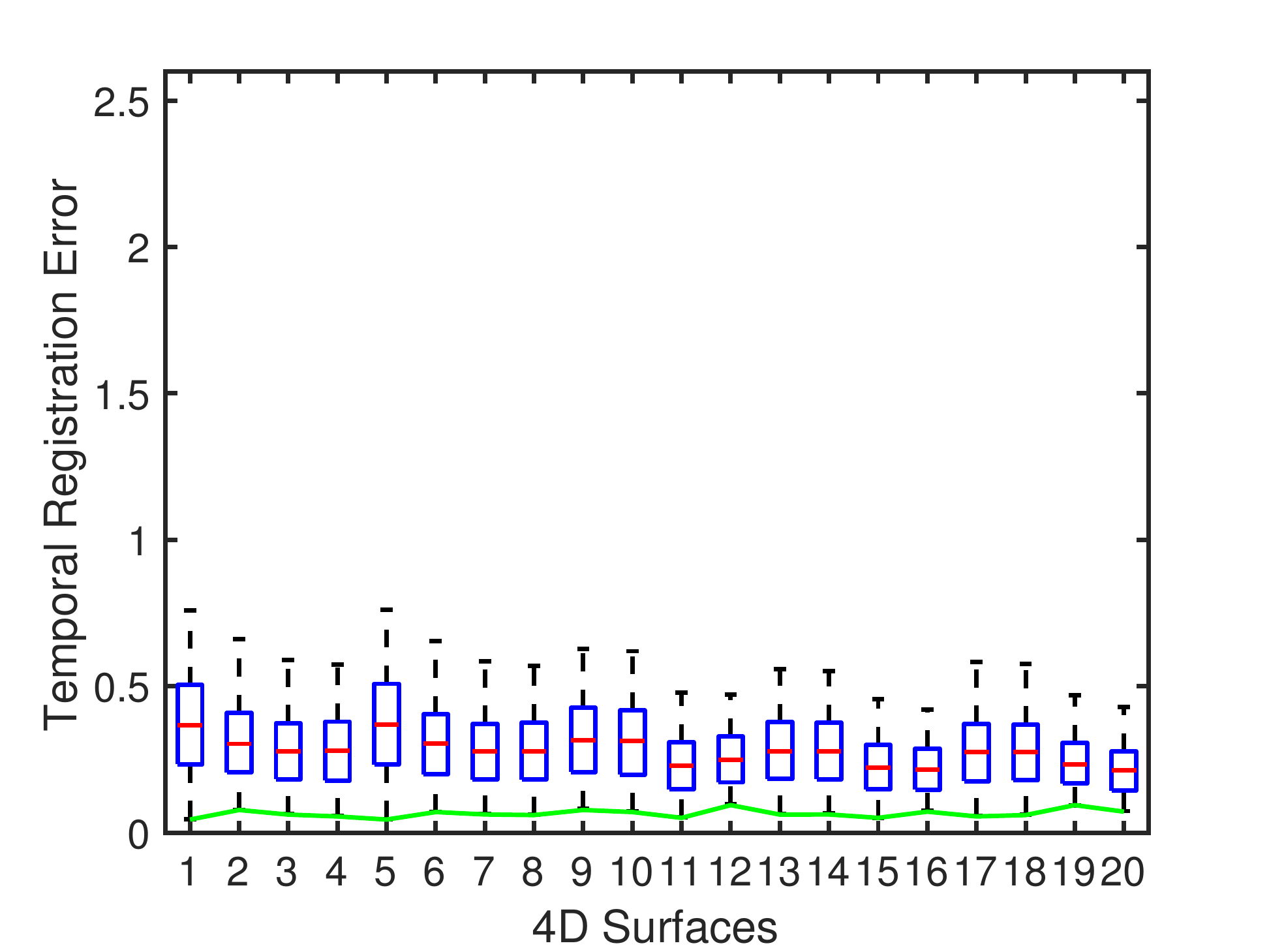} \\
		\small{(a) Before registration.} & \small{(b) After registration.}
	\end{tabular}
	\caption{\label{fig:errors_flame-fitting} Boxplots of errors between $20$ pairs of 4D surfaces:  (a) unregistered 4D surfaces generated using $100$ random diffeomorphic transformations of a single 4D surface for $5$ sequences and (b) spatiotemporally registered surfaces. The red lines represent the media error and the boxes represent its spread. The green curve in (a) and (b) is the distance between the perfectly registered 4D surfaces.}
\end{figure}

\begin{figure*}[!ht]
\center{
	\includegraphics[width=\textwidth]{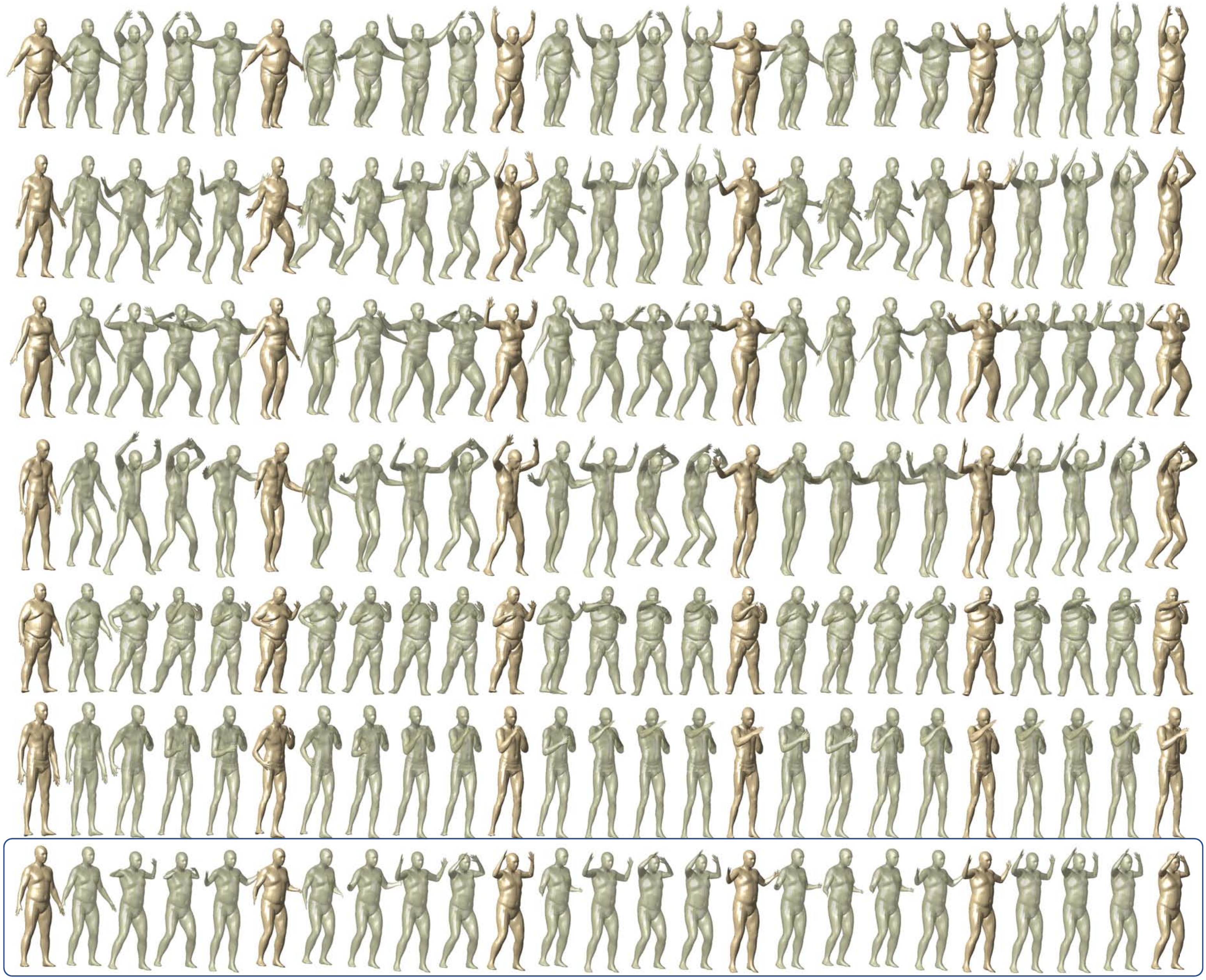} 
	\caption{\label{fig:means_humans2} Co-registration of multiple 4D surfaces. In this example, we consider four human body shapes performing a jumping action (first four rows) and two others performing a punching action (rows 5 and 6). Here, we show the spatiotemporally co-registered 4D surfaces and the 4D mean computed using the proposed algorithm. The supplementary material includes the input 4D surfaces before their spatiotemporal registration. It also includes the full video sequences. The surfaces are from the DFAUST dataset. }
	}
\end{figure*}

Now, given a pair of 4D surfaces $\curve_i$ and $\curve_j$, and for each pair of temporal diffeomosphims $\curvediffeo_k$ and $\curvediffeo_l$,   $\curve_i \circ  \curvediffeo_k$ and $\curve_j \circ \curvediffeo_l$ can be seen as a pair of 4D surfaces with different execution rates.  
Next, we compute, using the proposed framework, the optimal diffeomorphism $\curvediffeo^*_{i,j}$ that  aligns $\curve_i \circ  \curvediffeo_k$ onto $\curve_j \circ \curvediffeo_l$. Let $\curve^*_i = \curve_i \circ  \curvediffeo^*_{i,j} \circ \curvediffeo_k$.  To quantitatively evaluate the quality of the computed temporal registration, we compute:
\begin{itemize}
	\item  The distance between the perfectly registered 4D surfaces  $\curve_i$ and $\curve_j$; see the green curves in Fig.~\ref{fig:errors_flame-fitting}.
	\item The distance between $\curve_i \circ  \curvediffeo_k$ and $\curve_j  \circ \curvediffeo_l $ before temporal registration (Fig.~\ref{fig:errors_flame-fitting}-(a)), and the distance between   $\curve^*_i$ and $\curve_j$, \ie the distance between the two 4D surfaces after their temporal registration (Fig.~\ref{fig:errors_flame-fitting}-(b)). Ideally, the latter should be significantly smaller than the former. It should also be as close as possible to the  distance between the perfectly registered 4D surfaces  $\curve_i$ and $\curve_j$.
\end{itemize}


\noi Figs.~\ref{fig:errors_flame-fitting}-(a) and (b) report statistics of these errors for each pair of 4D surfaces, but aggregated over the $100$ random diffeomorphisms. As one can see, the median distance between the 4D surfaces after registration (Fig.~\ref{fig:errors_flame-fitting}-(b)) is significantly lower than the one before registration (Fig.~\ref{fig:errors_flame-fitting}-(a)). The former is significantly closer to the ground-truth (shown with green curves in Fig.~\ref{fig:errors_flame-fitting}) than the latter. 

%
%
%


\vspace{6pt}
\noi\textbf{Computation time.}   Our approach runs entirely on a CPU. The Matlab implementation of the spatiotemporal registration process takes less than $31.43$ seconds on $4.2$ GHz Intel Core i7  with $32$ GB of RAM. The visualization, which is needed when computing geodesics, means, and directions of variation, and when synthesizing random 4D surfaces, relies on the inversion of the SRNF maps. It requires $6$ seconds per frame and a total of $30$ minutes for the $300$ temporal frames used in this paper. All the experiments were performed using a high  spherical resolution  ($256\times 256$).

\begin{figure*}
	\begin{tabular}{@{}c@{}}
		\includegraphics[width=\textwidth]{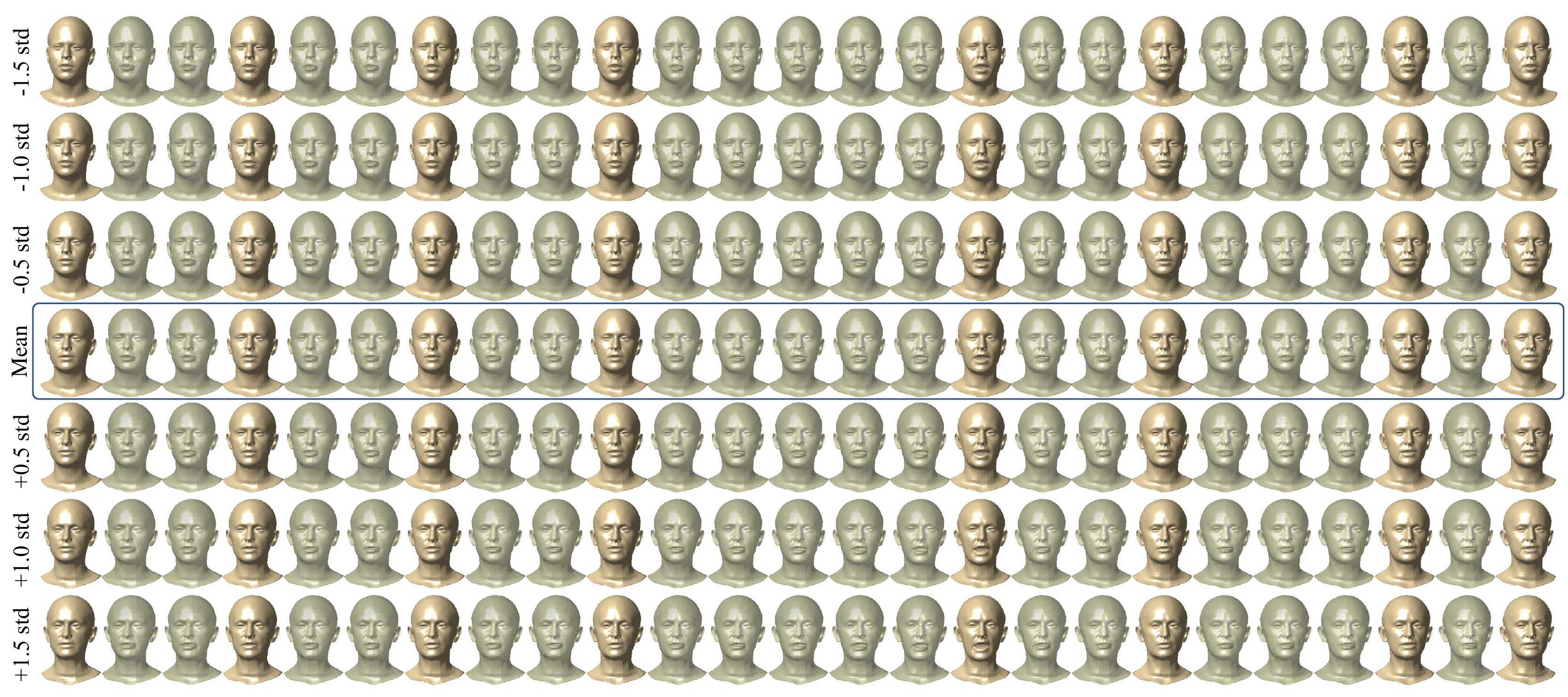}\\
		\small{(a) First mode of variation. }\\
		
		\includegraphics[width=\textwidth]{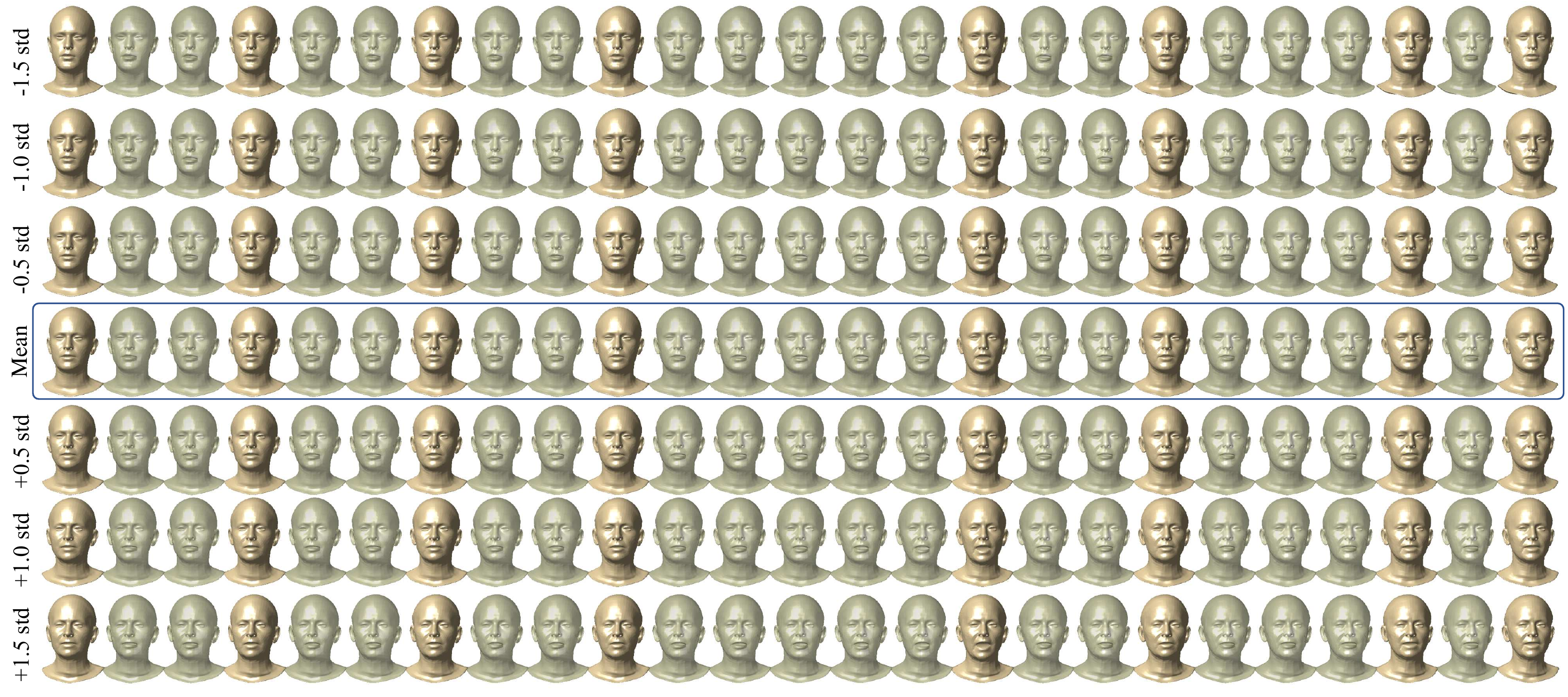}\\
		\small{(b) Second mode of variation. }\\
		
		\includegraphics[trim={0 120 0 0}, clip, width=\textwidth]{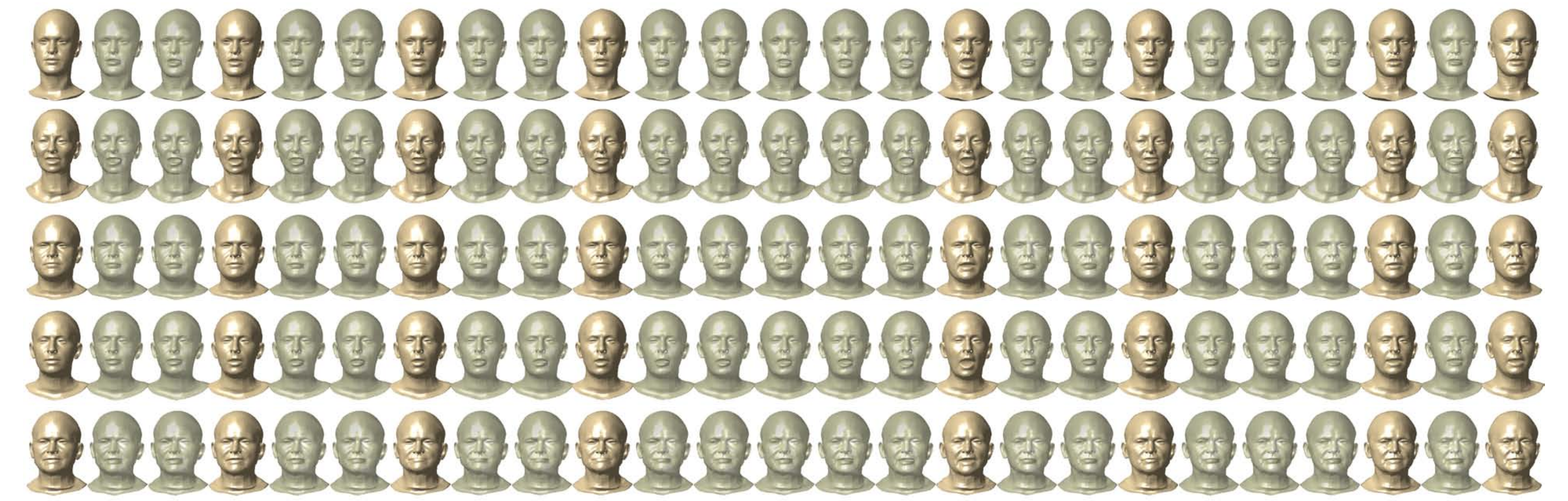}\\
		\small{(c) Five randomly synthesized 4D faces. }
	\end{tabular}
	
	\caption{\label{fig:modes_faces2} First (a) and second (b)  principal directions of variation (the mean 4D surface is highlighted in the middle). Each row corresponds to one 4D surface sampled between $-1.5$ to $1.5$ times the standard deviation along the principal direction of variation. We refer the reader to the supplementary material, which shows the input 4D faces (before their spatiotemporal registration). It also  includes  more modes of variation and random samples, as well as the complete video sequences.}
\end{figure*}

\subsection{Summary statistics}
\label{sec:results_statistics}
We now consider a set of unregistered 4D surfaces and compute their mean and principal directions of variation. Fig.~\ref{fig:means_humans2} shows the 4D mean (highlighted with a blue box) computed from six  4D human shapes performing different types of actions. The figure also shows the input 4D surfaces after their spatiotemporal registration; see the video in the supplementary material for an illustration of the input 4D surfaces before spatiotemporal registration.  Despite the large articulated motion, the large differences in the type of actions  and the significant differences in the execution rates of the 4D surfaces, our framework is able to co-register them and generate a plausible average 4D surface.  Figs.~\ref{fig:modes_faces2}-(a) and (b) show the mean and the first two principal modes of variation computed on  input 4D facial surfaces. As we can see, the computed mean also captures the main features of the dataset. The principal  directions of variation further capture relevant variability in the given data. The supplementary material  includes the input 4D surfaces prior to their registration. Please also refer to  the videos in the supplementary material for additional results.

 

\subsection{4D surface synthesis}
\label{sec:results_synthesis}

Fig.~\ref{fig:modes_faces2}-(c) shows five 4D facial expressions randomly sampled from a Gaussian distribution with parameters estimated from the VOCA dataset using the method of Sec.~\ref{sec:4Dstatistics}. To ensure that the synthesized 4D surfaces are plausible, we only consider those that are within $1.5$ standard deviations along each principal direction of variation. We refer the reader to the supplementary material for videos of all of the randomly generated 4D surfaces. The ability to synthesize novel 4D surfaces can benefit many applications in computer vision and graphics. It can be used to augment datasets for efficient training of deep learning models. 


\subsection{Ablation study}
\label{sec:ablation_study}
We undertake an ablation study to demonstrate the importance of each component of the proposed framework. 

\begin{figure}[t]
	\begin{tabular}{@{}c@{}|c@{}}
		\includegraphics[trim={70 0 60 0},clip, width=.24\textwidth]{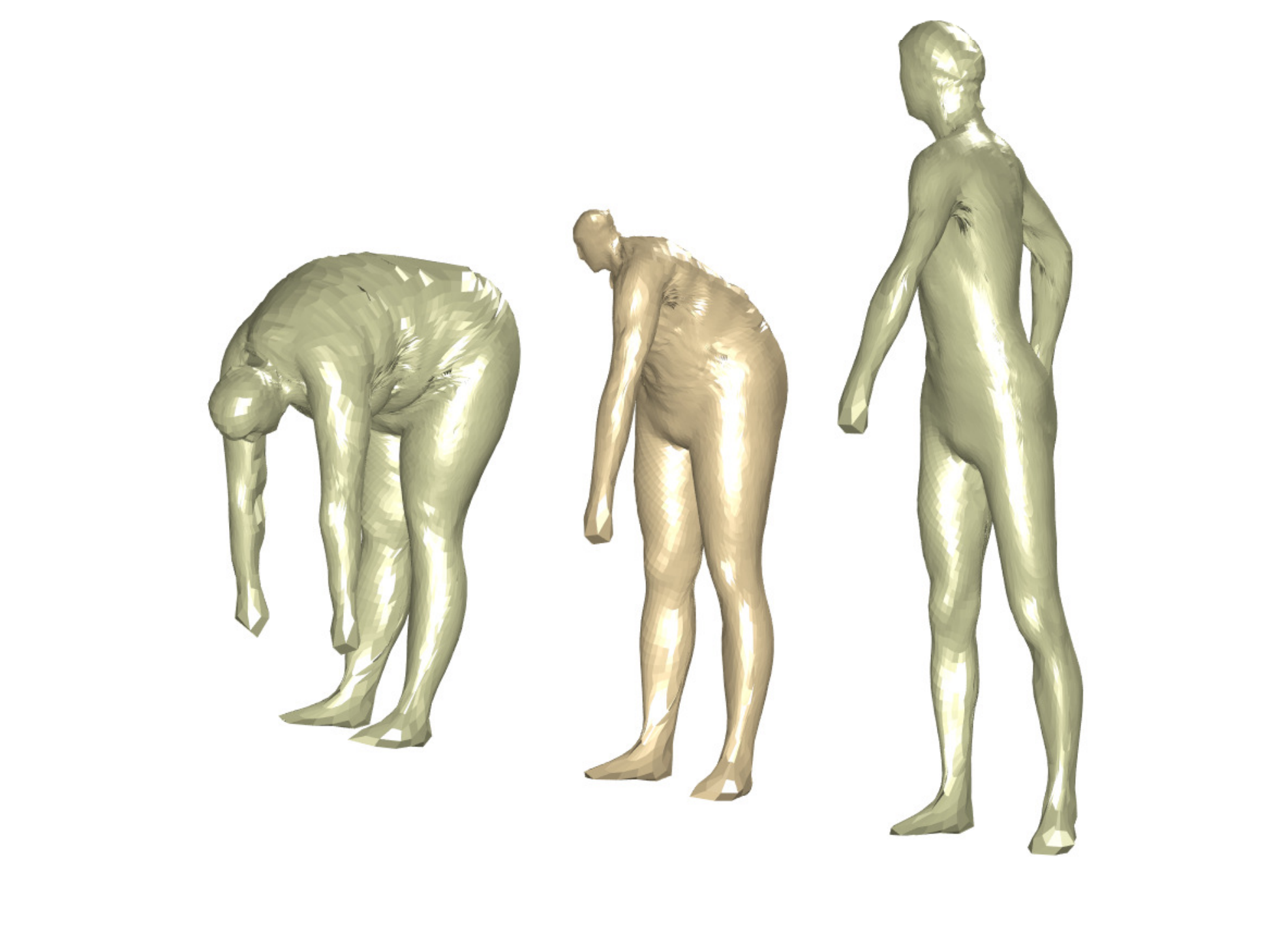} &
		\includegraphics[trim={70 0 60 0},clip, width=.24\textwidth]{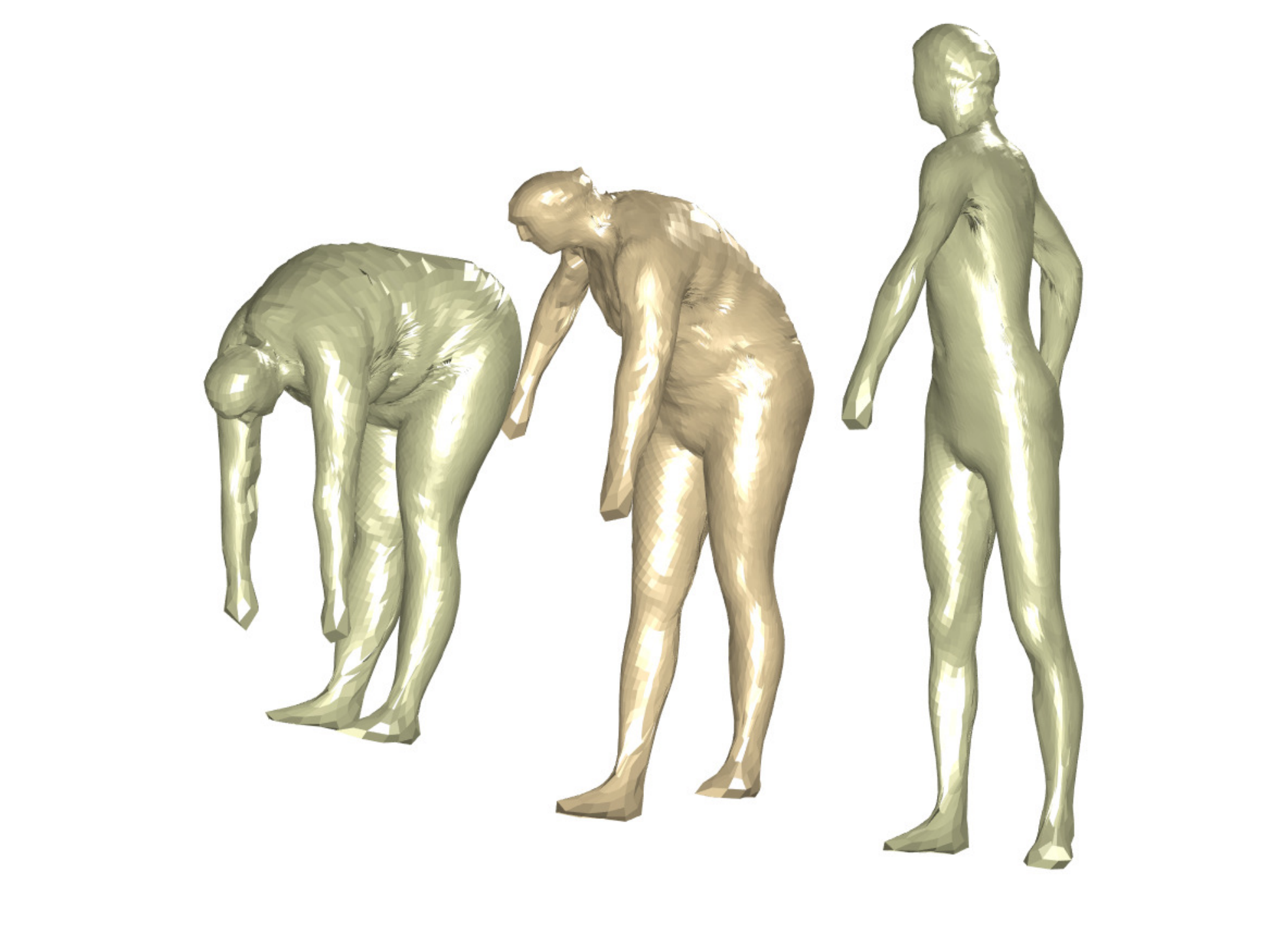}\\
		\small{(a) Without SRNF.} & \small{(b) With SRNF.}			 
	\end{tabular}
	
	\caption{\label{fig:srnf_vs_nonsrnf2}  The mean shape between the left and right surfaces, computed  \textbf{(a)} in the original surface space without the SRNF representation, and \textbf{(b)} in the SRNF space. In (a), the mean shape is distorted due to the use of the $\ltwo$ metric in the originals pace of surfaces. In both cases, the spatial registration is performed using the proposed registration method. }
\end{figure}

\vspace{6pt}
\noi\textbf{Importance of the SRNF representation.} In this experiment (Fig.~\ref{fig:srnf_vs_nonsrnf2}), we take two challenging 3D human body models, which  undergo a large articulated motion, perform their spatial registration using the proposed SRNF  approach,  and then compute their statistical mean using the $\ltwo$ metric in the original surface space (Fig.~\ref{fig:srnf_vs_nonsrnf2}-(a)) and the $\ltwo$ metric in the SRNF space (Fig.~\ref{fig:srnf_vs_nonsrnf2}-(b)). Fig.~\ref{fig:srnf_vs_nonsrnf2}-(a) shows that the articulated parts of the mean computed in the original surface space unnaturally shrink. This is predictable since, under the $\ltwo$ metric, geodesics correspond to straight lines. However, in the SRNF space, the $\ltwo$ metric is equivalent to the optimal bending and stretching of the surfaces, and thus the computed mean  is more natural; see Fig.~\ref{fig:srnf_vs_nonsrnf2}-(b).

\begin{figure*}[!ht]
\center{
    \includegraphics[clip, width=\textwidth]{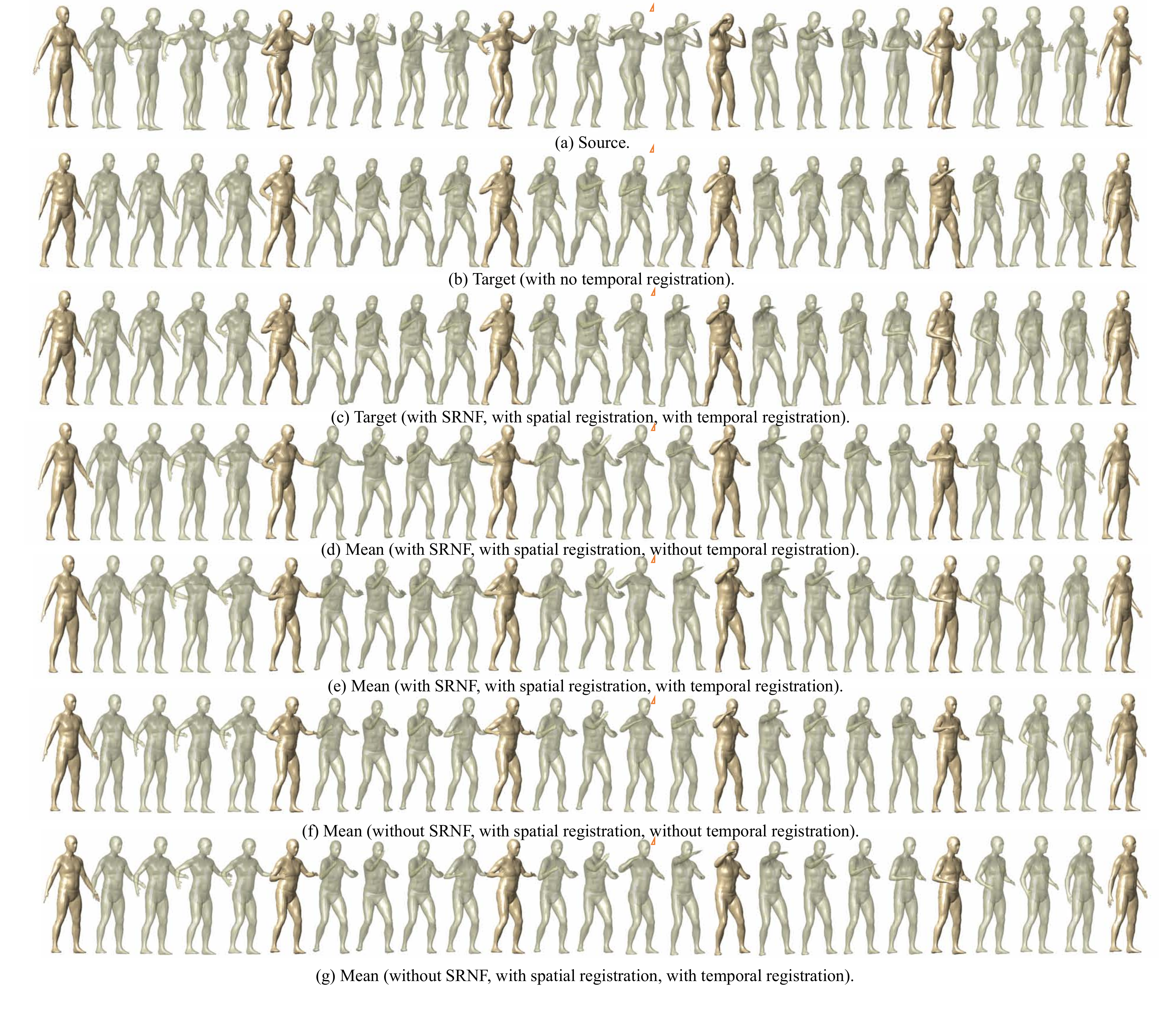}\\
    \includegraphics[trim={0 110 0 110},clip, width=.7\textwidth]{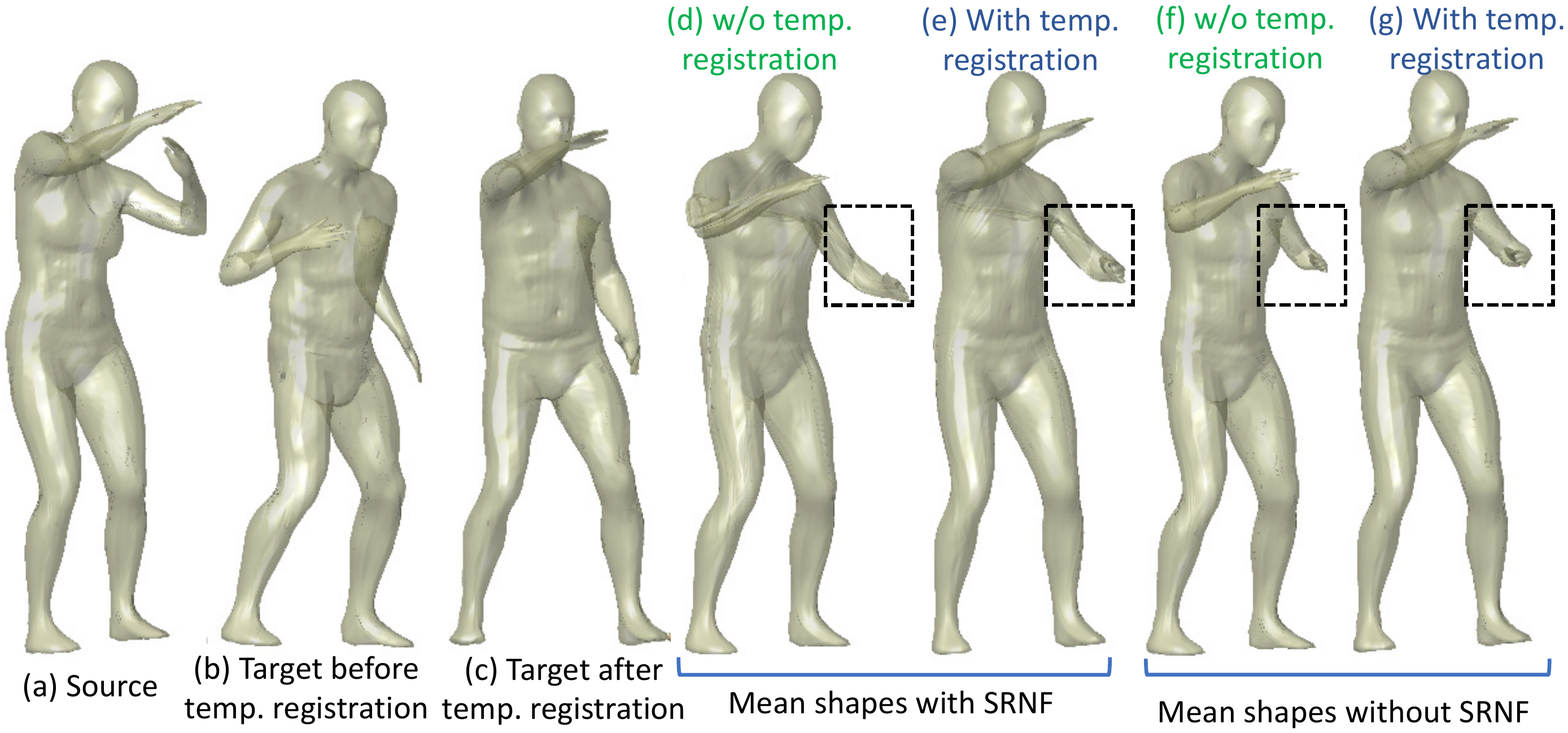}
\caption{\label{fig:4Dgedeosics_ablation1} Ablation study: illustration of the effect of the different components of the proposed framework on the quality of the computed mean 4D surface, which is the middle point along the geodesic between the source and target 4D surfaces. Te bottom row is a zoom on the frame highlighted in (a) to (g). The 4D surfaces are from the DFAUST dataset.   A video illustrating these sequences is included in the Supplementary Material.}
}
\end{figure*}

Next, we consider two full 4D surfaces of deforming human body shapes (Fig.~\ref{fig:4Dgedeosics_ablation1}-(a) and (b)) and show their  mean 4D surface obtained:  \textbf{(1)} with the SRNF representation, with spatial registration, and \textbf{without} temporal registration (Fig.~\ref{fig:4Dgedeosics_ablation1}-(d)),  \textbf{(2)} with the SRNF representation, with spatial registration, and with temporal registration (Fig.~\ref{fig:4Dgedeosics_ablation1}-(e)), \textbf{(3)} \textbf{without} SRNF representation, with spatial registration, and \textbf{without} temporal registration (Fig.~\ref{fig:4Dgedeosics_ablation1}-(f)), and \textbf{(4)} \textbf{without} SRNF representation, with spatial registration, and with temporal registration (Fig.~\ref{fig:4Dgedeosics_ablation1}-(g)). The last two cases are equivalent to  a linear interpolation in the original surface space, after spatial registration. In all cases, we perform the spatial registration using the SRNF framework.

\textbf{First}, we can see that the temporally-aligned target 4D surface (Fig.~\ref{fig:4Dgedeosics_ablation1}-(c)) is very close to the source 4D surface in Fig.~\ref{fig:4Dgedeosics_ablation1}-(a). We observe that the right hands became fully synchronized. As such, the mean 4D surface obtained after temporal registration  (Fig.~\ref{fig:4Dgedeosics_ablation1}-(e)) is fully synchronised with the source and the aligned target, unlike the mean 4D surface in Fig.~\ref{fig:4Dgedeosics_ablation1}-(d), which has been obtained without temporal registration. \textbf{Second}, in the mean 4D surfaces obtained without the SRNF framework (Figs.~\ref{fig:4Dgedeosics_ablation1}-(f) and (g)),   we can observe that the parts that undergo large articulated motion (\eg the arms) unnaturally shrink. This shrinkage is stronger in  Fig.~\ref{fig:4Dgedeosics_ablation1}-(f) since the mean is obtained without temporal registration. The bottom row of Fig.~\ref{fig:4Dgedeosics_ablation1} shows a zoom on the  time frame highlighted  in Figs.~\ref{fig:4Dgedeosics_ablation1}-(a) to (g).

Finally, we quantitatively evaluate the importance of the SRNF representation by comparing the expressive power of PCA on the original space of spatially registered surfaces and on the space of SRNFs. We randomly divide a data set equally into a training set and a testing set. We then fit a PCA model to the training set (both in the original space and in the space of SRNFs), project each model in the test set onto the PCA model, reconstruct it, and measure the error between the original  and the reconstructed models. We perform 5-fold cross-validation. Table~\ref{tab:surfaces_pca_expressiveness} reports the mean, the median, and the standard deviation of the error over the test set and averaged over the five runs. As one can see, PCA on the SRNF space has a significantly lower reconstruction error than PCA in the original space of surfaces. This demonstrates that the former is more suitable to characterize variability in the shape of 3D objects that bend and stretch.



\begin{table}[t]
    \centering
    \caption{\label{tab:surfaces_pca_expressiveness}Comparison of the expressive power of  PCA on the original space of surfaces and on the space of SRNFs. The lower the values are, the better.}
    \begin{tabular}{|@{ }c|c|c|c|c|c|c@{ }|}
        \hline
                    & \multicolumn{3}{c|}{\textbf{PCA on surfaces}}  & \multicolumn{3}{c|}{\textbf{PCA on SRNFs}}  \\
                    \cline{2-7}
                    & \textbf{Mean} & \textbf{Std} & \textbf{Median} & \textbf{Mean} & \textbf{Std} & \textbf{Median} \\
        \hline
        DFAUST      &  $0.26$ & $0.042$ & $0.270$ &  $\textbf{0.12}$ & $0.023$ & $\textbf{0.121}$ \\
        \hline
        VOCA        &   $0.030$ & $0.011$ & $0.028$ &  $\textbf{0.009}$ & $0.004$ & $\textbf{0.008}$ \\
        \hline
        CAPE        & $0.105$ & $0.041$ & $0.101$ &  $\textbf{0.053}$ & $0.021$ & $\textbf{0.050}$ \\
        \hline
    \end{tabular}
\end{table}

\begin{table}[t]
    \centering
    \caption{\label{tab:curves_pca_expressiveness}Comparison of the expressive power of PCA on the original space of curves and on the space of TSRVFs. The lower the values are, the better.}
    \begin{tabular}{|@{ }c|c|c|c|c|c|c@{ }|}
        \hline
                    & \multicolumn{3}{c|}{\textbf{PCA on curves}}  & \multicolumn{3}{c|}{\textbf{PCA on TSRVFs}}  \\
                    \cline{2-7}
                    & \textbf{Mean} & \textbf{Std} & \textbf{Median} & \textbf{Mean} & \textbf{Std} & \textbf{Median} \\
        \hline
        DFAUST      &  $0.926$ & $1.110$ & $0.693$ &  $\textbf{0.756}$ & $0.104$ & $\textbf{0.715}$ \\
        \hline
        VOCA        &   $0.676$ & $0.182$ & $0.64$ &  $\textbf{0.486}$ & $0.318$ & $\textbf{0.603}$ \\
        \hline
    \end{tabular}
\end{table}

\vspace{6pt}
\noi\textbf{Importance of the TSRVF representation for 4D surfaces.} We perform a similar ablation study, but on 4D surfaces,  to compare the expressive power of PCA on the original space of curves and on the space of TSRVFs. Table~\ref{tab:curves_pca_expressiveness} shows that PCA error on the TSRVF space is  lower than  the error in the original space. This demonstrates that the former is more suitable to characterize variability in 4D surfaces.

\section{Conclusion }
\label{sec:summary}

We have proposed a new framework for the statistical analysis of longitudinal 3D shape data (or 4D surfaces), \ie surfaces that deform over time, \eg 3D human body shapes performing actions at different execution rates or 3D human faces pronouncing sentences at different speeds. Unlike traditional techniques, which only consider how features such as landmarks or measurements vary over time, the proposed framework considers the deformation of the entire surface of a 3D object. Our key contribution is in representing 4D surfaces as trajectories in the space of SRNFs, and the use of Transported Square-Root Vector Fields to  analyze such trajectories statistically. The  proposed framework can spatiotemporally register 4D surfaces, even in the presence of large elastic deformations and significant variations in the execution rates. It is also able to compute geodesics and summary statistics, which in turn can be used to  synthesize new, unseen 4D surfaces randomly. 

Although we have demonstrated the proposed 4D analysis framework on human body shapes and facial surfaces, it is general and can be applied to other types of surfaces. Our current implementation is limited to surfaces that are homeomorphic to a sphere, but we plan  to extend the framework to higher-genus surfaces by exploring different parameterization methods, including mesh-based representations~\cite{bauer2021numerical}. The approach uses the numerical SRNF inversion procedure of Laga \etal~\cite{laga2017numerical}, which is sometimes not accurate near the poles of the parameterization domain; we plan to improve its performance via the use of charts. 

The framework deals with surfaces that bend and stretch but do not change in topology; as such, it does not apply to tree-like shapes, \eg botanical trees or roots. However, the concept of representing deformations as trajectories in a shape space also applies to  tree-shape spaces such as those used in~\cite{wang2018shape,wang2020statistical}. The framework is also limited to clean surfaces that are free of geometric and topological noise; as such, the proposed spatial registration method cannot be used to register partial scans to each other, or to register a template to partial scans. However, similar to statistical shape models such as 3D morphable models and SMPL, the proposed 4D atlas can be used as a prior; in conjunction with a data generation model, it can thereby be applied to noisy or partial data, \eg to  reconstruct entire 4D surfaces. The statistical analysis presented in this paper assumes that the population of the 4D surfaces follows a Gaussian distribution. We plan to extend the approach to other types of distributions, e.g., Gaussian Mixture Models, which can represent  populations that follow multimodal distributions

The proposed framework has various applications in computer vision, graphics, biology, and medicine. In computer vision, collecting large animations to train deep neural networks, \eg for 3D reconstruction or action recognition~\cite{han2019image,laga2020survey}, is complex and time-consuming. Our framework can contribute to solving this problem by automatically synthesizing new samples from a small dataset. Our current implementation has only considered random synthesis, which is very important for populating virtual environments and for data augmentation to train deep learning networks. However, there are many situations where we would like to control this process using a set of parameters. For instance, when dealing with 4D facial expressions, these parameters can be the degree of sadness, facial dimensions, etc. This type of control can be  implemented efficiently using regression in the TSRVF space. Finally, our framework can  be used to statistically analyze how anatomical organs deform due to  growth or disease progression.  


\vspace{6pt}
\noi\textbf{Acknowledgement. }  We would like to thank the authors of~\cite{hasler2009statistical,dfaust:CVPR:2017,ranjan2018generating,VOCA2019,ma2020learning} for making their datasets publicly available, and~\cite{ren2020maptree} and~\cite{pai2021fast} for sharing the codes of MapTree and the Fast Sinkhorn Filters-based surface registration. 



\appendices
\section{Spatial registration}
\label{sec:supp_spatial_registration}
This appendix discusses the implementation details of the spatial registration and provides additional results on various complex 3D shapes. 

\subsection{Algorithm}
\label{sec:supp_spatial_registration_algorithm}
In this section, we provide more details on the procedure used to solve the spatial registration problem in Eqn.~\eqref{eq:spatial_reg_in_srnfs} of the main manuscript. In our formulation, we consider spatial registration of a surface $\surface_2$ to a surface $\surface_1$ as the problem of  finding the rotation and reparameterization that bring the SRNF $\srnf_2$ of $\surface_2$ as close as possible to the SRNF $\srnf_1$ of $\surface_1$. We measure closeness using the $\ltwo$ metric in the SRNF space, which is equivalent to the partial elastic metric in the original space of surfaces. That is, we seek to solve the following optimization problem:
\begin{equation}
  	(\rotation^*, \diffeo^*) =\argmin_{\rotation \in \rotations, \diffeo\in \diffeos} \|\srnf_1 - \rotation (\srnf_2 \ast \diffeo)\|,
	 \label{eq:supp_spatial_reg_in_srnfs}
\end{equation}

\noi where $\ast$ is the composition operator  between an SRNF and a diffeomorphism $\diffeo\in\diffeos$.  This joint optimization over $SO(3)$ and $\Gamma$ can be solved by alternating, until convergence, between the two marginal optimizations:
\begin{itemize}
	\item Assuming a fixed parameterization, solve for the optimal rotation using Procrustes analysis via Singular Value Decomposition (SVD). 
	\item Assuming a fixed rotation, solve for the optimal reparameterization using a gradient descent algorithm. 
\end{itemize}

\noi Below we detail each of these steps.

\vspace{6pt}
\noi\textbf{Optimization over the rotation group.}  For a fixed $\diffeo \in \diffeos$, the minimization over $\rotations$ can be performed directly using Procrustes analysis. Let $\tilde{\srnf}_{2}$ denote $(\srnf_{2}, \diffeo)\equiv \sqrt{\jacobian[\diffeo]}\isp (\srnf_{2} \circ \diffeo)$ in Eqn.~\eqref{eq:supp_spatial_reg_in_srnfs}. (Here, $\jacobian[\diffeo]$ is the determinant of the Jacobian of $\diffeo$.) The optimal rotation matrix
	\begin{equation}
		\rotation^{*} = \argmin_{\rotation\in\rotations} \norm{\srnf_1 - \rotation\tilde{\srnf}_2}^2
	\end{equation}
\noi can then be obtained using Algorithm~\ref{alg:optimal_rotation}. 

\begin{algorithm}[!ht]
\hspace*{-6pt} \textit{Input:}  Two surfaces $\{\surface_1, \surface_2\} \in {\surfaces}$.\newline
\hspace*{-7pt}\textit {Output:} Optimal rotation matrix $\rotation^*$ and optimally rotated surface $\surface_2^*$. 

\begin{algorithmic}[1]
 \STATE Compute the SRNFs $\srnf_1=\srnfmap(\surface_1)$ and $\srnf_2=\srnfmap(\surface_2)$.
 \STATE Compute the $3 \times 3$ matrix $A = \int_{\domain} \srnf_{1}(s) {\tilde \srnf}_{2}(s)^{\top} ds$.
 \STATE Compute the singular value decomposition $A = U \Sigma V^{T}$.
 \STATE Compute the optimal rotation as $\rotation^{*} = UV^{\top}$. (If the determinant of $A$ is negative, the last column of $V$ changes sign.)
 \STATE Compute the optimally rotated surface $\surface_2^*=\rotation^*\surface_2$. 
\end{algorithmic}
\caption{Optimal rotational alignment of two surfaces.}
\label{alg:optimal_rotation}
\end{algorithm}

\vspace{6pt}
\noi\textbf{Optimization over the space of spatial diffeomorphisms.}  We use a gradient descent approach to solve the optimization problem over $\diffeos$. Although this approach has an obvious limitation of converging to a local solution, it is still general enough to be applicable to general surfaces and provides plausible results. In order to specify the gradient, we focus on the current iteration, and define the reduced cost function $\Ereg: \diffeos \map \rnonneg$:
	\begin{equation}
		\Ereg(\diffeo)  = \norm{\srnf_{1}  - (\tilde{\srnf}_{2}, \diffeo)}^2  = \norm{\srnf_{1}  - \phi(\diffeo)}^2,
	\end{equation}

\noi where $\tilde{\srnf}_2 = (\srnf_2, \diffeo_0)$, $\diffeo_0$ and $\diffeo$ denote  the current and the incremental reparameterizations respectively, and $\phi:\diffeos \map [\srnf_2]$ is defined to be $\phi(\diffeo) = (\tilde{\srnf}_2, \diffeo)$. Let $b$  be a unit vector in $T_{\diffeo_\id}(\diffeos)$ for $\diffeo_{\id}(s) = s$. Then, the directional derivative of $\Ereg$ at
$\diffeo_{\id}$, in the direction of $b$, is given by $\inner{\srnf_1-\phi(\diffeo_{\id})}{d\phi(b)} b$, where $\phi_{\ast}$ is the differential of $\phi$ and $\inner{\cdot}{\cdot}$ is the $\ltwo$ inner  product. If we have an orthonormal basis for
$T_{\diffeo_\id}(\diffeos)$, we can specify the full gradient of $\Ereg$ with respect to $\diffeo$, which is an element of $T_{\diffeo_\id}(\diffeos)$ given by $\partial
\diffeo\backsimeq\sum_{b_i\in{\cB}_I} \innertwo{\srnf_1-\tilde{\srnf}_2}{d\phi(b_i)} b_i$. This linear combination of the orthonormal basis elements of $T_{\diffeo_\id}(\diffeos)$ provides the incremental update of $\tilde{\srnf}_2$ in the orbit $[\srnf_2]$. 

This leaves two remaining issues: \textbf{(1)} the specification of an orthonormal basis of $T_{\diffeo_\id}(\diffeos)$, and \textbf{(2)} an expression for $\phi_{\ast}$. In our implementation, we use gradients of the spherical harmonic basis to define the orthonormal basis of $T_{\diffeo_\id}(\diffeos)$; see Section 3.2.2 of~\cite{jermyn2017elastic}. The expression of $\phi_{\ast}$ is also derived in Section 3.2.2 of~\cite{jermyn2017elastic}. With this, the optimization over the space of diffeomorphisms can be performed using Algorithm~\ref{alg:spatial_reg}.

\begin{algorithm}[!h]
\hspace*{-6pt} \textit{Input:} Two surfaces $\{f_1,\ f_2\}\in{\surfaces}$ and small step size $\epsilon$.\newline
\hspace*{-8pt}  \textit {Output:} Optimal registration $\diffeo^*$ and optimally registered surface $f_2^*$. 

\begin{algorithmic}[1]
 \STATE Generate basis $\cB_I=\{b_i,\ i=1,\dots,N\}$ using\\ spherical harmonics.
 \STATE Compute the SRNFs $\srnf_1=\srnfmap(\surface_1)$ and $\srnf_2=\srnfmap(\surface_2)$.
 \STATE Initialize $\diffeo_0=\diffeo_{id}$, $\srnf_2^0=\srnf_2$ and $j=0$.
 \STATE For each $b_i,\ i=1,\dots,N$, compute $d\phi(b_i)$.
 \STATE Compute the registration update $\partial\diffeo=\sum_{b_i\in{\cB}_I} \innertwo{\srnf_1-{\srnf}^j_2}{d\phi(b_i)} b_i$.
 \STATE Apply the registration update using $\diffeo_{j+1}=\diffeo_{j}\circ(\diffeo_{id}+\epsilon\partial\diffeo)$.
 \STATE Update $\srnf_2^{j+1}=(\srnf_2^0,\diffeo_{j+1})$ and $j=j+1$.
 \STATE Iterate steps 4-7 until convergence.
 \STATE Let $\diffeo^*=\diffeo_{j}$ and $\surface_2^*= \surface_2\circ\diffeo^*$. 
\end{algorithmic}
\caption{Optimal registration of two surfaces.}
\label{alg:spatial_reg}
\end{algorithm}

\begin{figure}[!t]
	\begin{tabular}{@{}c@{ }|c@{}}
		\includegraphics[trim={84 0 60 0},clip, width=.24\textwidth]{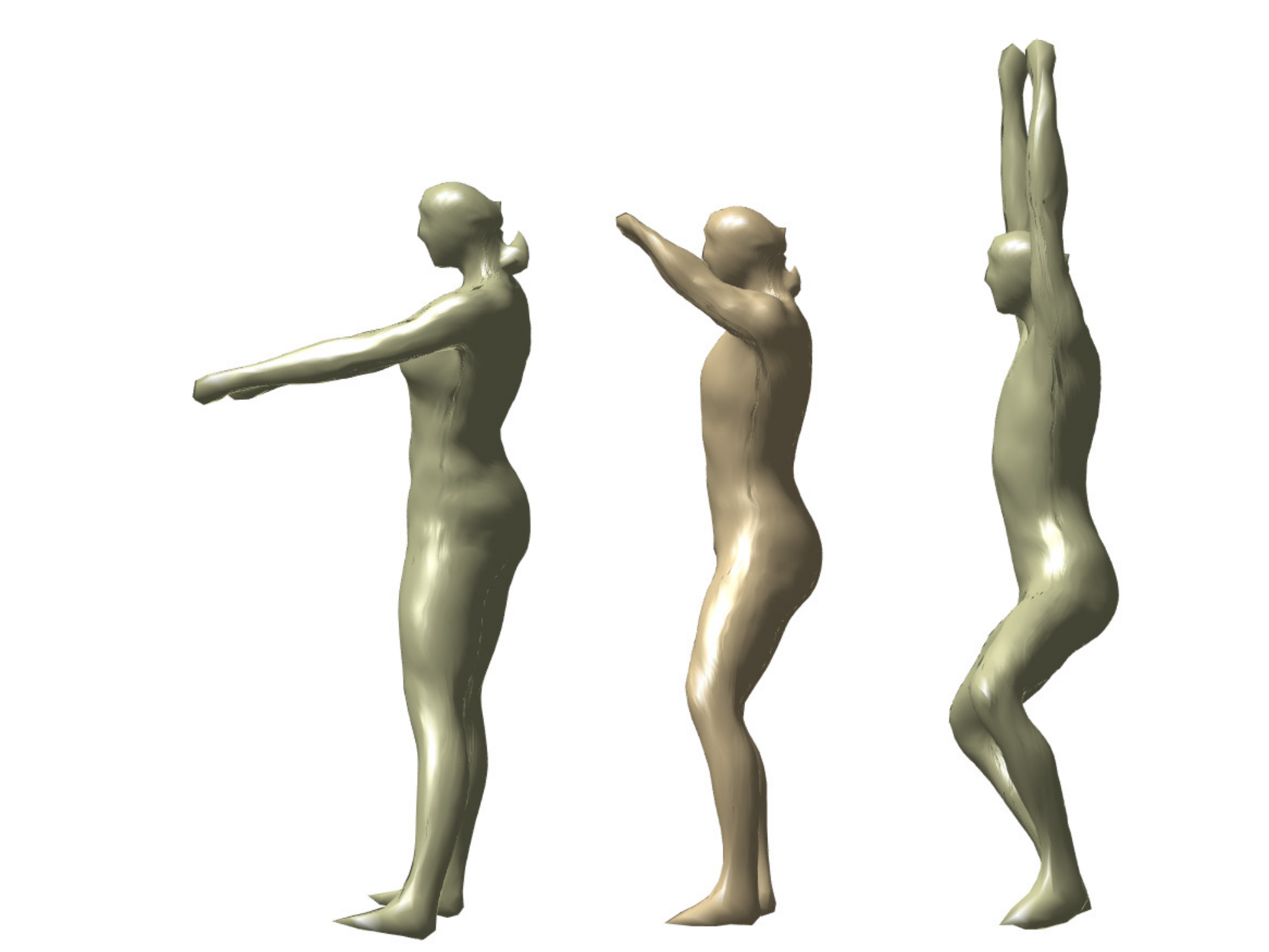} &
		\includegraphics[trim={84 0 60 0},clip, width=.24\textwidth]{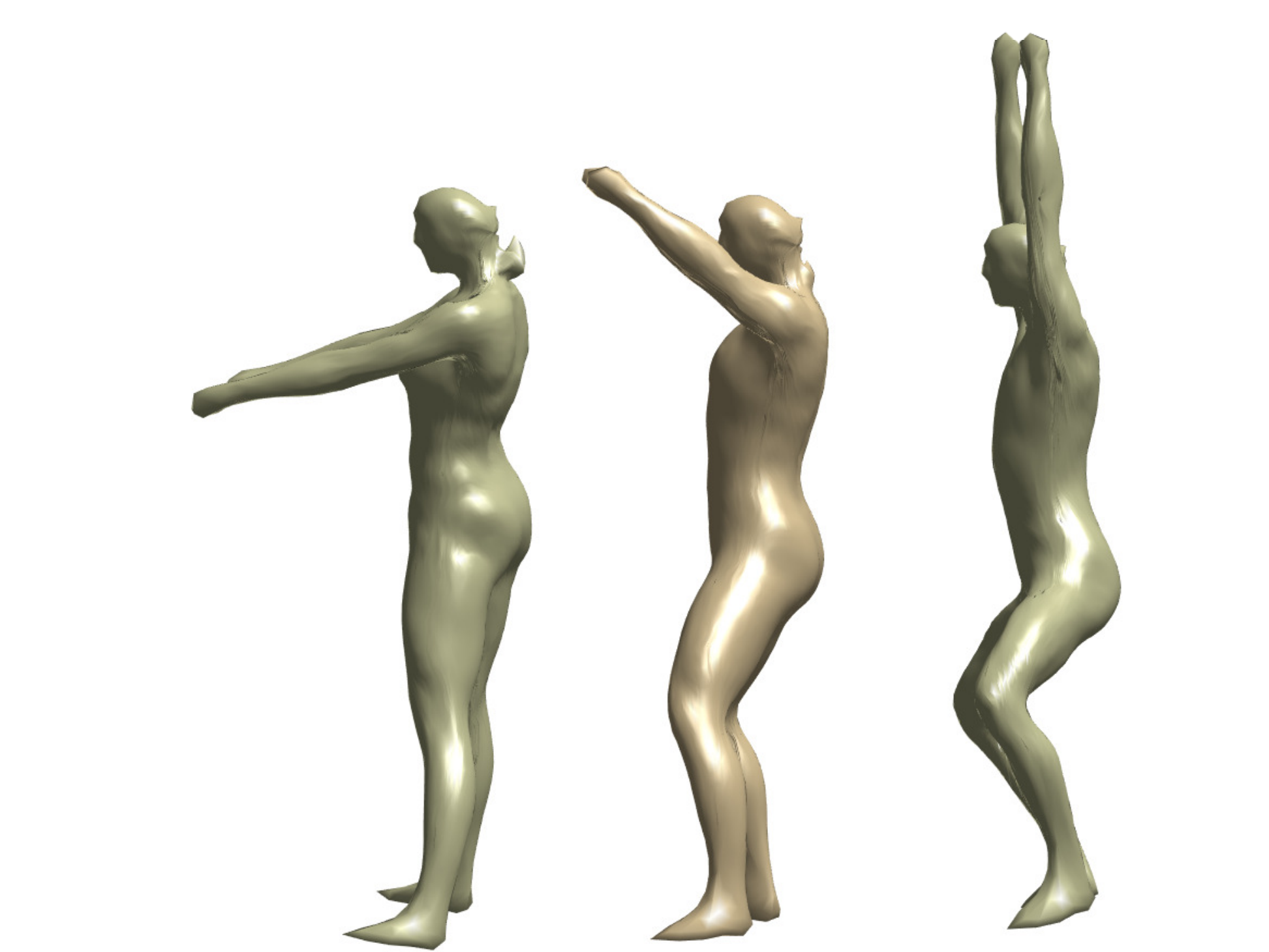}\\
		\small{(a) Without SRNF.} & \small{(b) With SRNF.}			 
	\end{tabular}
	
	\caption{\label{fig:supp_srnf_vs_nonsrnf1}  The mean shape between the left and right surfaces, computed \textbf{(a)} in the original surface space without the SRNF representation, and \textbf{(b)} in the SRNF space. The computed mean shape is shown in the middle of each subfigure. Observe that in (a), elongated parts such as the arms significantly shrink while in (b) they bend in a natural fashion.}
\end{figure}

\begin{figure}[b]
	 \begin{tabular}{@{}cc@{}}
	 	\includegraphics[trim={170 0 170 0},clip, width=0.25\textwidth]{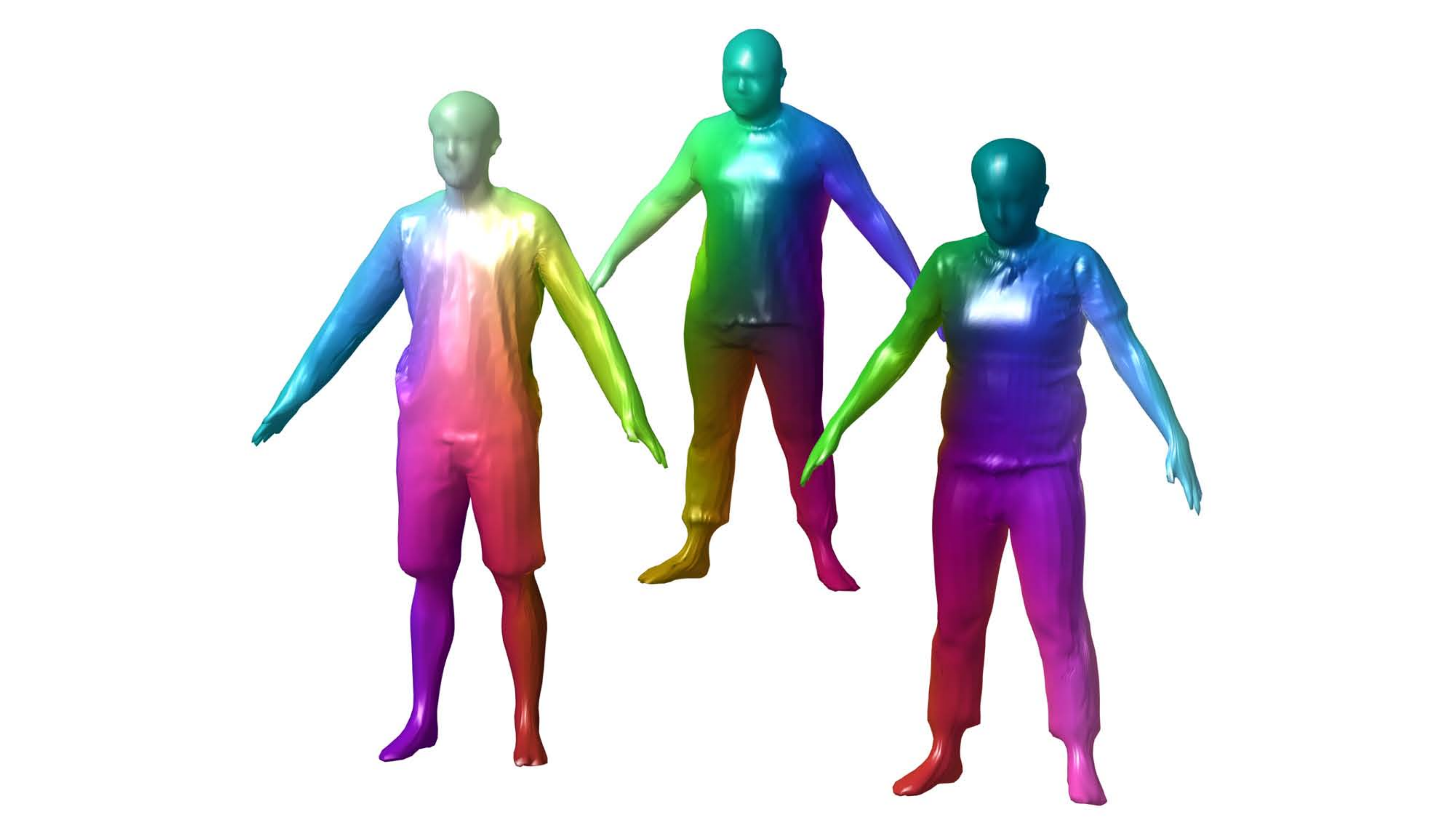} &
		\includegraphics[trim={220 0 220 0},clip, width=0.22\textwidth]{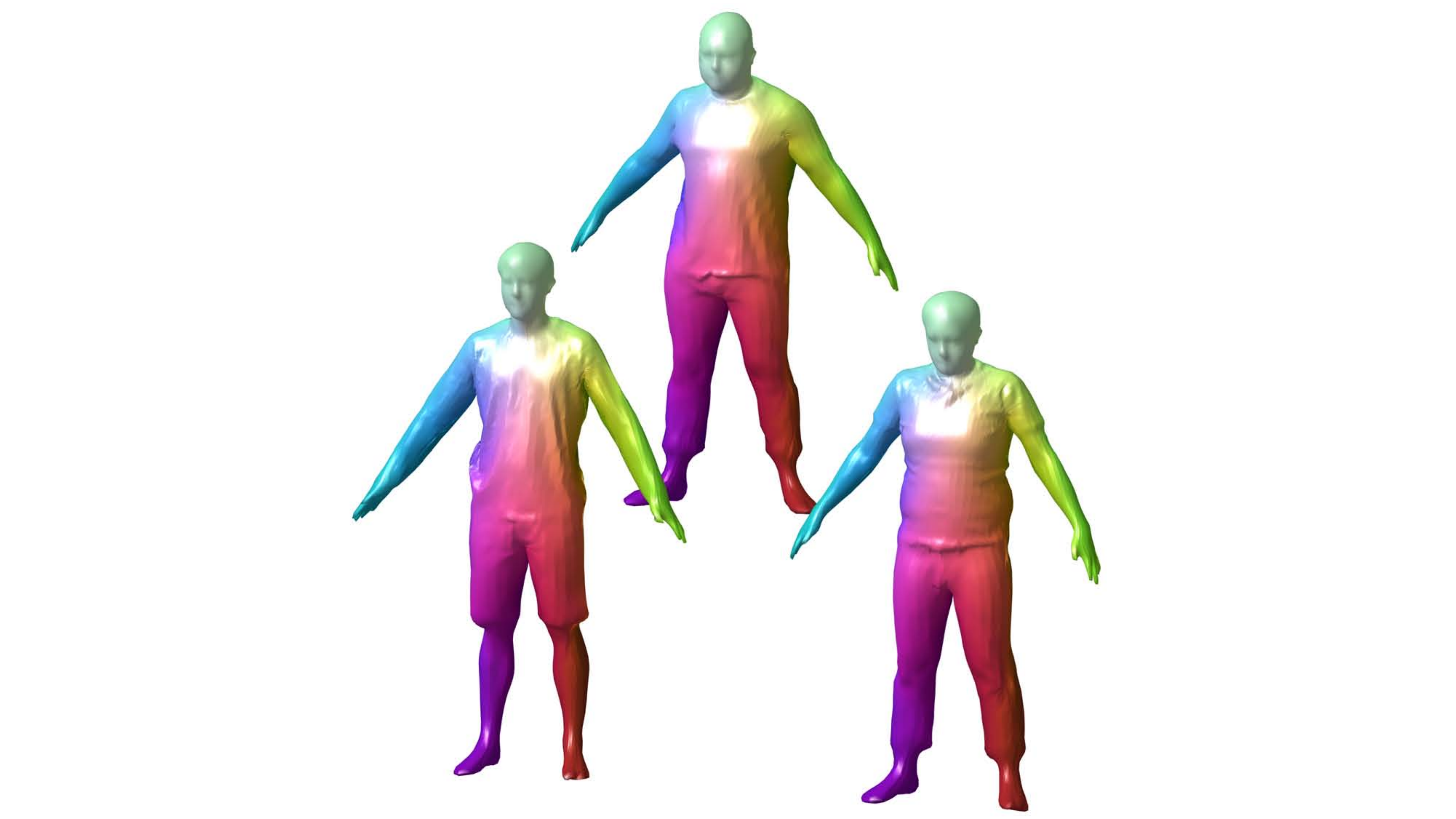}\\
		\small{(a) Input (before  registration).} & \small{(b) After  registration.}
	\end{tabular}
	 \caption{\label{fig:cape_spatial_registrations} Illustration of the  spatial correspondences, on samples from the CAPE dataset~\cite{ma2020learning},  before and after applying the proposed spatial registration algorithm. Correspondences are color-coded, \ie points that are in correspondence are rendered with the same color.  }
\end{figure}

\begin{figure*}[tb]
	 \begin{tabular}{@{}cc@{}}
	 	\includegraphics[trim={100 0 100 0},clip, width=0.5\textwidth]{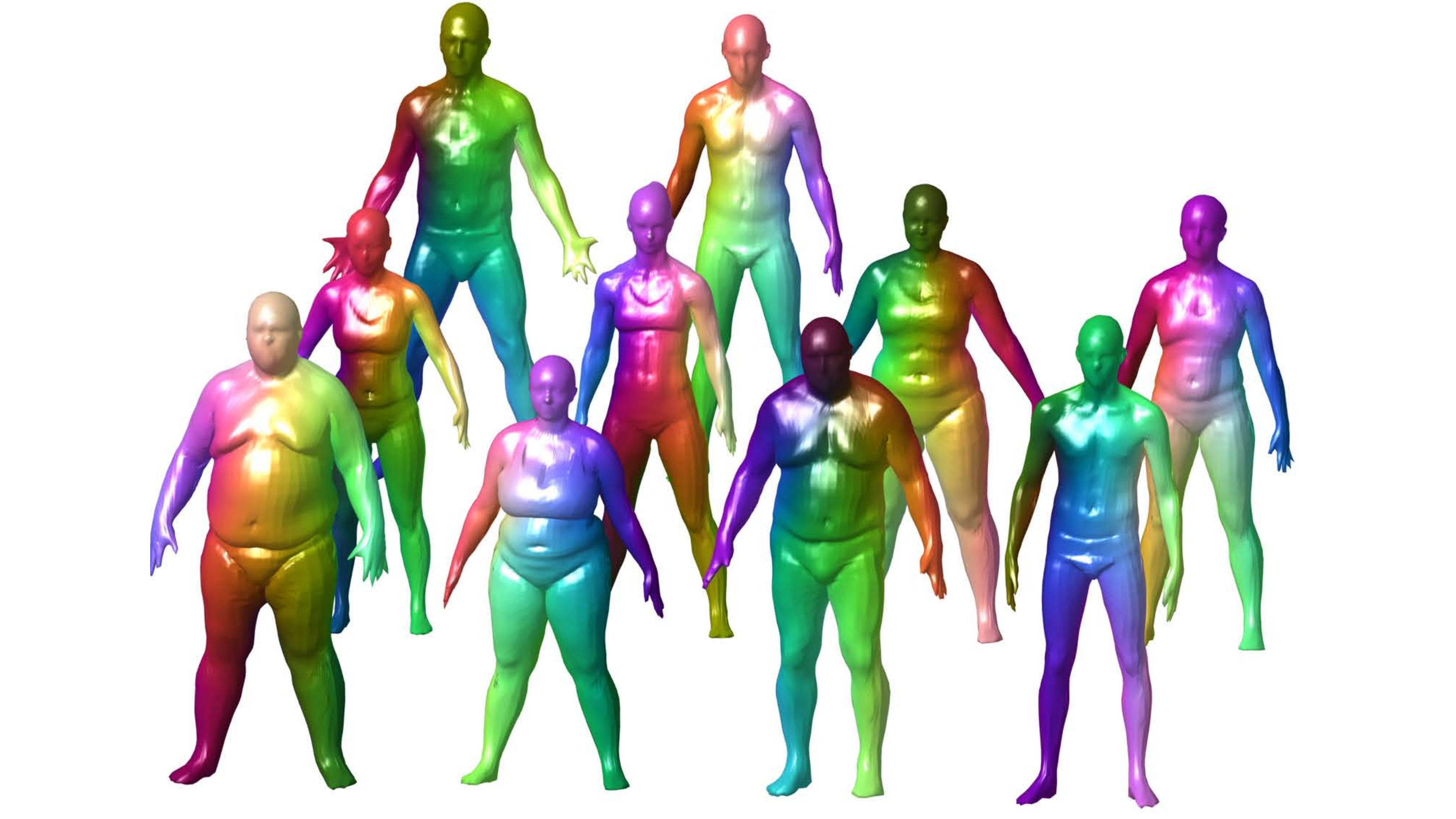} & \includegraphics[trim={100 0 100 0},clip, width=0.5\textwidth]{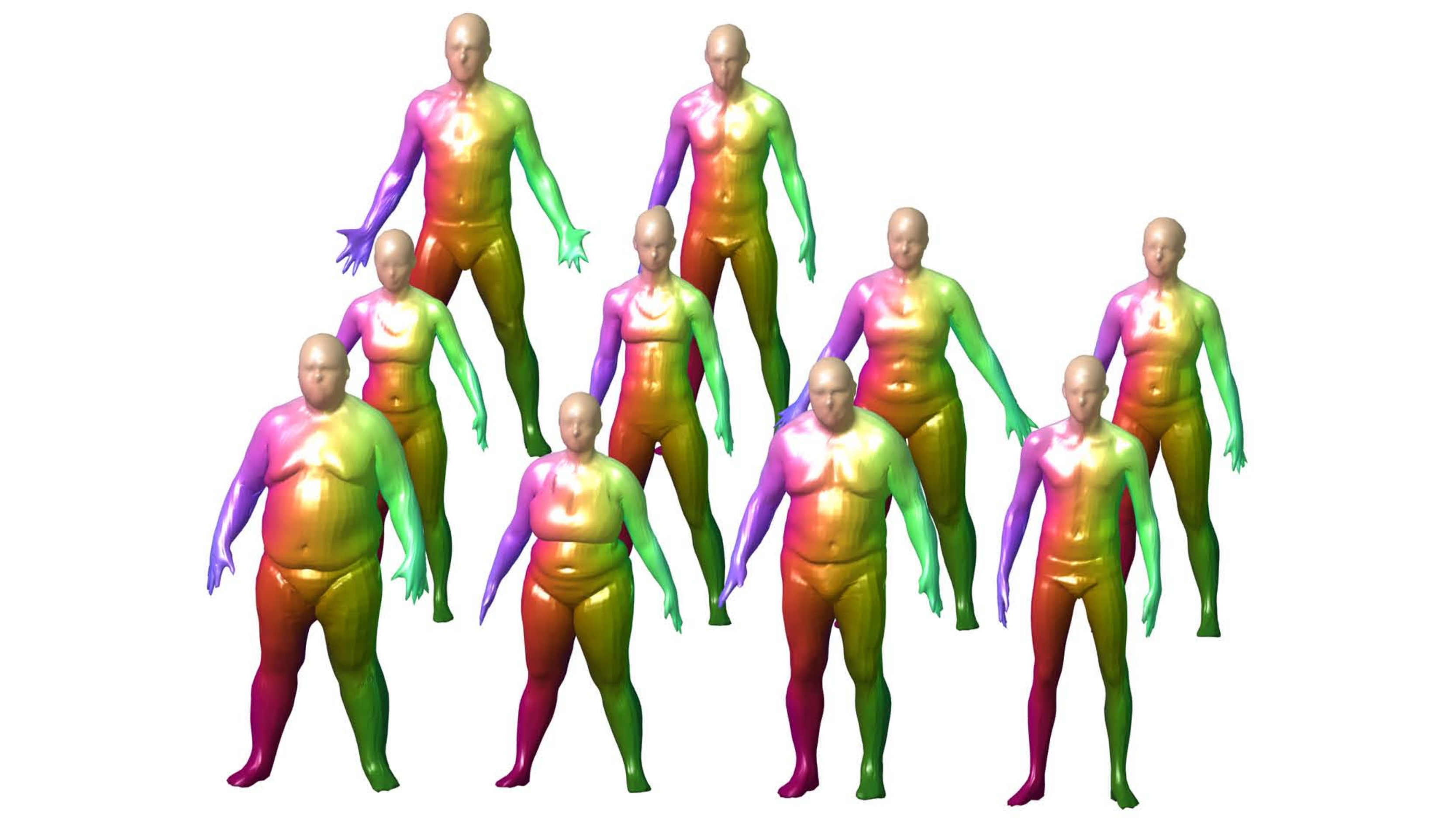}\\
		\small{(a) Input (before spatial registration).}&	
		 \small{(b) After spatial registration.}
	\end{tabular}
	 \caption{\label{fig:defaust_spatial_registrations} Illustration of the  spatial correspondences, on some samples from the DFAUST dataset~\cite{dfaust:CVPR:2017},  before and after applying the proposed spatial registration algorithm. Correspondences are color-coded, \ie points that are in correspondence are rendered with the same color.  }
\end{figure*}

\begin{figure*}[!h]
	 \begin{tabular}{@{}c@{}}
	 	\includegraphics[trim={0 140 0 140 },clip, width=\textwidth]{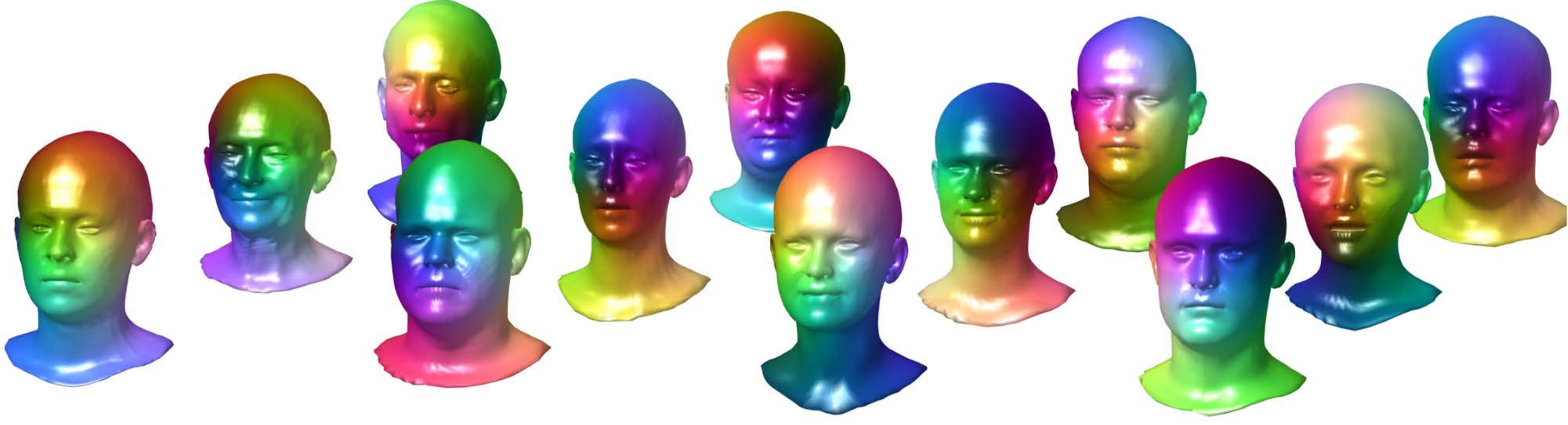}\\
		\small{(a) Input (before spatial registration).}\\
		\includegraphics[trim={0 140 0 140},clip, width=\textwidth]{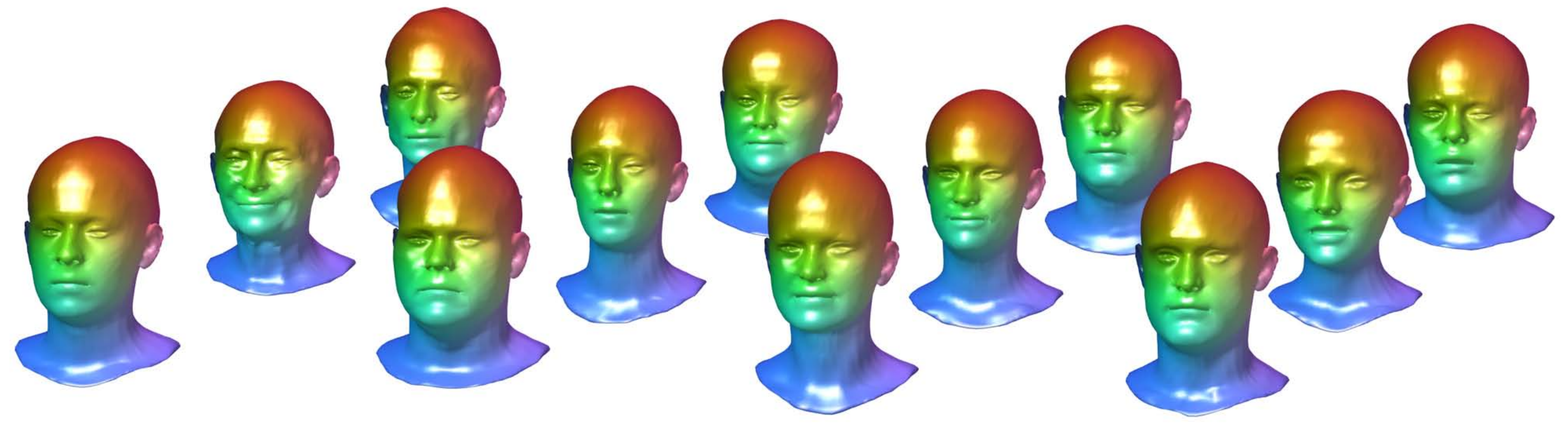}\\
		\small{(b) After spatial registration.}
	\end{tabular}
	 \caption{\label{fig:coma_spatial_registrations} Illustration of the  spatial correspondences, on samples from the COMA dataset~\cite{ranjan2018generating}, before and after applying the proposed spatial registration algorithm. Correspondences are color-coded, \ie points that are in correspondence are rendered with the same color.  }
\end{figure*}

\begin{figure*}[!ht]
	 \begin{tabular}{@{}c@{}}
	 	\includegraphics[width=\textwidth]{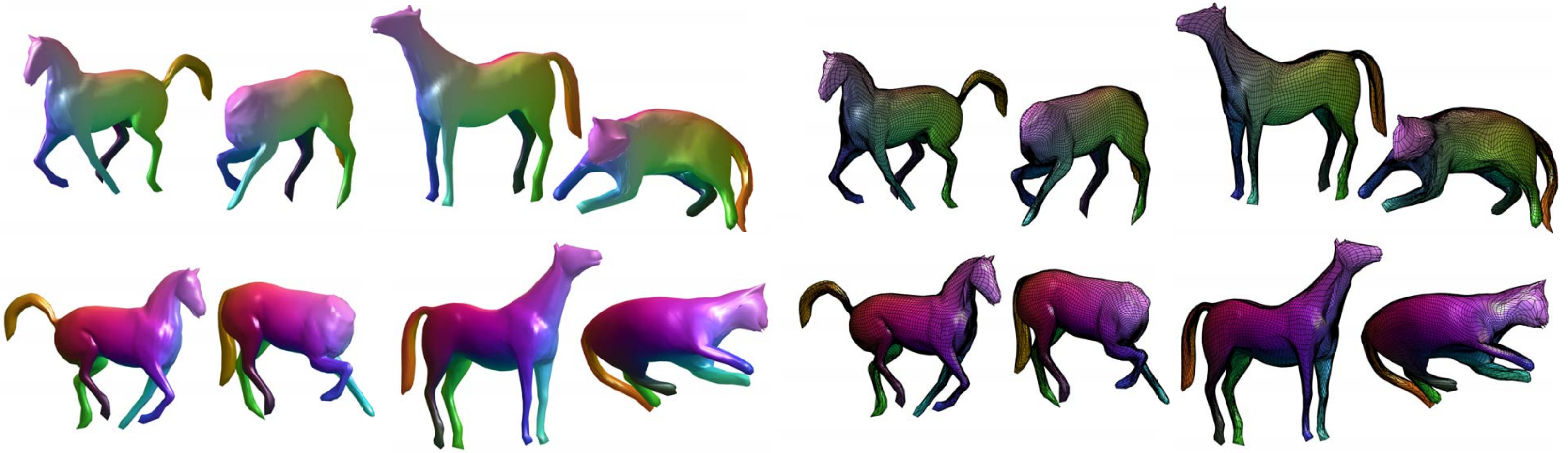}
	\end{tabular}
	 \caption{\label{fig:missing_parts_quadruples} Examples of  spatial correspondences between objects with missing parts. Correspondences are color-coded. The top row shows one view of the 3D models while the bottom row shows another view of the same models. In both cases, we show the smooth rendering as well as the wireframe rendering of the surfaces  so that the reader can appreciate the quality of the correspondences. }
\end{figure*}

\begin{figure}[!ht]
	 \begin{tabular}{@{}c@{}}
	 	\includegraphics[width=\linewidth]{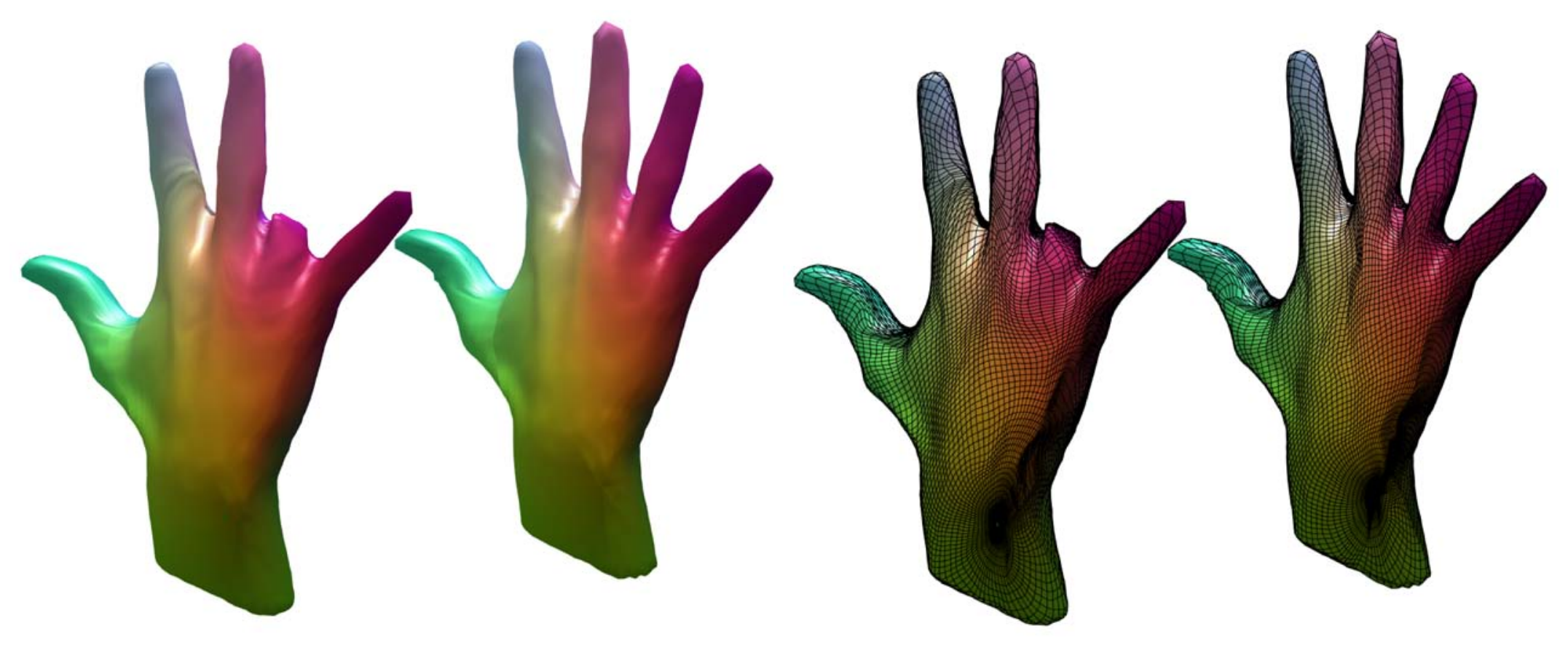}
	\end{tabular}
	 \caption{\label{fig:missing_parts_quadruples2} Examples of  spatial correspondences between 3D hands with one finger missing. Correspondences are color-coded, \ie points that are in correspondence are rendered with the same color.  We show the smooth rendering as well as the wireframe rendering so that the reader can appreciate the quality of the correspondences.}
\end{figure}

\subsection{Additional spatial registration results}
In this section, we provide more results to show the importance of each component of the proposed framework and to demonstrate the efficiency of the proposed spatial registration.

\vspace{6pt}
\noi\textbf{Importance of the SRNF representation.} Fig.~\ref{fig:supp_srnf_vs_nonsrnf1} illustrates, with a practical example, the importance of the SRNF representation for computing geodesics and statistics between 3D surfaces that undergo complex articulated and elastic motion. In this example, we use two human body shapes, perform their spatial registration using the proposed SRNF framework,  and then compute their mean using the $\ltwo$ metric  in the original space of surfaces (Fig.~\ref{fig:supp_srnf_vs_nonsrnf1}-(a)) and using the $\ltwo$ metric in the space of SRNFs (Fig.~\ref{fig:supp_srnf_vs_nonsrnf1}-(b)). In the former case, we can see that the parts that undergo large articulated motion (arms in this example) significantly shrink. This is predictible since a geodesic under the $\ltwo$ metric in the original space of surfaces corresponds to straight lines. In the SRNF space, the arms naturally bend since the $\ltwo$ metric in this space is equivalent to the optimal  bending and stretching of surfaces; see Fig.~\ref{fig:supp_srnf_vs_nonsrnf1}-(b).

\begin{figure*}[!ht]
	 \begin{tabular}{@{}c@{}}
	 	\includegraphics[width=\textwidth]{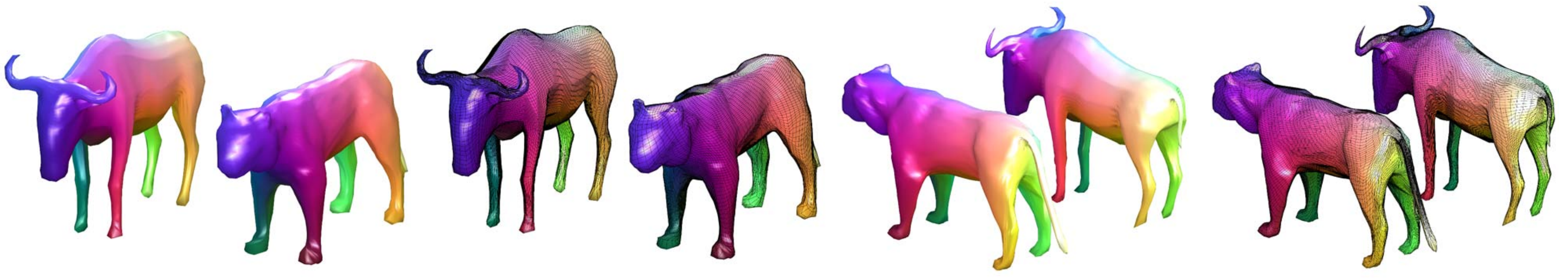}\\
		\small{(a) Correspondences using the SRNF framework.}\\
		
		\includegraphics[width=0.8\textwidth]{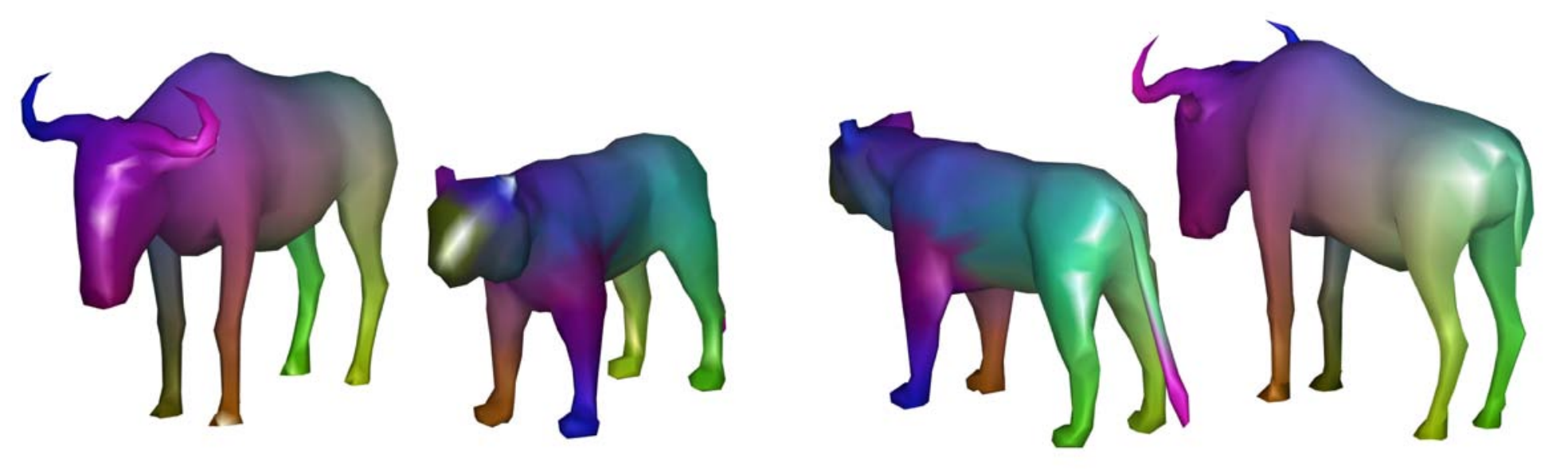}\\
		\small{(b) Correspondences using  MAPTree~\cite{ren2020maptree}.}\\
		
	\end{tabular}
	 \caption{\label{fig:animals_registrations2} Examples of  spatial correspondences between pairs of quadruped animals from the ACSM animal dataset~\cite{kulkarni2020articulation}. Correspondences are color-coded.  We also show in (a) the wireframe rendering of the surfaces.  Correspondences are color-coded, \ie points that are in correspondence are rendered with the same color. Correspondences are color-coded, \ie points that are in correspondence are rendered with the same color. }
\end{figure*}

\vspace{6pt}
\noi\textbf{Human bodies and faces.} Fig.~\ref{fig:cape_spatial_registrations} shows three dressed 3D human models, from the CAPE dataset, before  registration (Fig.~\ref{fig:cape_spatial_registrations}-(a)) and after registration (Fig.~\ref{fig:cape_spatial_registrations}-(b)) using the proposed SRNF-based approach. In both cases, the correspondences are color-coded. CAPE dataset is particularly challenging since it contains dressed 3D human bodies. Despite the significant differences in the initial parameterizations (see Fig.~\ref{fig:cape_spatial_registrations}-(a)) and in the shape of the bodies and clothes,  the proposed approach is able to find plausible correspondences between the surfaces (see Fig.~\ref{fig:cape_spatial_registrations}-(b)). Figs.~\ref{fig:defaust_spatial_registrations} and~\ref{fig:coma_spatial_registrations} show more examples using, respectively, the DFAUST human body dataset~\cite{dfaust:CVPR:2017} and the COMA human faces~\cite{ranjan2018generating}. In both cases, we show the surface before and after registration with correspondences color-coded.

\vspace{6pt}
\noi\textbf{Surfaces with missing parts.} Figs.~\ref{fig:missing_parts_quadruples} and~\ref{fig:missing_parts_quadruples2} show examples of spatial registrations between surfaces that have missing parts. Since our framework uses an elastic metric, which allows bending and stretching of surfaces, it is able to find one-to-one correspondences even under these challenging cases. Note that, similar to functional maps, the framework requires closed manifold surfaces and thus it is not able to handle (noisy) partial scans.

\begin{figure*}[!ht]
	 \begin{tabular}{@{}c@{}}
	 	\includegraphics[width=\textwidth]{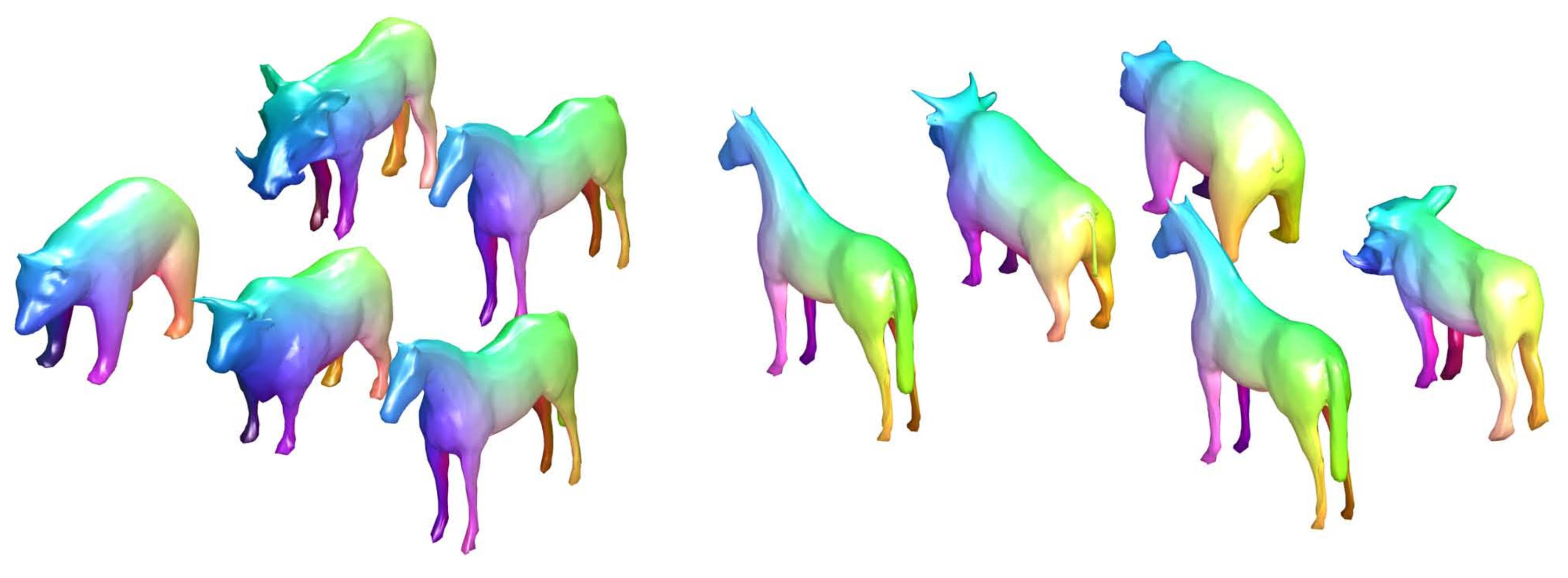}\\
		\small{View 1. \hspace{8cm} View 2.}\\
		\includegraphics[width=\textwidth]{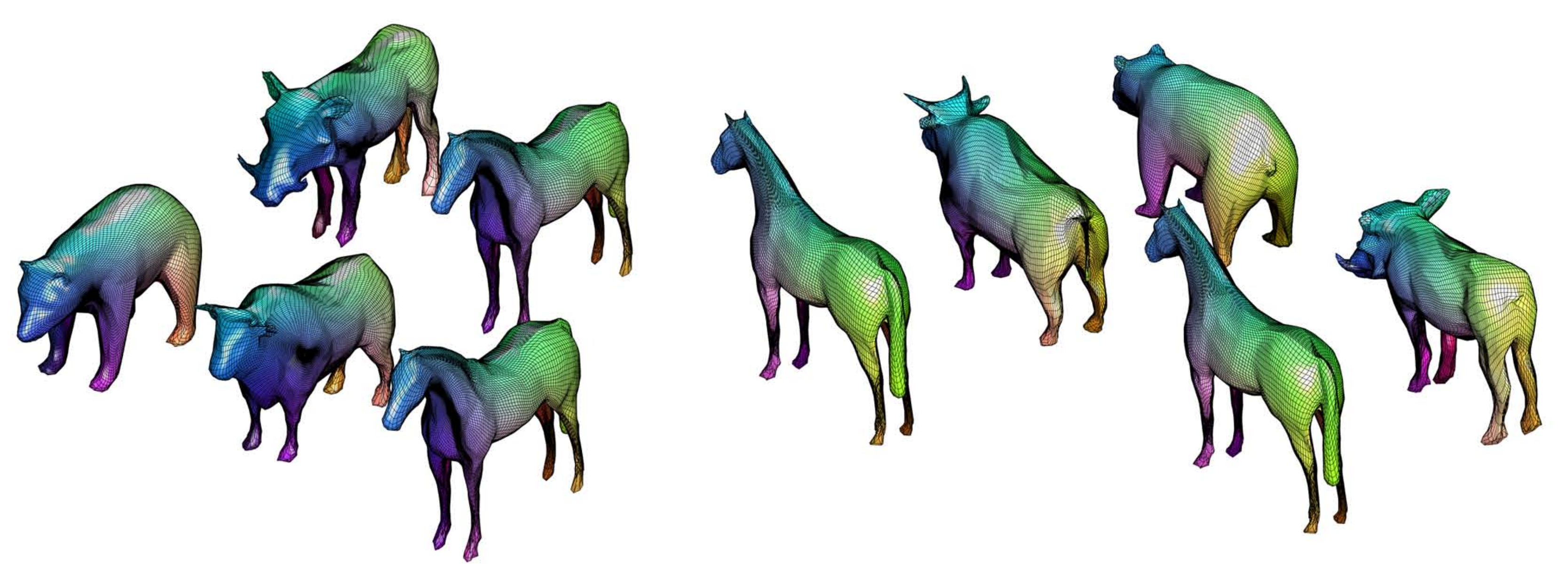}\\
		\small{View 1 with tessellation. \hspace{6cm} View 2 with tessellation.}\\
		\small{(a) Correspondences using the SRNF framework.}\\
		
		\includegraphics[width=\textwidth]{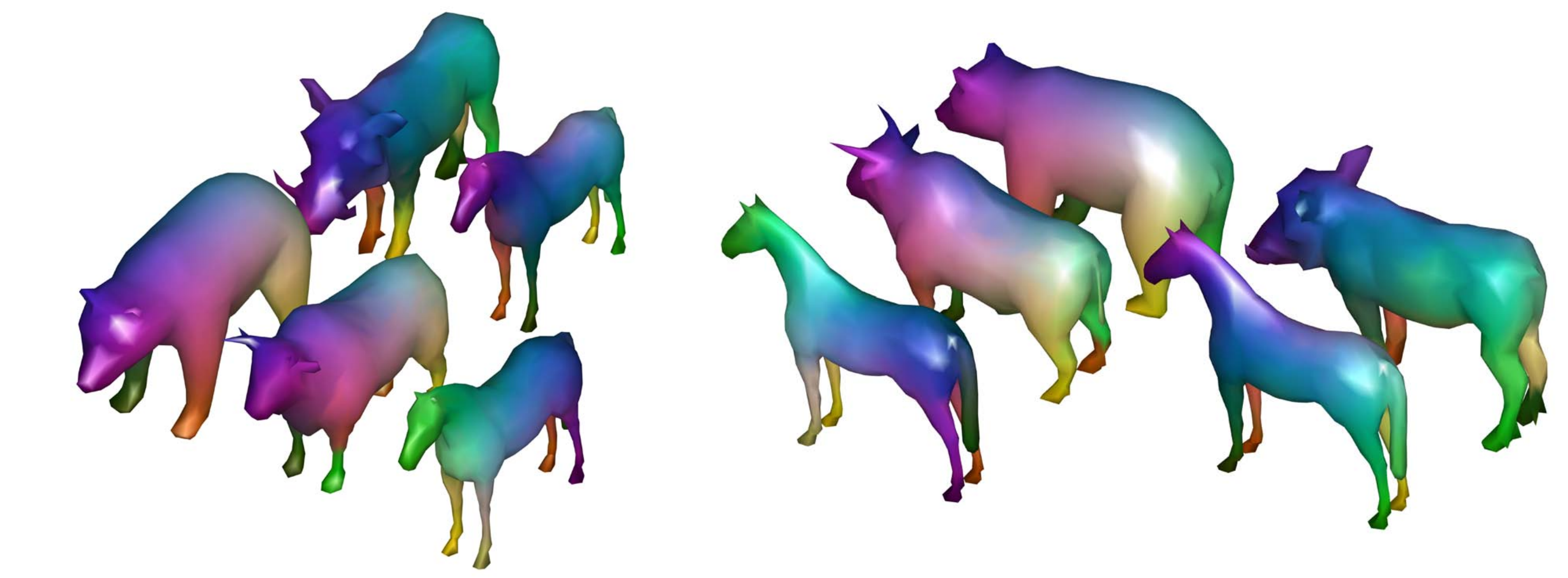}\\
		\small{View 1. \hspace{8cm} View 2.}\\
		\small{(b) Correspondences using  MAPTree~\cite{ren2020maptree}.}\\
	\end{tabular}
	 \caption{\label{fig:animals_registrations1} Examples of  spatial correspondences between quadruped animals from the ACSM animal dataset~\cite{kulkarni2020articulation}. Correspondences are color-coded. We also show in (a) the tessellation grid of the surfaces. Correspondences are color-coded, \ie points that are in correspondence are rendered with the same color. }
\end{figure*}

\begin{figure*}[!ht]
	 \begin{tabular}{@{}c@{}}
	 	\includegraphics[width=\textwidth]{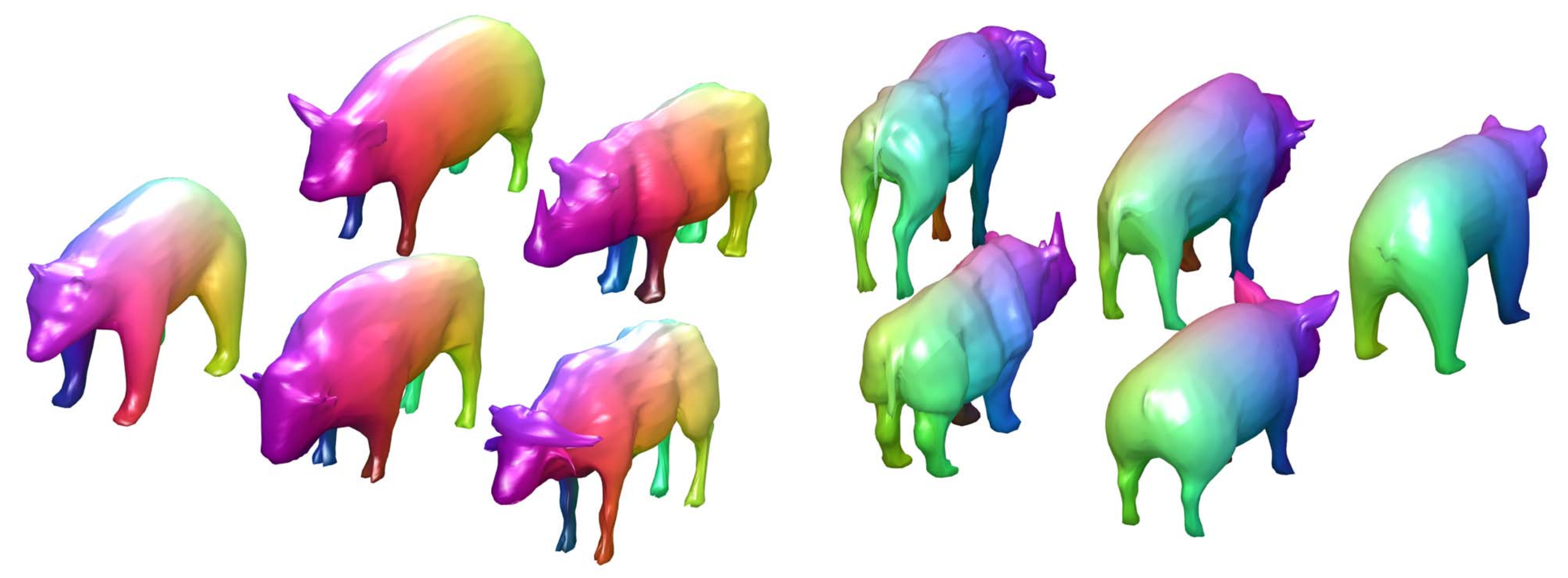}\\
		\small{View 1. \hspace{8cm} View 2.}\\
		\includegraphics[width=\textwidth]{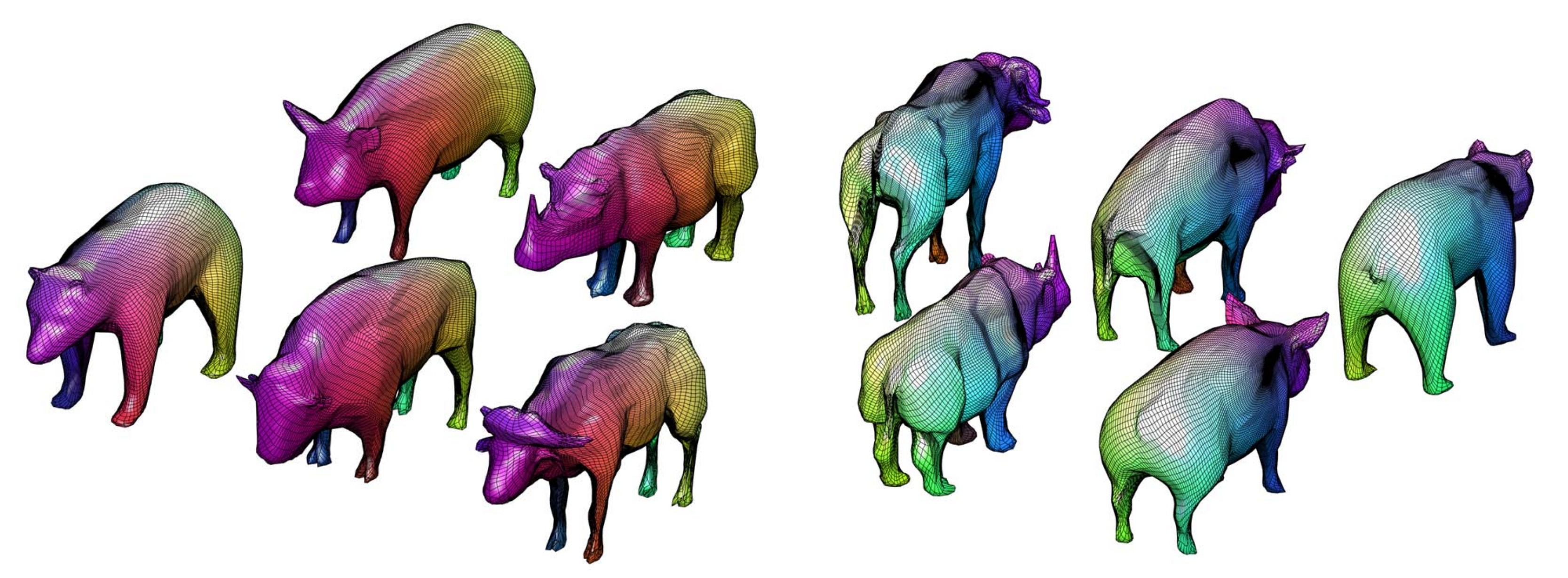}\\		
		\small{View 1 with tessellation. \hspace{6cm} View 2 with tessellation.}\\
		\small{(a) Correspondences using the SRNF framework.}\\
		
		\includegraphics[width=\textwidth]{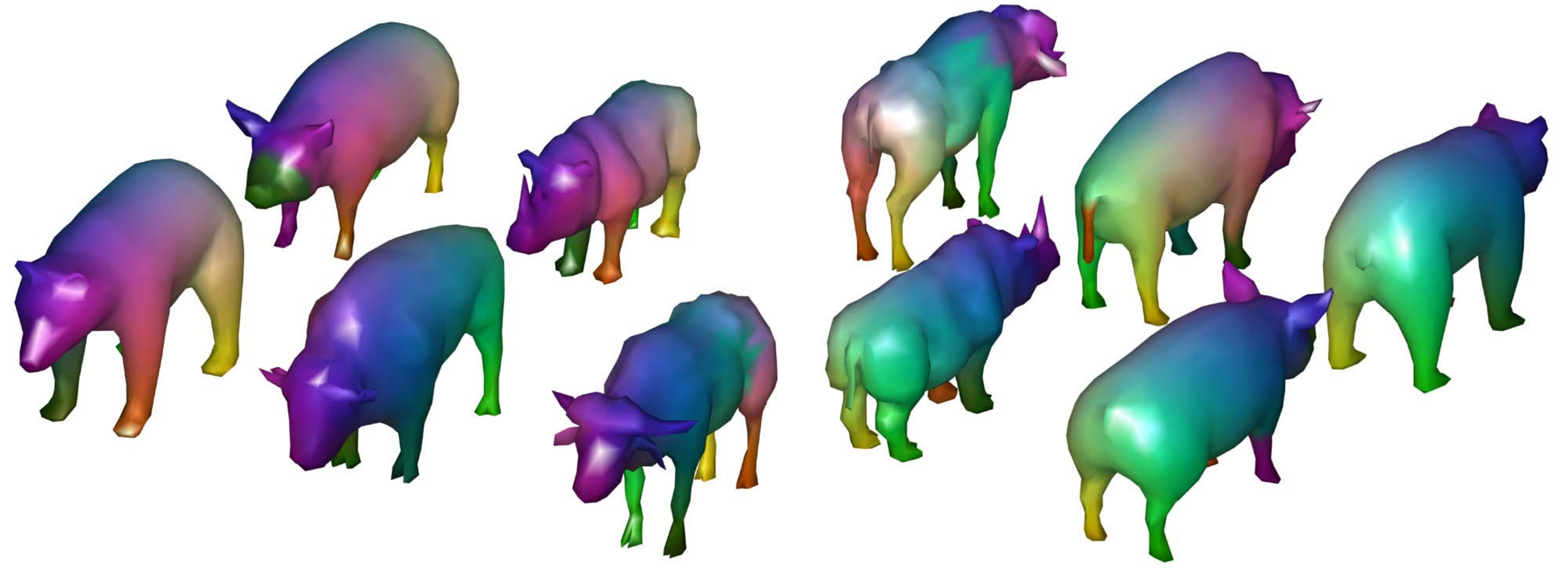}\\
		\small{View 1. \hspace{8cm} View 2.}\\
		\small{(b) Correspondences using  MAPTree~\cite{ren2020maptree}.}\\
	\end{tabular}
	 \caption{\label{fig:animals_registrations3} Examples of  spatial correspondences between quadruped animals from the ACSM animal dataset~\cite{kulkarni2020articulation}. Correspondences are color-coded. We also show in (a) the tessellation grid of the surfaces.  Correspondences are color-coded, \ie points that are in correspondence are rendered with the same color.}
\end{figure*}

\vspace{6pt}
\noi\textbf{Quadruped animals.} We test the proposed SRNF-based spatial registration on the quadruped animal dataset of~\cite{kulkarni2020articulation}. Figs.~\ref{fig:animals_registrations2}-(a),~\ref{fig:animals_registrations1}-(a), and~\ref{fig:animals_registrations3}-(a) show the correspondence results obtained using our approach (correspondences are color-coded). We also compare, visually, our results to the registrations obtained using the state-of-the-art functional map-based methods such as MapTree~\cite{ren2020maptree}; see  Figs.~\ref{fig:animals_registrations1}-(b),~\ref{fig:animals_registrations2}-(b), and~\ref{fig:animals_registrations3}-(b). As one can see, the mappings obtained using MapTree are often not correct in many regions of the shapes. More importantly, the maps obtained using MapTree are not  one-to-one since a point on the source shape can be mapped to multiple points on the target. Our approach finds one-to-one correspondences and thus is more suitable for  computing geodesics and shape statistics.

\subsection{Karcher mean of surfaces}
Algorithm~\ref{alg:karcher_mean} summarizes the Karcher mean algorithm used to compute the mean 4D surface of a collections of 4D surfaces; see Section~\ref{sec:4Dstatistics} of the main manuscript. We refer the reader to the main manuscript for the definition of each symbol.

\begin{algorithm}[!h]
\textit{Input:} A set of spatially-registered 4D surfaces $\{\curve_1,\dots, \curve_n\}\in{\curves}$ and their corresponding SRNF trajectories $\{\srnfcurve_1,\dots,\srnfcurve_n\}\in \srnfcurves$.\\
\noi\textit {Output:} Karcher mean $\meancurve \in \curves$.
 
 \begin{algorithmic}[1]
	\STATE Compute the TSRVF maps: $ \srvf_i = \srvfmap(\srnfcurve_i), i=1, \dots, n$. 
 	\STATE Set $\meansrvf \leftarrow \srvf_1$ as an initial estimate of the Karcher mean. 
	\STATE Set $\curvediffeo_i,\ i=1, \dots,n$ to the identity $\curvediffeo_{id}(t)=t$.
	\STATE \label{alg:stepKarcher}For $i=1,\dots,n$, 
		\begin{itemize}
			\item Temporally register $\srvf_i$ to $\meansrvf$, resulting in $\srvf^*_i= \srvf_i \odot \curvediffeo_i^*$, by solving Eqn~\eqref{eq:optimalTemporalRegistration} of the main manuscript. 
			\item $\srvf_i \leftarrow \srvf^*_i$ and $\curvediffeo_i \leftarrow \curvediffeo_i \circ \curvediffeo_i^*$. 
		\end{itemize}
    \STATE Update the Karcher mean $\meansrvf = \frac{1}{n}\sum_{i=1}^n \srvf_i$.
    \STATE If the change in $\left\| \meansrvf \right\|$ is large then go to Step~\ref{alg:stepKarcher}.
    \STATE Find $\meancurve$ by TSRVF inversion of $\meansrvf$ followed by SRNF inversion. 
    \STATE Return $\meancurve$, $\curvediffeo_i, q_i, \i=1, \dots, n$. 
\end{algorithmic}
\caption{\label{alg:karcher_mean} Karcher mean of 4D surfaces.}
\end{algorithm}

\clearpage
\section{Temporal registration}
\label{sec:supp_temporal_registration}
In this appendix, we provide additional results and additional quantitative evaluations  to further demonstrate the quality and performance of the  temporal registration framework proposed in this paper.

\begin{figure}[t]
	\begin{tabular}{@{}c@{}c@{}}
		\includegraphics[trim={90 50 90 50}, clip, width=0.24\textwidth]{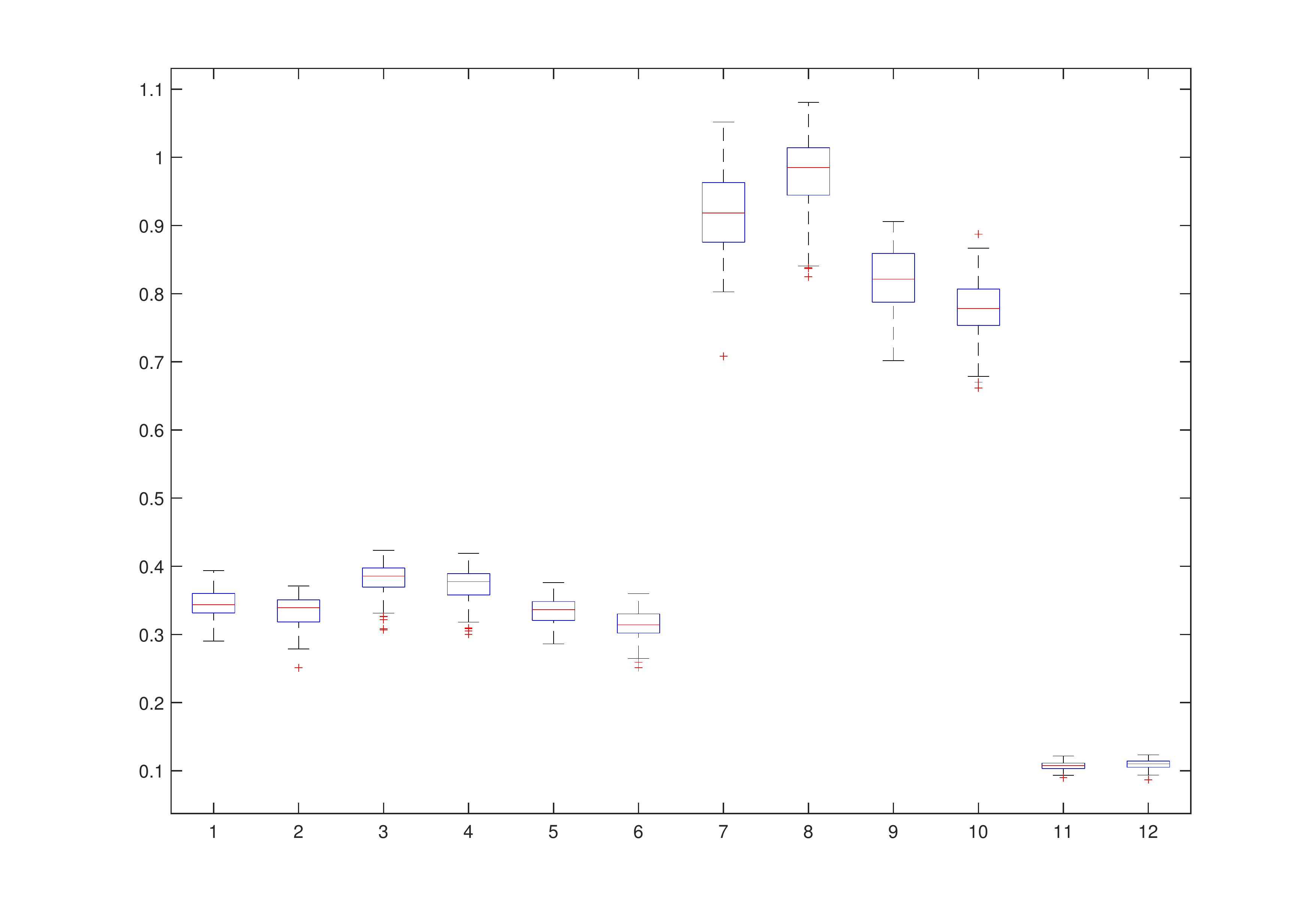} & 
		\includegraphics[trim={90 50 90 50}, clip,width=0.24\textwidth]{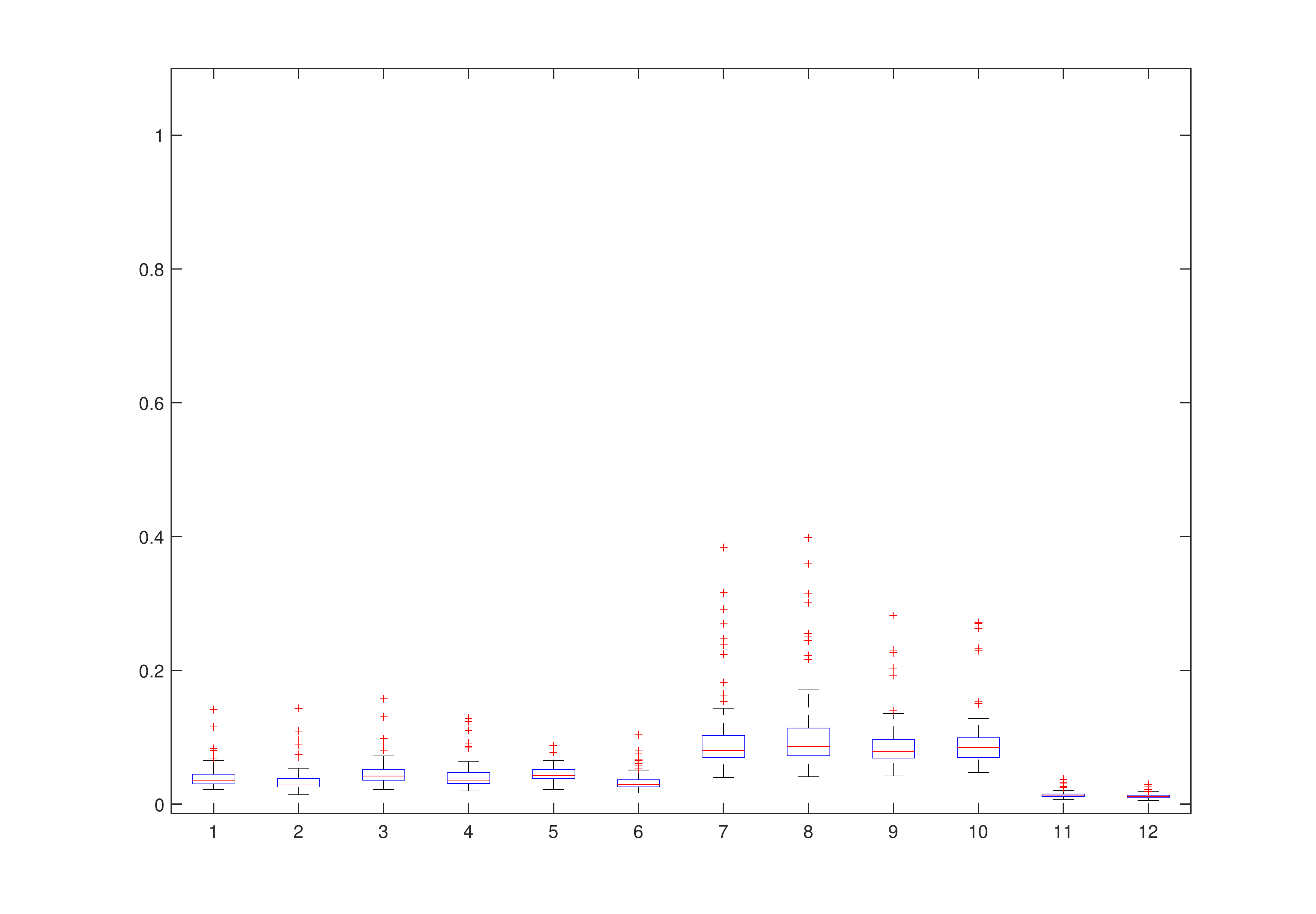} \\
		\small{(a) Before registration.} & \small{(b) After registration.}
	\end{tabular}
	\caption{\label{fig:suppmat_errors}  We consider twelve 4D surfaces from the VOCA dataset, perturb each of them using $100$ random temporal diffeomorphisms, and then measure the distance between each original 4D surface and its perturbed versions. We show the boxplots of distances  \textbf{(a)} before the temporal registration and \textbf{(b)} after the temporal registration using the proposed framework for each of the twelve 4D surfaces. In each plot, the X axis corresponds to the 4D surfaces and the Y axis to the temporal alignment  error. The lower the error is the better is the alignment. }
\end{figure}

\begin{figure}[t]
	\begin{tabular}{@{}c@{}c@{}}
		\includegraphics[trim={40 1 40 1}, clip, width=0.24\textwidth]{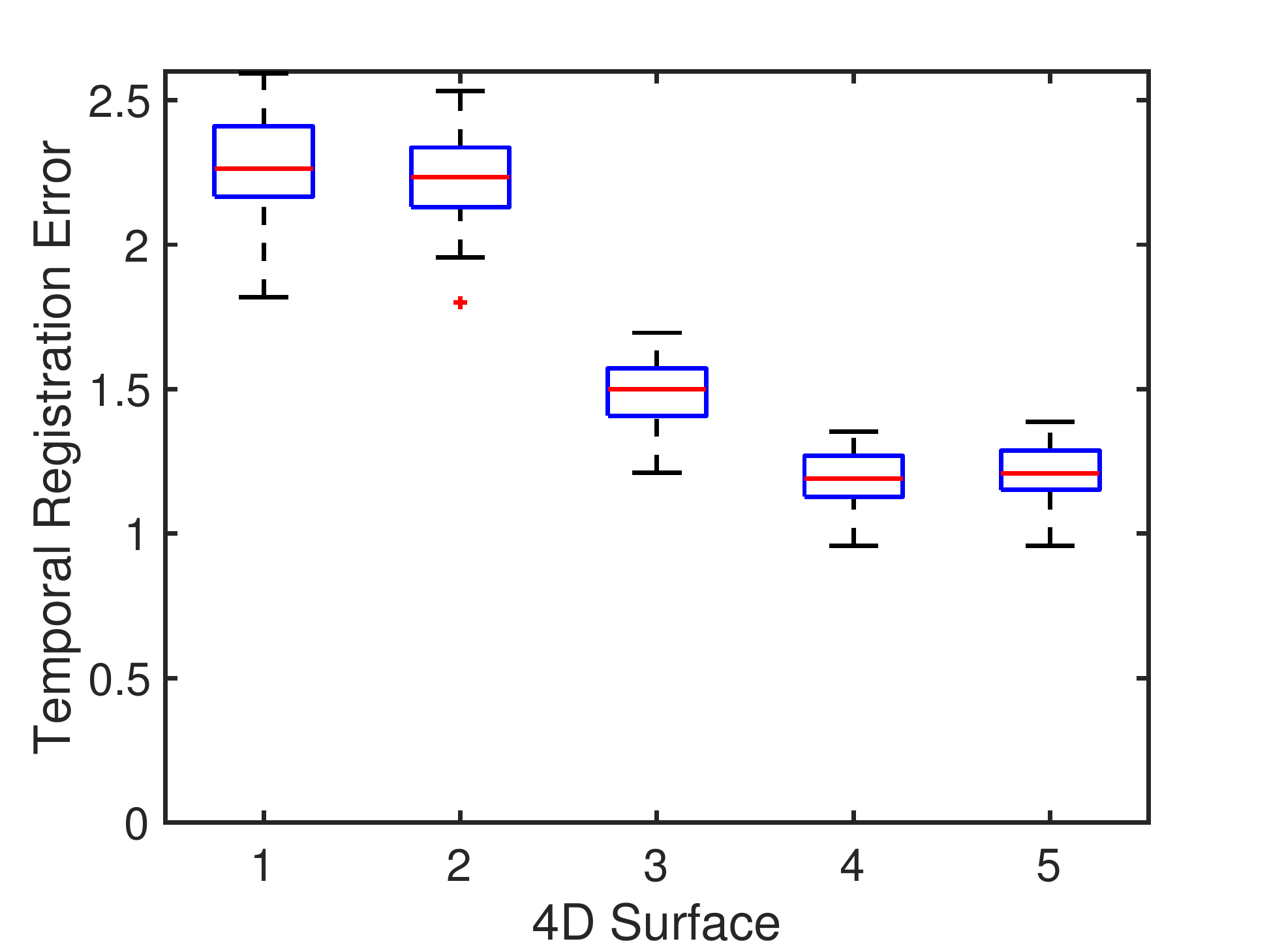} & 
		\includegraphics[trim={40 1 40 1}, clip, width=0.24\textwidth]{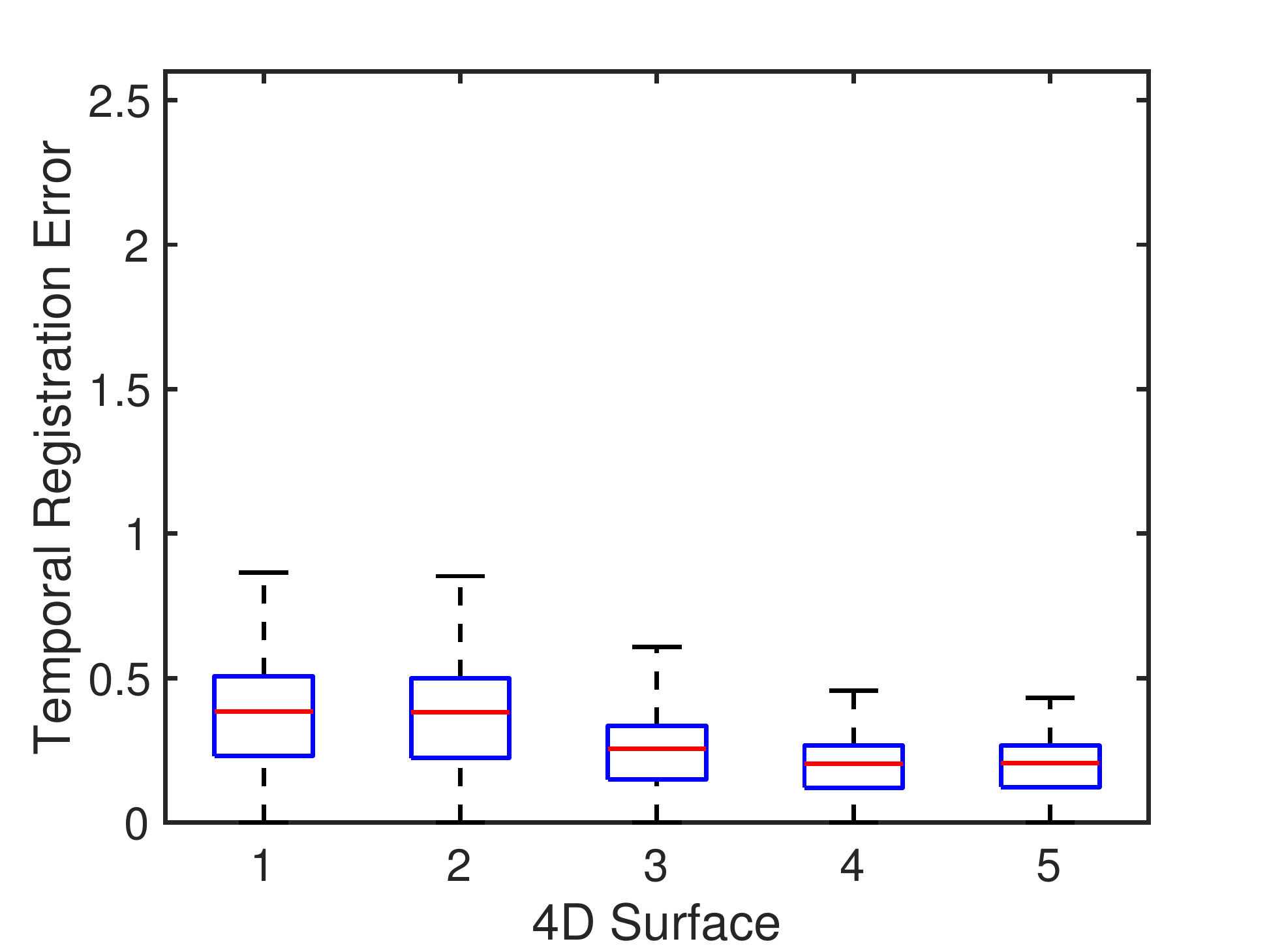} \\
		\small{(a) Before registration.} & \small{(b) After registration.}
	\end{tabular}
	\caption{\label{fig:suppmat_errors2} We consider five 4D surfaces generated using FLAME framework, perturb each of them using $100$ random temporal diffeomorphisms, and then measure the distance between each original 4D surface and its perturbed versions. We show the boxplots of distances  \textbf{(a)} before the temporal registration and \textbf{(b)} after the temporal registration using the proposed framework for each of the five 4D surfaces. In each plot, the X axis corresponds to the 4D surfaces and the Y axis to the temporal alignment  error. The lower the error is the better is the alignment. }
\end{figure}

\subsection{Evaluation of the temporal registration}

To quantitatively evaluate the performance of the proposed temporal registration, we first consider a 4D surface $\curve$ and parameterize it with randomly generated temporal diffeomorphims $\curvediffeo_i, i\in \{1, \cdots, 100\}$, to obtain new 4D surfaces $\curve_i = \curve \circ \curvediffeo_i$. Next, we compute the distances $d_i,\ i=1,\dots,100$, between $\curve$ and $\curve_i$, using the expression given in  Eqn.~\eqref{eq:TSRVFltwo} of the main manuscript, before temporal registration. Fig.~\ref{fig:suppmat_errors}-(a) shows the box plots of the resulting distances for $12$ sequences from the VOCA dataset before the temporal registration. Ideally, $\forall i,\ d_i = 0$. However, since the 4D surfaces are not temporally registered, the distances are large in most of the cases. 

In Fig.~\ref{fig:suppmat_errors2}, we perform the same experiment but this time on five 4D surfaces simulated using the FLAME framework; see Section~\ref{sec:results_registration} of the main manuscript for a detailed description of how these 4D surfaces have been generated. Fig.~\ref{fig:suppmat_errors2} reports the temporal registration errors before and after applying the proposed framework for each of the five 4D surfaces. As one can see, our approach is able to properly align the perturbed 4D surfaces to their original 4D surfaces. 

Next, we compute the optimal temporal registration, \ie for every 4D surface $\curve_i$, we find the optimal diffeomorphism $\curvediffeo^*_i$ that  aligns $\curve_i$  to $\curve$. Let $\curve^*_i = \curve_i \circ  \curvediffeo^*_i$.  Fig.~\ref{fig:suppmat_errors2}-(b) shows boxplots of the distances between $\curve^*_i$ and $\curve$ for the same five sequences. Compared to the distances between the original unregistered 4D expressions shown in Fig.~\ref{fig:suppmat_errors2}-(a), these distances are significantly lower. This shows that the proposed temporal alignment framework brings the 4D surfaces as close as possible to each other.

\begin{figure}[t]
	\begin{tabular}{@{}c@{}c@{}}
		\includegraphics[trim={85 1 85 1}, clip, width=0.24\textwidth]{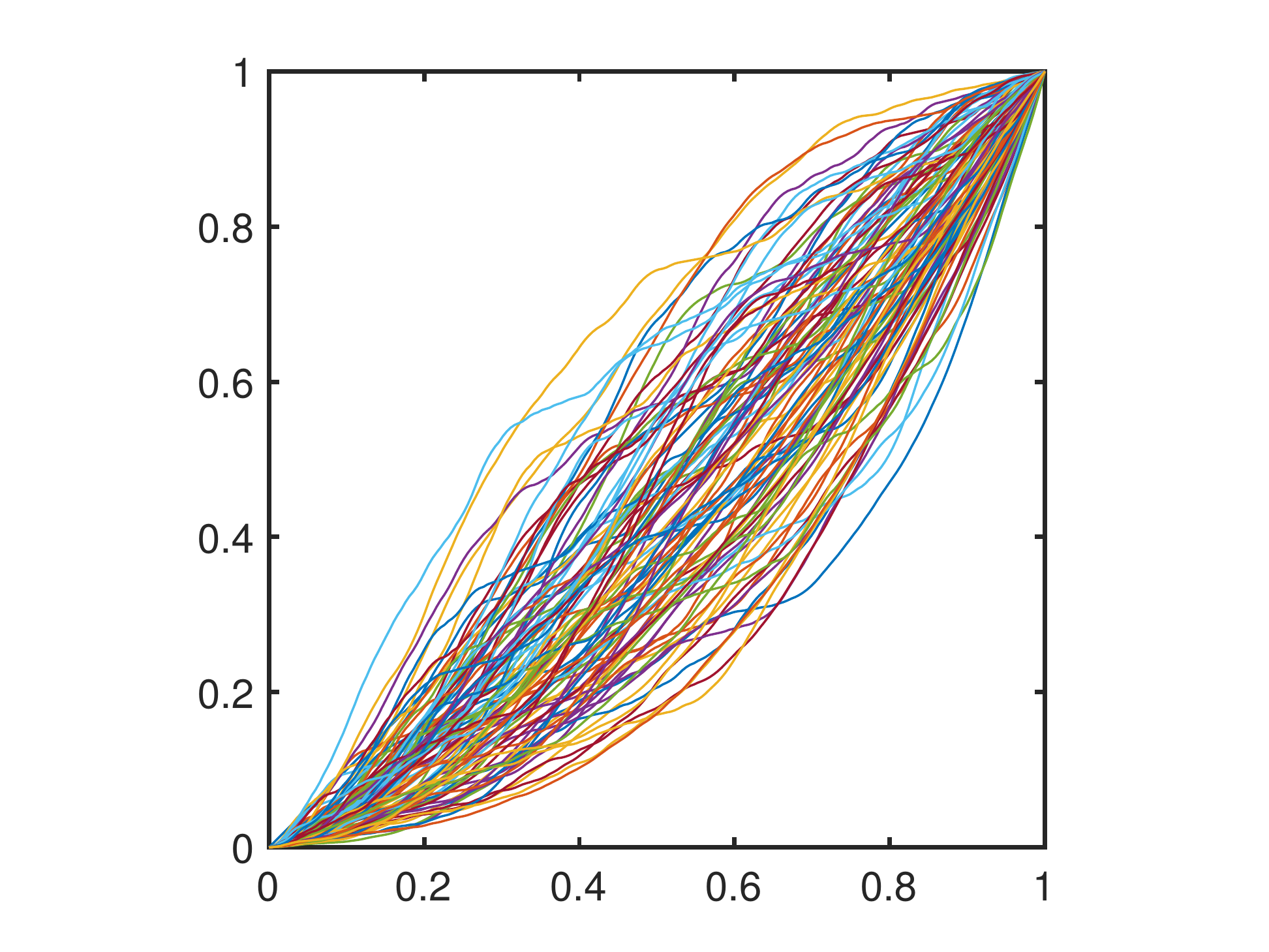} &
		\includegraphics[trim={85 1 85 1}, clip, width=0.24\textwidth]{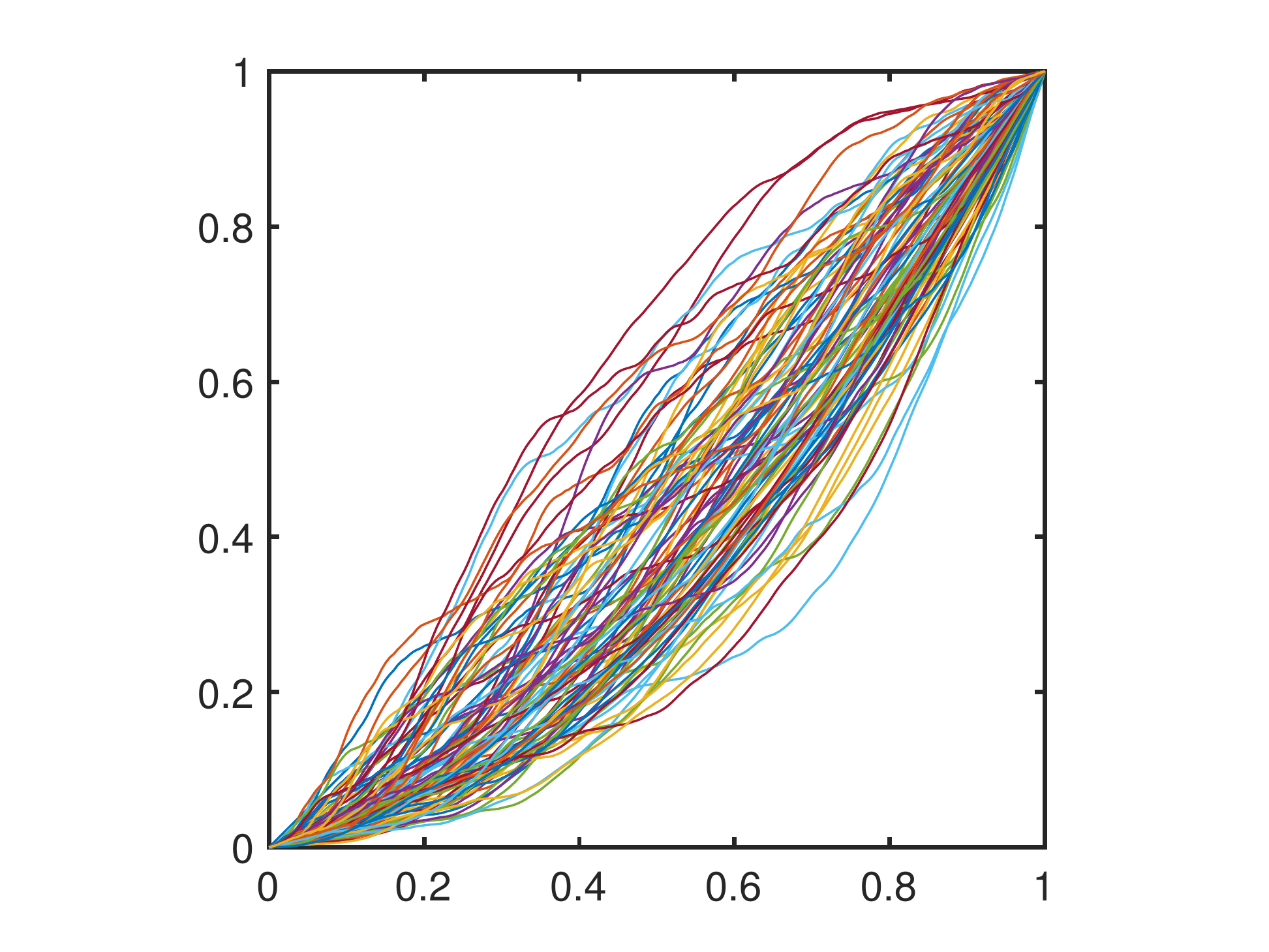} \\
		\small{(a)} & \small{(b)}
	
	\end{tabular}
	\caption{\label{fig:suppmat_diffeos} The 100 random temporal diffeomorphisms that have been applied to the original 4D surfaces to simulate the data \textbf{(a)} in Figs.~\ref{fig:suppmat_errors} and~\ref{fig:suppmat_errors2} of this supplementary material, and  \textbf{(b)} in Fig.~\ref{fig:errors_flame-fitting} of the main manuscript.  In each plot, the X and U axis both correspond to the temporal domain $[0, 1]$ since,  in our formulation, a temporal diffeomorphism $\curvediffeo$ is defined as the mapping of the temporal domain $[0, 1]$ to itself, \ie $\curvediffeo: [0, 1] \to [0, 1] \text{ such that } 0<\frac{d\curvediffeo}{dt}<\infty, \curvediffeo(0) = 0 \text{ and } \curvediffeo(1) = 1.$  }
\end{figure}

\begin{figure*}
		\includegraphics[trim={15 0 12 0},clip, width=\textwidth]{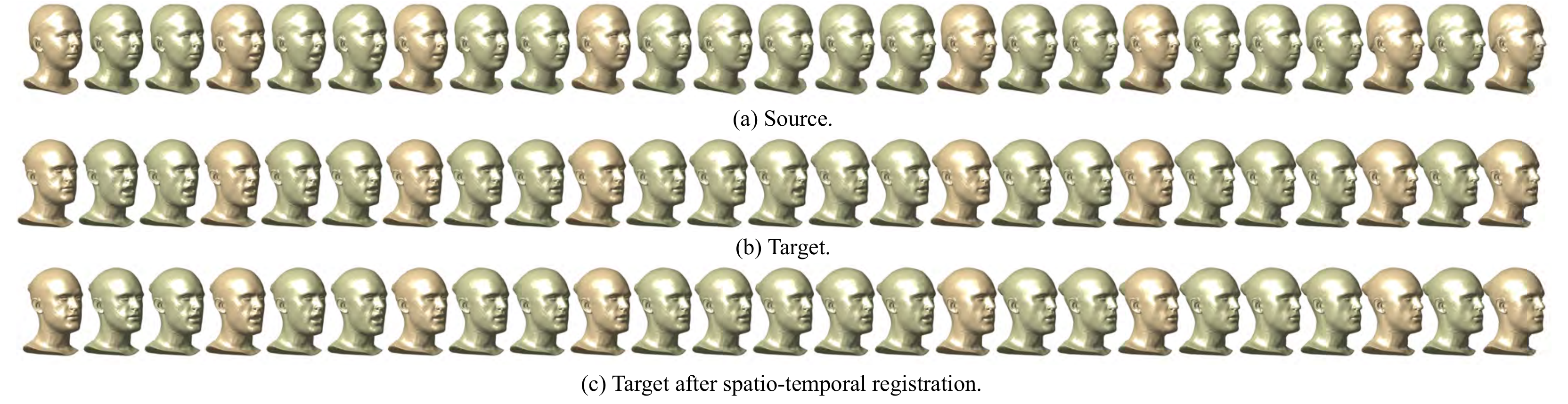} 
		 \caption{\label{fig:suppmat_faces_spatio_temporal_faces_1} Examples of the spatiotemporal registration of two facial expressions. Note how the spatiotemporally registered target 4D surface in (c) became fully synchronised with the source 4D surface (a).   The full video sequence is provided in the supplementary material.} 
\end{figure*}

Finally, Figs.~\ref{fig:suppmat_diffeos}-(a) and (b) show the $100$ random temporal diffeomorphisms that have been applied to perturb the 4D surfaces used for quantitative evaluation shown in  Fig.~\ref{fig:errors_flame-fitting} of the main manuscript, and in Figs.~\ref{fig:suppmat_errors} and~\ref{fig:suppmat_errors2} in this Supplementary Material.

\subsection{Additional temporal registration results}
Fig.~\ref{fig:suppmat_faces_spatio_temporal_faces_1} shows another example of the spatiotemporal registration of two 4D facial surfaces. In this example, we show the source surface, the target 4D surface before the spatiotemporal registration, and the target 4D surface after spatiotemporal registration. We also highlight some key frames to illustrate the quality of the temporal registration.

\section{Geodesics between 4D surfaces}
\label{sec:supp_geodesics}
\begin{figure*}[t]
\center{
\begin{tabular}{@{}c@{}}
	\midrule
	\includegraphics[width=\textwidth]{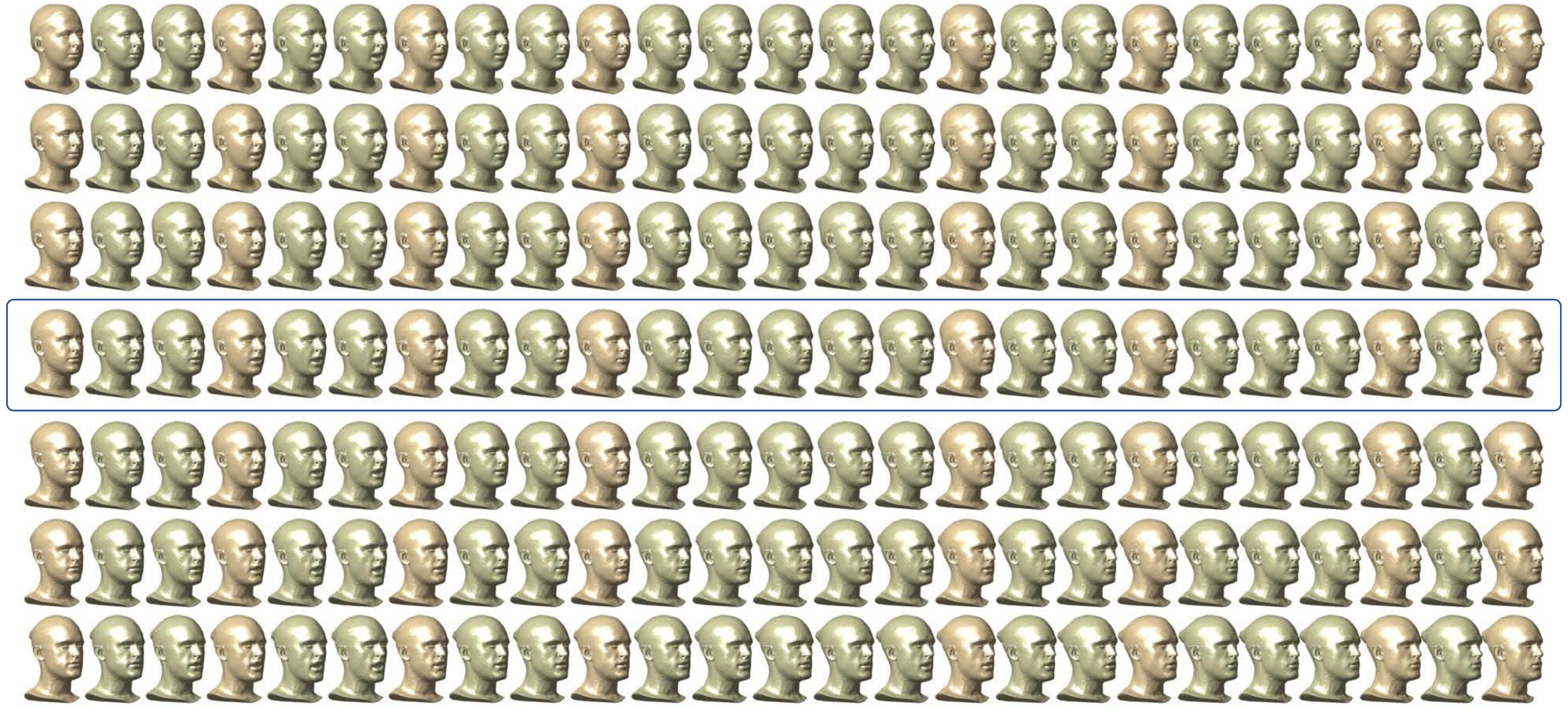} \\
	(a) Example 1.\\
	\midrule
	\includegraphics[width=\textwidth]{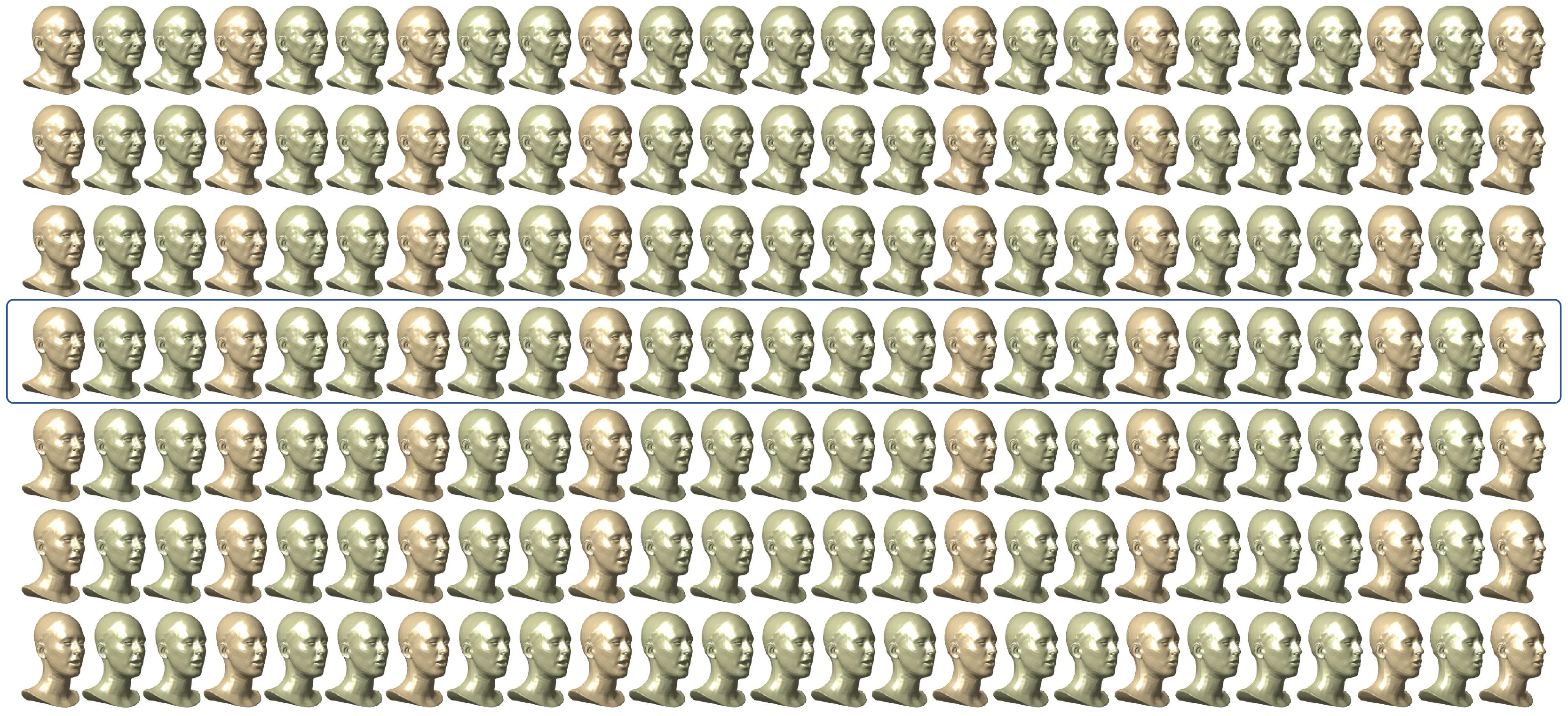}\\
	(b) Example 2.\\
	\midrule
\end{tabular}
	\caption{\label{fig:suppmat_4Dgedeosics_faces2} Two additional examples of geodesics between 4D surfaces.  In each example, we show the source 4D surface, the target 4D surface after spatiotemporal registration, and five intermediate 4D surfaces along the geodesic between the source and the target. The highlight 4D surface corresponds to the mean. A video illustrating this sequence is included in the supplementary material.}
}
\end{figure*}

\begin{figure*}[t]
	\includegraphics[width=\textwidth]{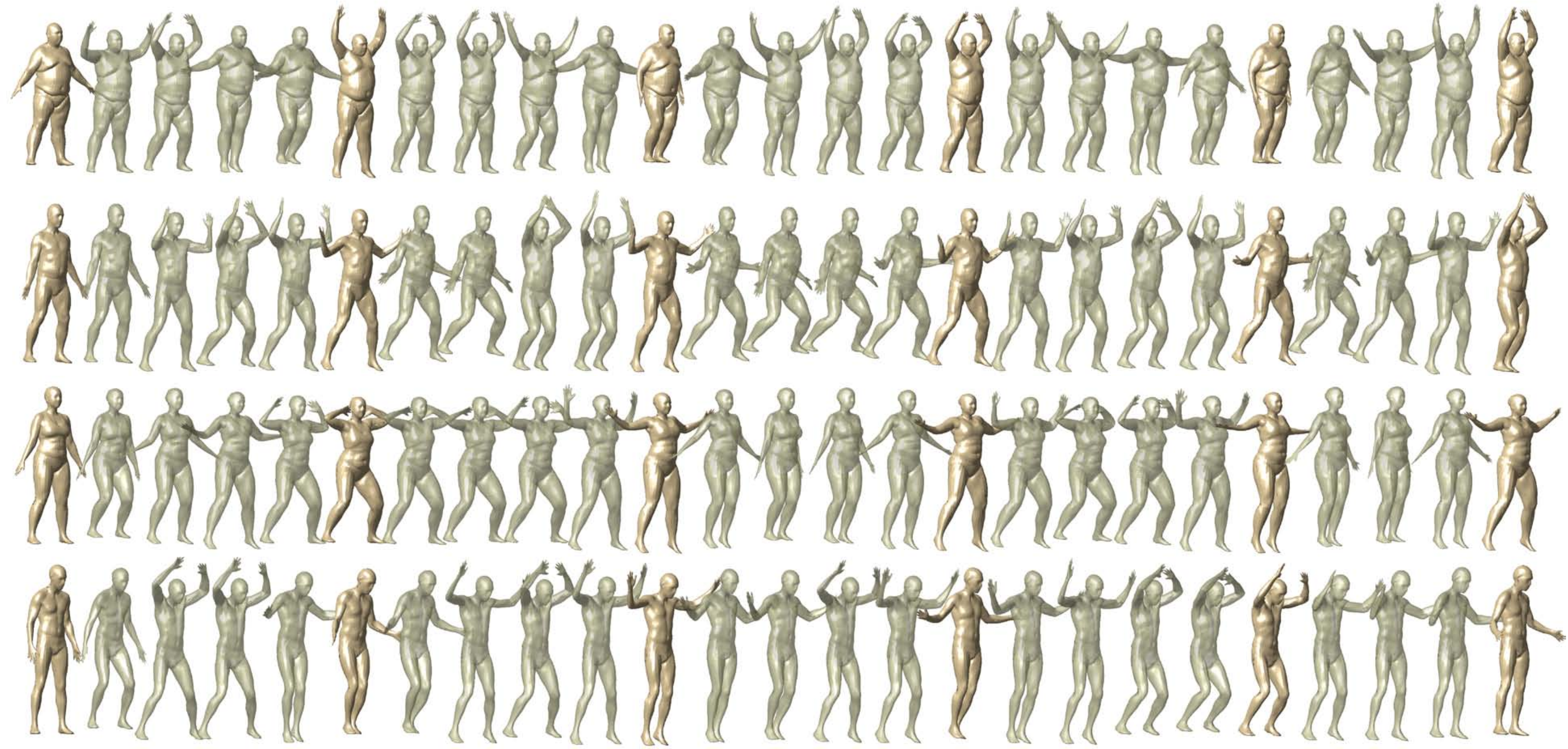} 
	\caption{\label{fig:suppmat_means_humans2_input}  Input 4D surfaces, prior to the spatiotemporal registration, used to generate the mean 4D surface of Fig.~\ref{fig:means_humans2} of the main manuscript. Each row correspond to one 4D surface.  The supplementary material also includes the full video sequences. }
\end{figure*}

This appendix  presents additional results on the computation of geodesics between 4D surfaces.   Fig.~\ref{fig:suppmat_4Dgedeosics_faces2} shows two additional examples of gedoesics between 4D facial surfaces.  Fig.~\ref{fig:suppmat_means_humans2_input}, on the other hand, shows the input 4D surfaces, prior to the spatiotemporal registration, used to generate the mean 4D surface in Fig.~\ref{fig:means_humans2} in the main manuscript. 


\section{Statistics between 4D surfaces}
\label{sec:supp_statistics}
\begin{figure*}[t]
	\begin{tabular}{@{}c@{}}		
		\includegraphics[width=\textwidth]{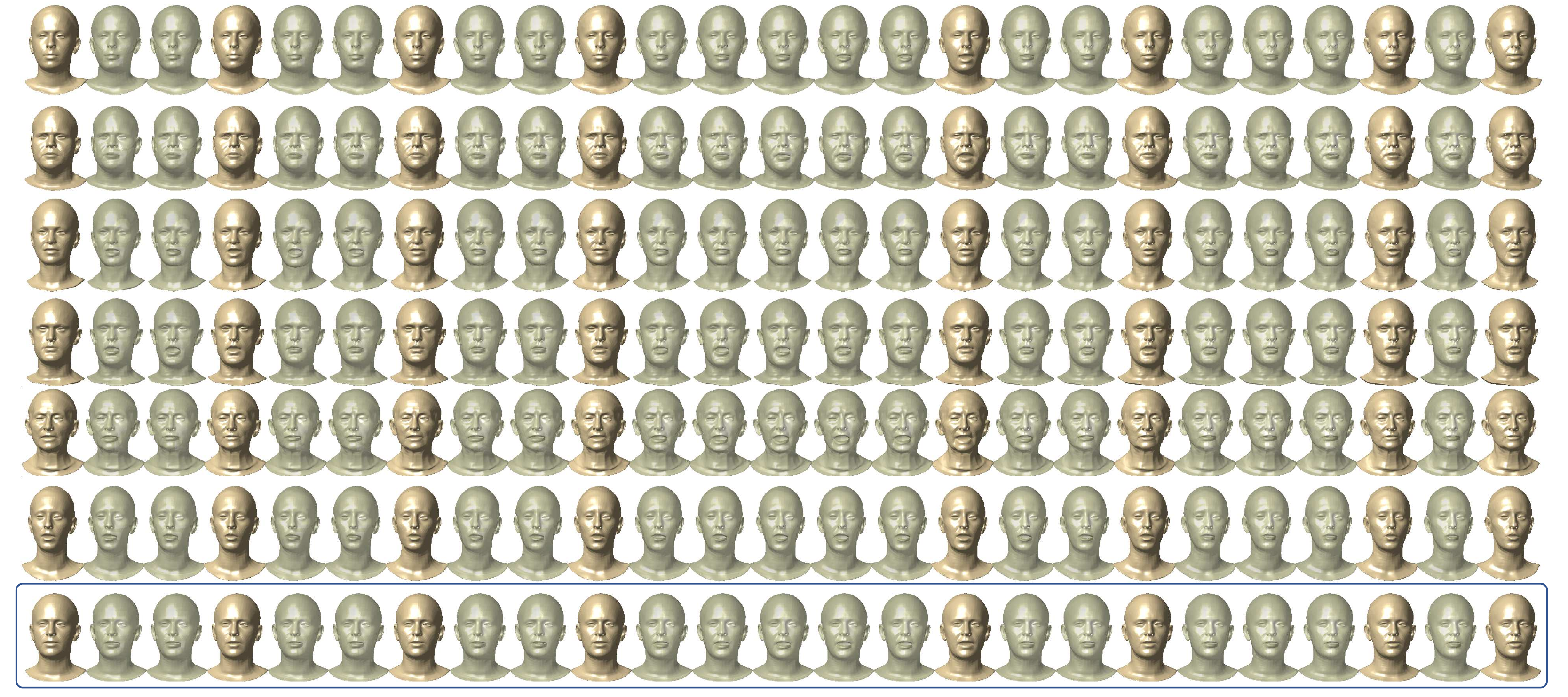}\\
		\small{(a) Input unregistered 4D surfaces. The last row is the mean obtained without temporal registration. }\\
		
		\includegraphics[width=\textwidth]{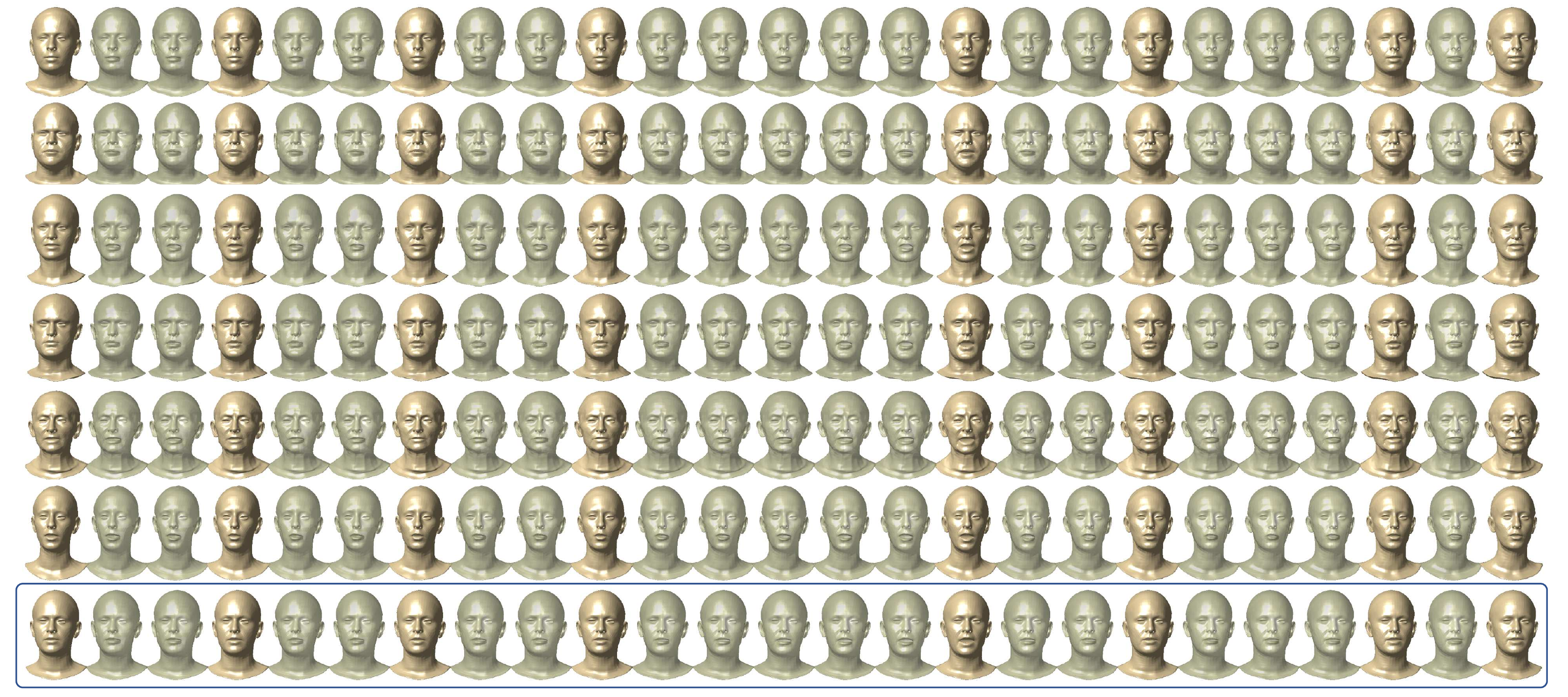}\\
		
		 \small{(b) Spatio-temporally co-registered 4D surfaces. The last row is the mean obtained  after co-registration} \\
		 \small{The mean is shown in the last column.} 		
	\end{tabular}
	
	\caption{\label{fig:suppmat_means_faces2}  Input 4D faces and mean 4D surface computed before (a) and after (b) co-registration.  Each row corresponds to one 4D surfaces. The supplementary material includes the full video sequences.  }
\end{figure*}

\begin{figure*}[t]
	\begin{tabular}{@{}c@{}}
		\includegraphics[width=0.95\textwidth]{voca_6samples_mode1.pdf}\\
		\small{(a) First mode of variation. }\\
		
		\includegraphics[width=0.95\textwidth]{voca_6samples_mode2.pdf}\\
		\small{(b) Second mode of variation. }\\
		
		\includegraphics[width=0.95\textwidth]{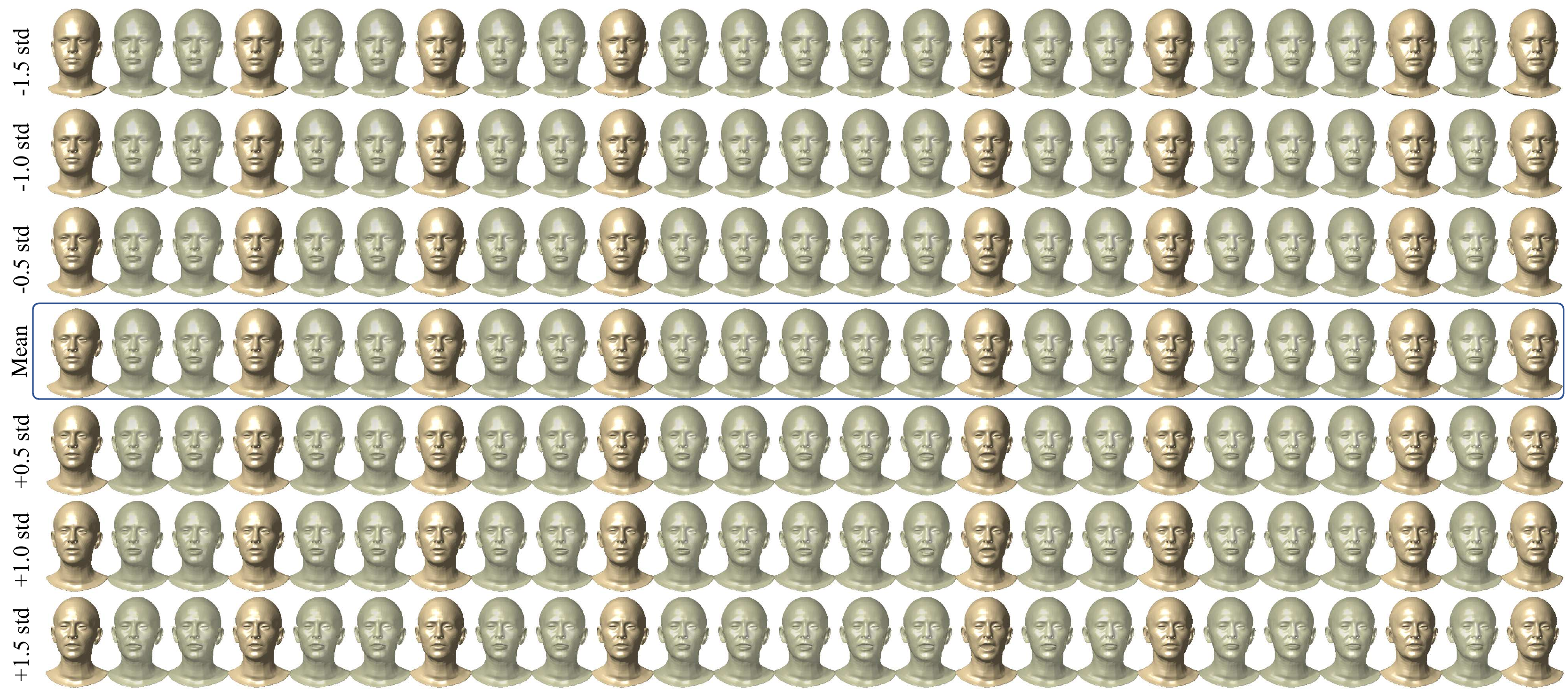}\\
		\small{(c) Third mode of variation. }
	\end{tabular}
	
	\caption{\label{fig:suppmat_modes_faces2} Three principal  modes of variation (the mean 4D surface is highlighted in the middle). Each row corresponds to one 4D surface. The supplementary material includes the full video sequences.}
\end{figure*}

Finally, Section~\ref{sec:supp_statistics}  provides additional results on the computation of summary statistics on collections of 4D surfaces. 
Figs.~\ref{fig:suppmat_means_faces2} and~\ref{fig:suppmat_modes_faces2} show statistics computed on a collection of six 4D facial expressions from the VOCA dataset. In Fig.~\ref{fig:suppmat_means_faces2}, we show the input 4D surfaces before and after the spatiotemporal registration. In both cases, we highlight the computed mean 4D  surface. Fig.~\ref{fig:suppmat_modes_faces2}  shows the leading three modes of variation; each mode of variation is discretized  as a path from $-1.5$ to $+1.5$ standard deviations.  


\bibliographystyle{IEEEtran}
\bibliography{references}
\end{document}